\documentclass[10pt,twocolumn,letterpaper]{article}

\usepackage[pagenumbers]{iccv} 

\definecolor{iccvblue}{rgb}{0.21,0.49,0.74}
\usepackage[pagebackref,breaklinks,colorlinks,allcolors=iccvblue]{hyperref}
\usepackage{xspace}
\usepackage{diagbox}
\usepackage{multirow}

\usepackage{graphicx}
\usepackage{tabularray}
\usepackage[abs]{overpic} 
\usepackage{tikz}
\usepackage{comment}
\usepackage{amsmath,amssymb} 
\usepackage{pifont}
\usepackage{color}
\usepackage{nicefrac}
\usepackage{enumitem}

\usepackage{booktabs}
\usepackage{multirow}
\usepackage{xspace}
\usepackage{rotating}

\usepackage[bottom]{footmisc}
\usepackage{url}
\usepackage{nicefrac}
\usepackage{soul}
\usepackage{wrapfig}
\usepackage{subcaption}

\usepackage{float}
\usepackage{tabularray}
\UseTblrLibrary{booktabs} 

\usepackage{pgfplotstable}
\pgfplotsset{compat=1.17} 
\usetikzlibrary{pgfplots.groupplots}

\title{MicroFlow: Domain-Specific Optical Flow for Ground Deformation \\ Estimation in Seismic Events}

\author{%
  Juliette Bertrand$^{1}$ \quad Sophie Giffard-Roisin$^{2}$ \quad James Hollingsworth$^2$ \quad Julien Mairal$^1$ \\ 
  \small $^1$Univ. Grenoble Alpes, Inria, CNRS, Grenoble INP, LJK \\ \quad
  \small $^2$Univ. Grenoble Alpes, Univ. Savoie Mont Blanc, CNRS, IRD, Univ. Gustave Eiffel, ISTerre, 38000 Grenoble
}

\begin{document}
\maketitle
\newcommand{\jb}[1]{{\color{Red}{Juliette: #1}}}
\newcommand{\jm}[1]{{\color{NavyBlue}{#1}}}
\newcommand{\julien}[1]{{\color{NavyBlue}{Julien: #1}}}
\newcommand{\sg}[1]{{\color{RubineRed}{#1}}}
\newcommand{\sophie}[1]{{\color{RubineRed}{#1}}}
\newcommand{\myparagraph}[1]{\vspace{0.1cm}\noindent {\bf #1.}}

\newcommand{\microflow}[0]{MicroFlow\xspace}
\newcommand{\microflownospace}[0]{MicroFlow}
\newcommand{\raft}[0]{RAFT\xspace}
\newcommand{\cosicorr}[0]{COSI-Corr\xspace}
\newcommand{\micmac}[0]{MicMac\xspace}
\newcommand{\gfn}[0]{GeoFlowNet\xspace}
\newcommand{\gfnnospace}[0]{GeoFlowNet}
\newcommand{\cnndis}[0]{CNN-DIS\xspace}
\newcommand{\iterative}[0]{GeoEvoNet\xspace}
\newcommand{\iterativeltv}[0]{MicroFlow\xspace}

\newcommand{\ltv}[0]{LTV\xspace}

\newcommand{\ads}[0]{ADS-80\xspace}
\newcommand{\landsat}[0]{Landsat-8\xspace}
\newcommand{\pleiades}[0]{Pleiades\xspace}
\newcommand{\sentinel}[0]{Sentinel-2\xspace}

\newcommand{\matching}[0]{patch matching\xspace}



\definecolor{cl}{HTML}{00008B}
\newcommand{\cl}{\textcolor{cl}{\bf CI}\xspace}
\definecolor{lu}{HTML}{3045D0}
\newcommand{\lu}{\textcolor{lu}{\bf ND}\xspace}
\newcommand{\ir}{\textcolor{Orchid}{\bf IR}\xspace}
\definecolor{wpg}{HTML}{F62681}
\newcommand{\wpg}{\textcolor{wpg}{\bf W}\xspace}
\newcommand{\ws}{\textcolor{Plum}{\bf SW}\xspace}
\newcommand{\il}{\textcolor{Bittersweet}{\bf IL}\xspace}
\newcommand{\ltvm}{\textcolor{Brown}{\bf LTV}\xspace}


\newcommand{\gtc}{Black}
\newcommand{\noregc}{PineGreen}
\newcommand{\tvc}{Purple}
\newcommand{\ltvc}{Red}
\newcommand{\gfnc}{gray}
\newcommand{\micmacc}{cyan}
\newcommand{\cosicorrc}{blue}



\newcommand{\stddev}[1]{\scriptsize{$\pm#1$}}
\def\l1{\ensuremath{\ell_1}\xspace}
\def\l2{\ell_2\xspace}
\newcommand{\nn}[1]{\ensuremath{\text{NN}_{#1}}\xspace}

\newcommand{\tran}{^\top}
\newcommand{\mtran}{^{-\top}}
\newcommand{\zcol}{\mathbf{0}}
\newcommand{\zrow}{\zcol\tran}

\newcommand{\ind}{\mathds{1}}
\newcommand{\expect}{\mathbb{E}}
\newcommand{\nat}{\mathbb{N}}
\newcommand{\zahl}{\mathbb{Z}}
\newcommand{\real}{\mathbb{R}}
\newcommand{\proj}{\mathbb{P}}
\newcommand{\prob}{\mathbf{Pr}}

\newcommand{\mif}{\textrm{if }}
\newcommand{\other}{\textrm{otherwise}}
\newcommand{\minimize}{\textrm{minimize }}
\newcommand{\maximize}{\textrm{maximize }}

\newcommand{\id}{\operatorname{id}}
\newcommand{\const}{\operatorname{const}}
\newcommand{\sgn}{\operatorname{sgn}}
\newcommand{\var}{\operatorname{Var}}
\newcommand{\mean}{\operatorname{mean}}
\newcommand{\trace}{\operatorname{tr}}
\newcommand{\diag}{\operatorname{diag}}
\newcommand{\vect}{\operatorname{vec}}
\newcommand{\cov}{\operatorname{cov}}

\newcommand{\softmax}{\operatorname{softmax}}
\newcommand{\clip}{\operatorname{clip}}

\newcommand{\defn}{\mathrel{:=}}
\newcommand{\peq}{\mathrel{+\!=}}
\newcommand{\meq}{\mathrel{-\!=}}

\newcommand{\floor}[1]{\left\lfloor{#1}\right\rfloor}
\newcommand{\ceil}[1]{\left\lceil{#1}\right\rceil}
\newcommand{\inner}[1]{\left\langle{#1}\right\rangle}
\newcommand{\norm}[1]{\left\|{#1}\right\|}
\newcommand{\frob}[1]{\norm{#1}_F}
\newcommand{\card}[1]{\left|{#1}\right|\xspace}
\newcommand{\diff}{\mathrm{d}}
\newcommand{\der}[3][]{\frac{d^{#1}#2}{d#3^{#1}}}
\newcommand{\pder}[3][]{\frac{\partial^{#1}{#2}}{\partial{#3^{#1}}}}
\newcommand{\ipder}[3][]{\partial^{#1}{#2}/\partial{#3^{#1}}}
\newcommand{\dder}[3]{\frac{\partial^2{#1}}{\partial{#2}\partial{#3}}}

\newcommand{\wb}[1]{\overline{#1}}
\newcommand{\wt}[1]{\widetilde{#1}}

\def\nsp{\hspace{-3pt}}
\def\xssp{\hspace{1pt}}
\def\ssp{\hspace{3pt}}
\def\msp{\hspace{6pt}}
\def\lsp{\hspace{12pt}}
\def\xlsp{\hspace{20pt}}

\newcommand{\cA}{\mathcal{A}}
\newcommand{\cB}{\mathcal{B}}
\newcommand{\cC}{\mathcal{C}}
\newcommand{\cD}{\mathcal{D}}
\newcommand{\cE}{\mathcal{E}}
\newcommand{\cF}{\mathcal{F}}
\newcommand{\cG}{\mathcal{G}}
\newcommand{\cH}{\mathcal{H}}
\newcommand{\cI}{\mathcal{I}}
\newcommand{\cJ}{\mathcal{J}}
\newcommand{\cK}{\mathcal{K}}
\newcommand{\cL}{\mathcal{L}}
\newcommand{\cM}{\mathcal{M}}
\newcommand{\cN}{\mathcal{N}}
\newcommand{\cO}{\mathcal{O}}
\newcommand{\cP}{\mathcal{P}}
\newcommand{\cQ}{\mathcal{Q}}
\newcommand{\cR}{\mathcal{R}}
\newcommand{\cS}{\mathcal{S}}
\newcommand{\cT}{\mathcal{T}}
\newcommand{\cU}{\mathcal{U}}
\newcommand{\cV}{\mathcal{V}}
\newcommand{\cW}{\mathcal{W}}
\newcommand{\cX}{\mathcal{X}}
\newcommand{\cY}{\mathcal{Y}}
\newcommand{\cZ}{\mathcal{Z}}

\newcommand{\vA}{\mathbf{A}}
\newcommand{\vB}{\mathbf{B}}
\newcommand{\vC}{\mathbf{C}}
\newcommand{\vD}{\mathbf{D}}
\newcommand{\vE}{\mathbf{E}}
\newcommand{\vF}{\mathbf{F}}
\newcommand{\vG}{\mathbf{G}}
\newcommand{\vH}{\mathbf{H}}
\newcommand{\vI}{\mathbf{I}}
\newcommand{\vJ}{\mathbf{J}}
\newcommand{\vK}{\mathbf{K}}
\newcommand{\vL}{\mathbf{L}}
\newcommand{\vM}{\mathbf{M}}
\newcommand{\vN}{\mathbf{N}}
\newcommand{\vO}{\mathbf{O}}
\newcommand{\vP}{\mathbf{P}}
\newcommand{\vQ}{\mathbf{Q}}
\newcommand{\vR}{\mathbf{R}}
\newcommand{\vS}{\mathbf{S}}
\newcommand{\vT}{\mathbf{T}}
\newcommand{\vU}{\mathbf{U}}
\newcommand{\vV}{\mathbf{V}}
\newcommand{\vW}{\mathbf{W}}
\newcommand{\vX}{\mathbf{X}}
\newcommand{\vY}{\mathbf{Y}}
\newcommand{\vZ}{\mathbf{Z}}

\newcommand{\va}{\mathbf{a}}
\newcommand{\vb}{\mathbf{b}}
\newcommand{\vc}{\mathbf{c}}
\newcommand{\vd}{\mathbf{d}}
\newcommand{\ve}{\mathbf{e}}
\newcommand{\vf}{\mathbf{f}}
\newcommand{\vg}{\mathbf{g}}
\newcommand{\vh}{\mathbf{h}}
\newcommand{\vi}{\mathbf{i}}
\newcommand{\vj}{\mathbf{j}}
\newcommand{\vk}{\mathbf{k}}
\newcommand{\vl}{\mathbf{l}}
\newcommand{\vm}{\mathbf{m}}
\newcommand{\vn}{\mathbf{n}}
\newcommand{\vo}{\mathbf{o}}
\newcommand{\vp}{\mathbf{p}}
\newcommand{\vq}{\mathbf{q}}
\newcommand{\vr}{\mathbf{r}}
\newcommand{\Vs}{\mathbf{s}}
\newcommand{\vt}{\mathbf{t}}
\newcommand{\vu}{\mathbf{u}}
\newcommand{\vv}{\mathbf{v}}
\newcommand{\vw}{\mathbf{w}}
\newcommand{\vx}{\mathbf{x}}
\newcommand{\vy}{\mathbf{y}}
\newcommand{\vz}{\mathbf{z}}
\newcommand{\vdf}{\mathbf{df}}

\newcommand{\vone}{\mathbf{1}}
\newcommand{\vzero}{\mathbf{0}}

\newcommand{\valpha}{{\boldsymbol{\alpha}}}
\newcommand{\vbeta}{{\boldsymbol{\beta}}}
\newcommand{\vgamma}{{\boldsymbol{\gamma}}}
\newcommand{\vdelta}{{\boldsymbol{\delta}}}
\newcommand{\vepsilon}{{\boldsymbol{\epsilon}}}
\newcommand{\vzeta}{{\boldsymbol{\zeta}}}
\newcommand{\veta}{{\boldsymbol{\eta}}}
\newcommand{\vtheta}{{\boldsymbol{\theta}}}
\newcommand{\viota}{{\boldsymbol{\iota}}}
\newcommand{\vkappa}{{\boldsymbol{\kappa}}}
\newcommand{\vlambda}{{\boldsymbol{\lambda}}}
\newcommand{\vmu}{{\boldsymbol{\mu}}}
\newcommand{\vnu}{{\boldsymbol{\nu}}}
\newcommand{\vxi}{{\boldsymbol{\xi}}}
\newcommand{\vomikron}{{\boldsymbol{\omikron}}}
\newcommand{\vpi}{{\boldsymbol{\pi}}}
\newcommand{\vrho}{{\boldsymbol{\rho}}}
\newcommand{\vsigma}{{\boldsymbol{\sigma}}}
\newcommand{\vtau}{{\boldsymbol{\tau}}}
\newcommand{\vupsilon}{{\boldsymbol{\upsilon}}}
\newcommand{\vphi}{{\boldsymbol{\phi}}}
\newcommand{\vchi}{{\boldsymbol{\chi}}}
\newcommand{\vpsi}{{\boldsymbol{\psi}}}
\newcommand{\vomega}{{\boldsymbol{\omega}}}

\newcommand{\rLambda}{\mathrm{\Lambda}}
\newcommand{\rSigma}{\mathrm{\Sigma}}

\begin{abstract}

Dense ground displacement measurements are crucial for geological studies but
are impractical to collect directly. Traditionally, displacement fields are
estimated using patch matching on optical satellite images from 
different acquisition times. While deep learning-based optical flow models are
promising, their adoption in ground deformation analysis is hindered by
challenges such as the absence of real ground truth, the need for sub-pixel
precision, and temporal variations due to geological or anthropogenic changes.
In particular, we identify that deep learning models relying on explicit
correlation layers struggle at estimating small displacements in real-world
conditions.  Instead, we propose a model that employs iterative refinements
with explicit warping layers and a correlation-independent backbone, enabling
sub-pixel precision. Additionally, a non-convex variant of Total Variation
regularization preserves fault-line sharpness while maintaining smoothness
elsewhere.  Our model significantly outperforms widely used geophysics methods
on semi-synthetic benchmarks and generalizes well to challenging real-world
scenarios captured by both medium- and high-resolution sensors. Project page: \url{https://jbertrand89.github.io/microflow/}.

\end{abstract}

\begin{figure*}[t]
\vspace{-10pt} 
\centering
\setlength\tabcolsep{2pt}
\begin{tblr}{
  colspec = {X[c,0.2]X[c,h]X[c,h]X[c,0.02]X[c,h]X[c,h]X[c,0.02]X[c,h]X[c,h]X[c,h]X[c,0.5]},
  stretch = 0,
  rowsep = 2pt,
  hlines = {red5, 0pt},
  vlines = {red5, 0pt},
  cell{1}{2} = {c=2}{c}, 
}
& \small\textbf{Image pair}& & &\small\textbf{\shortstack{1. \cosicorr}} & \textbf{\small2. \micmac} & & \textbf{\shortstack{\small3. \gfn}} & \textbf{\small4. RAFT} & \textbf{\small5. Ours} \\
  (a) & \includegraphics[width=2cm]{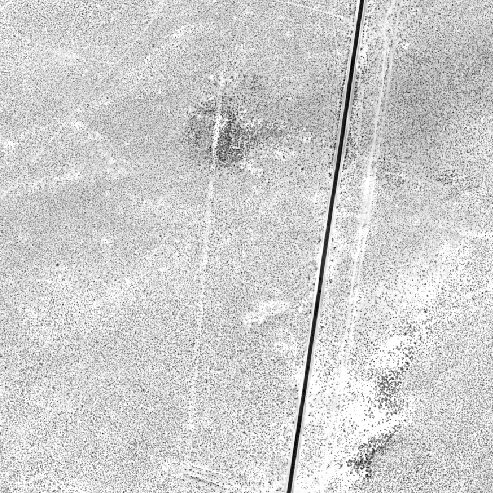} &  \includegraphics[width=2cm]{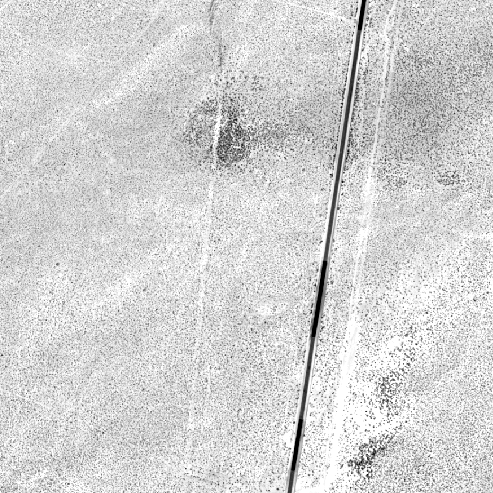}& &
\includegraphics[width=2cm]{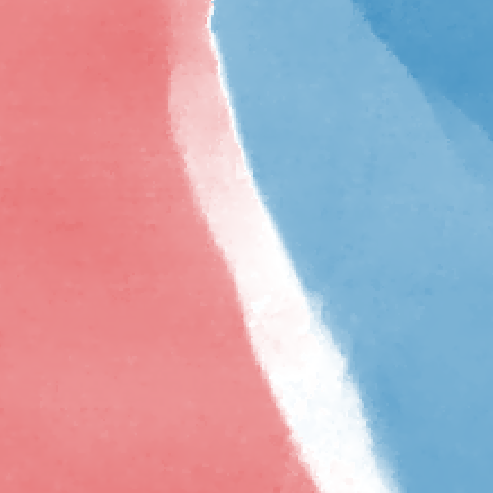}&
\includegraphics[width=2cm]{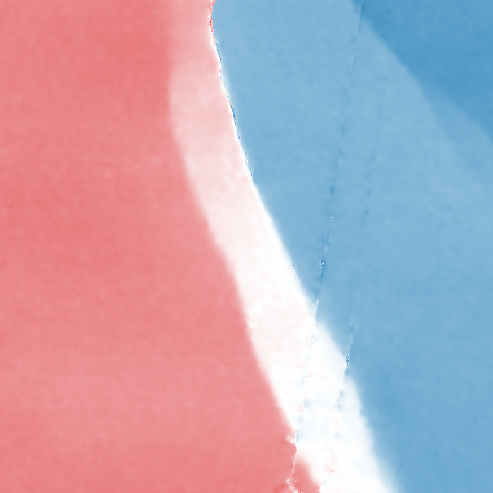}& & 
\includegraphics[width=2cm]{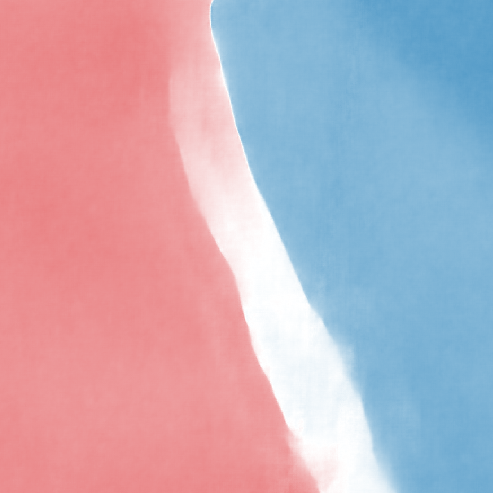}&
\includegraphics[width=2cm]{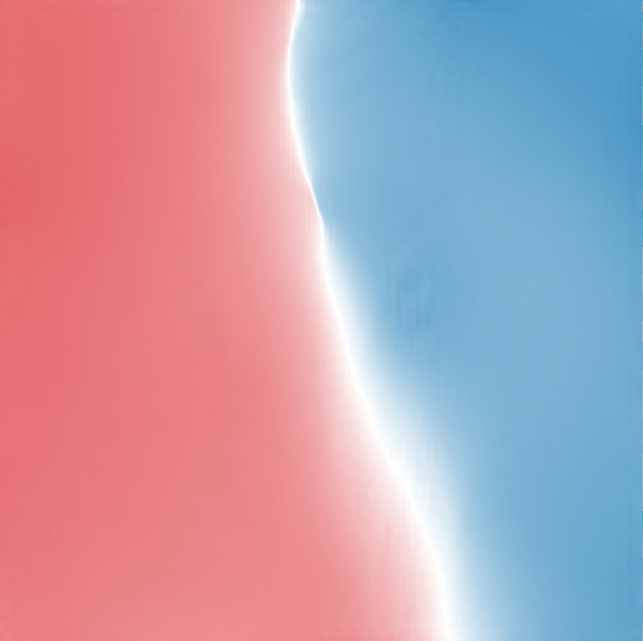}& 
\includegraphics[width=2cm]{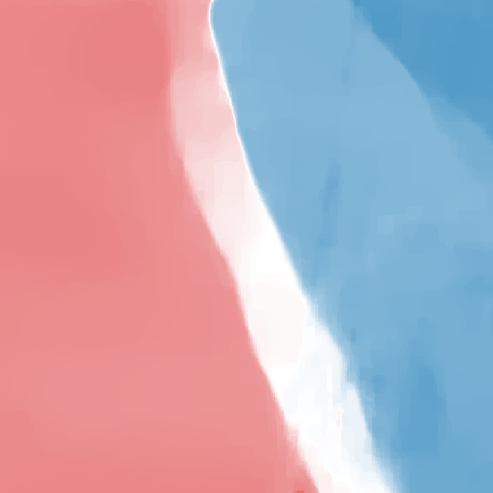} &
\includegraphics[height=1.7cm]{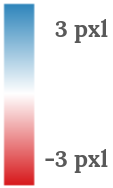}
\\
(b) & \includegraphics[width=2cm]{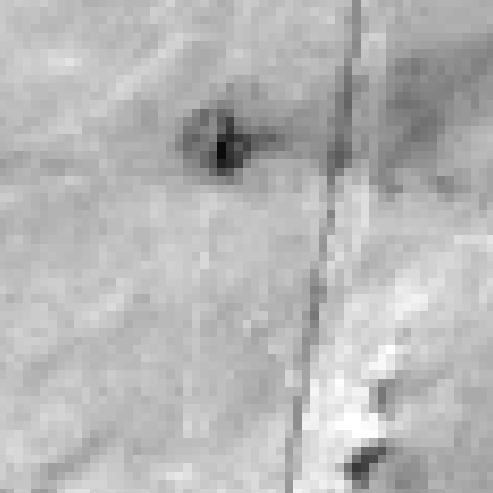}&
\includegraphics[width=2cm]{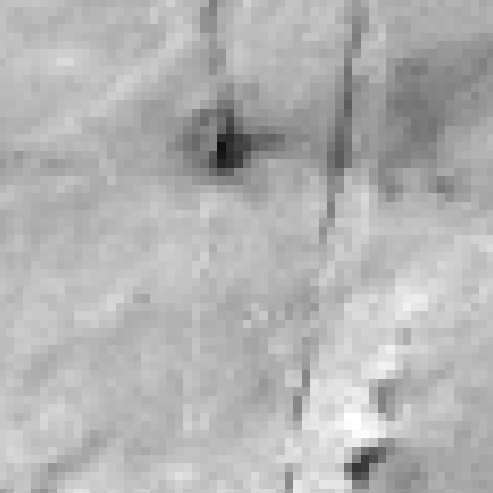}& & 
\includegraphics[width=2cm]{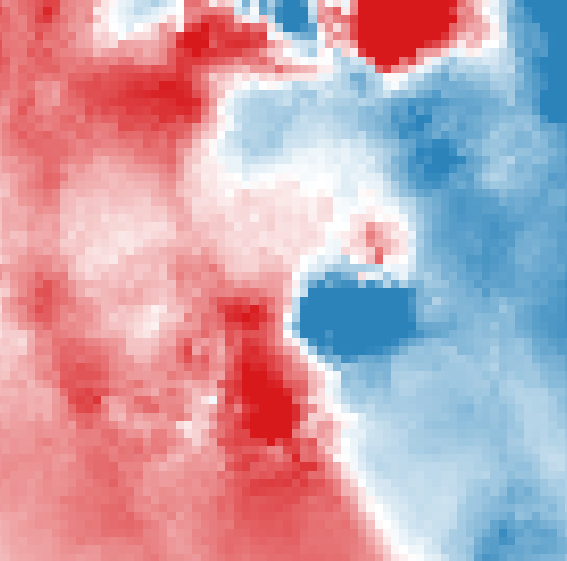}& 
\includegraphics[width=2cm]{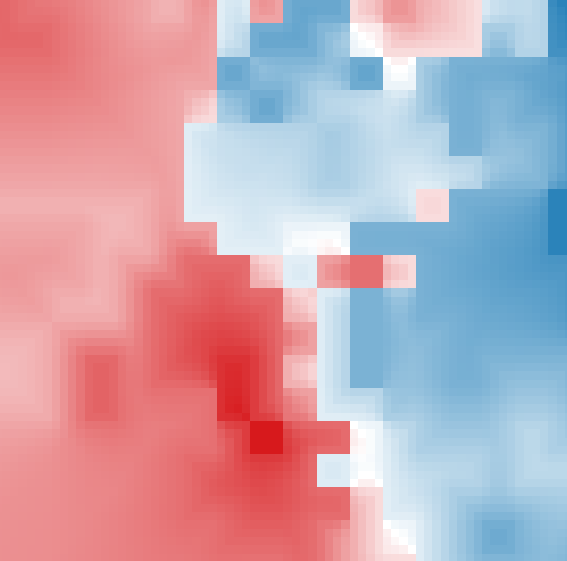}& & 
\includegraphics[width=2cm]{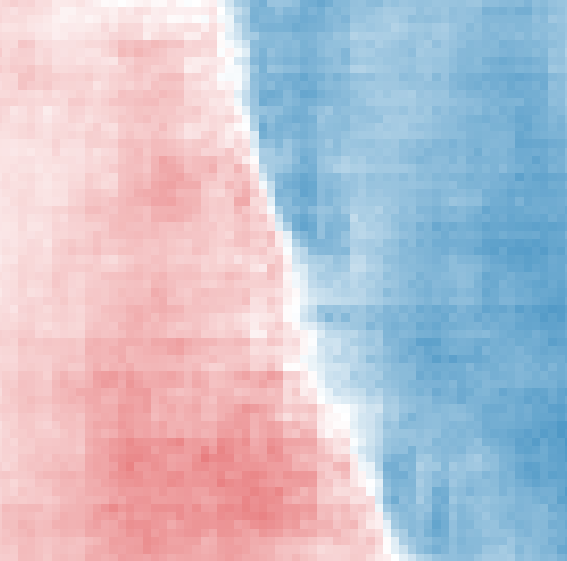}& 
\includegraphics[width=2cm]{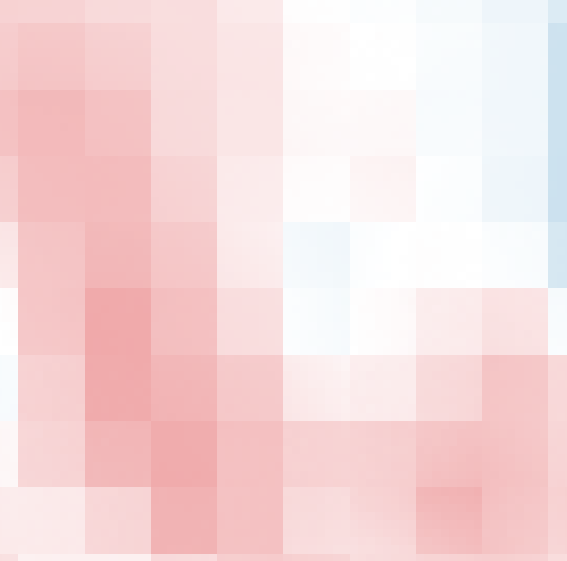}&
\includegraphics[width=2cm]{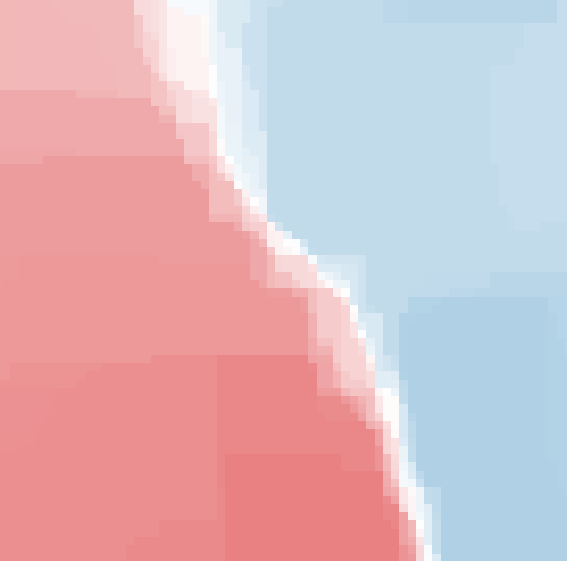} &
\includegraphics[height=1.8cm]{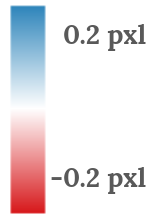}
\\
(c) & \includegraphics[width=2cm]{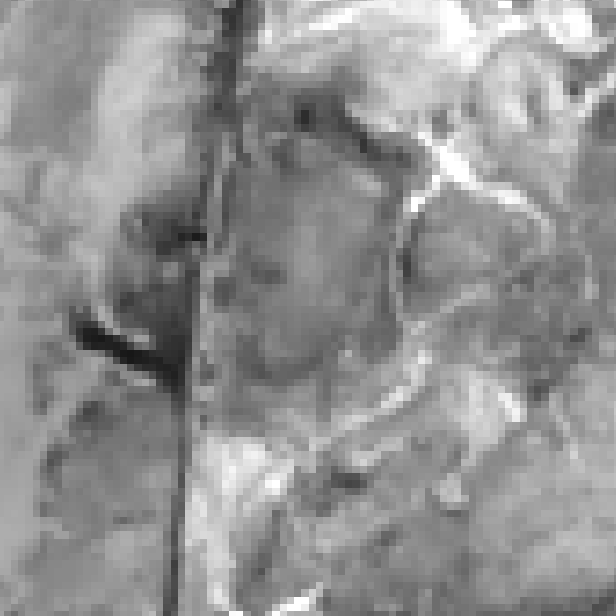}&
\includegraphics[width=2cm]{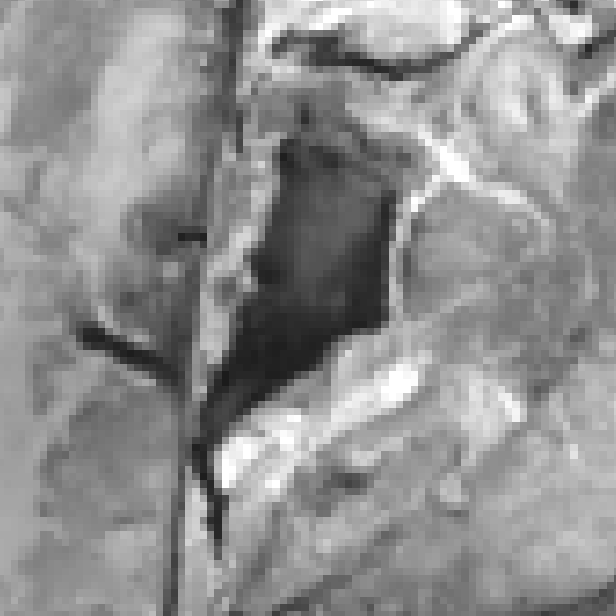}& &
\includegraphics[width=2cm]{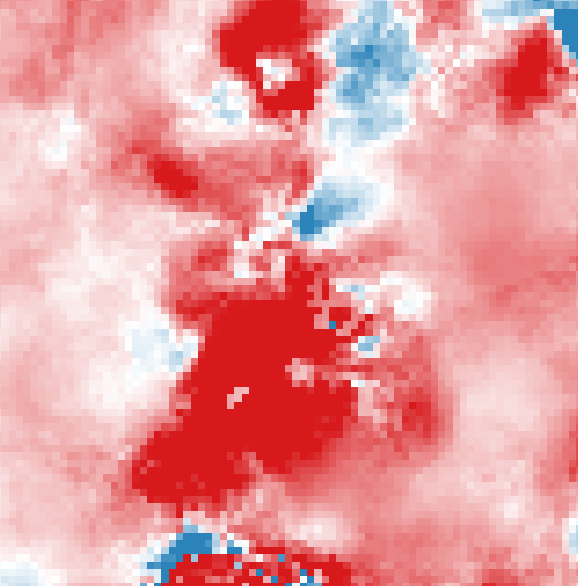}&  
\includegraphics[width=2cm]{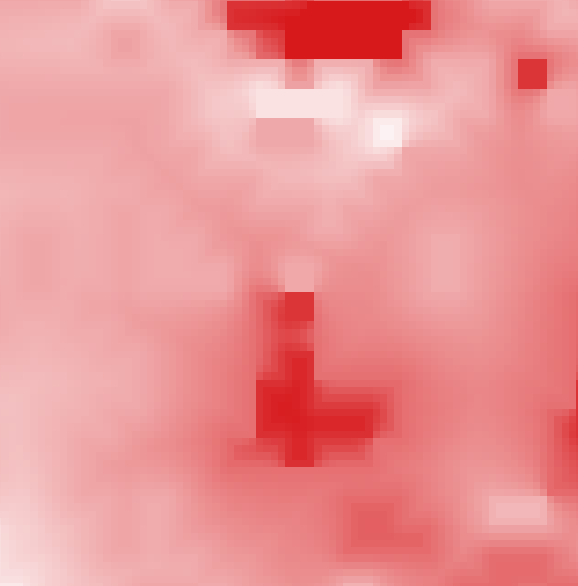}& &
\includegraphics[width=2cm]{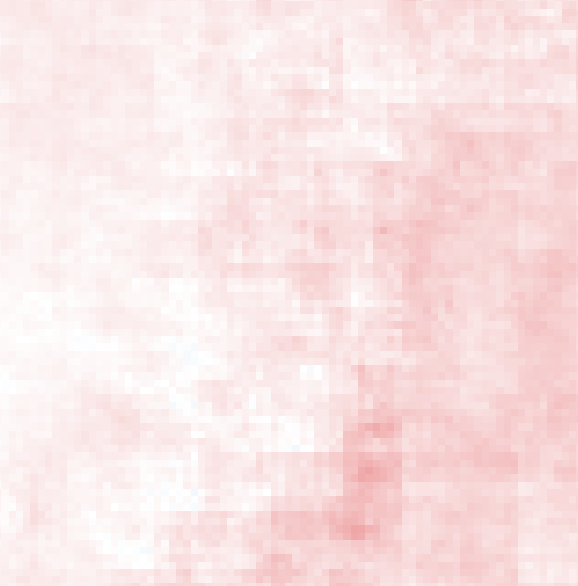}&  
\includegraphics[width=2cm]{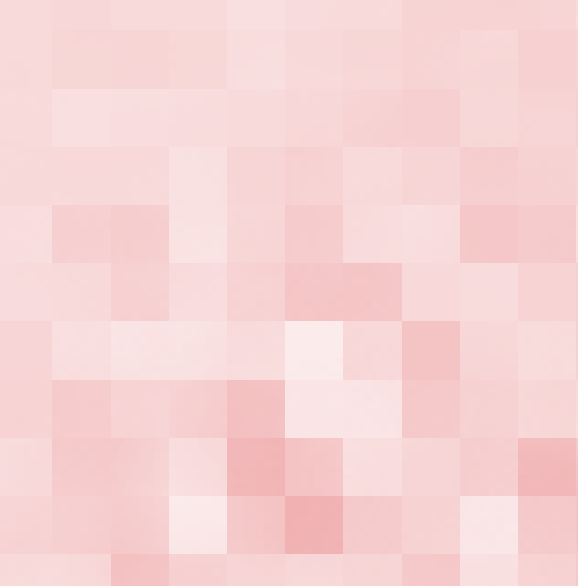}&
\includegraphics[width=2cm]{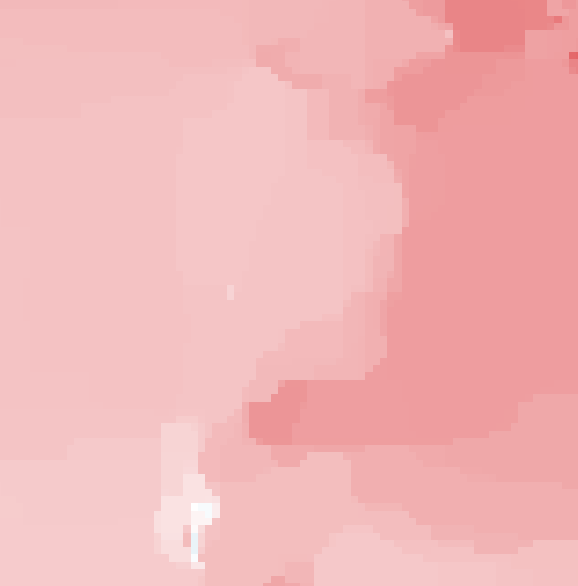}&
\includegraphics[height=1.8cm]{fig/00_scales/scale_landsat.png}
\\
(d) & \includegraphics[width=2cm]{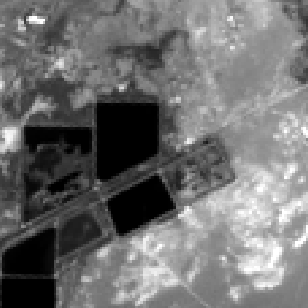}&
\includegraphics[width=2cm]{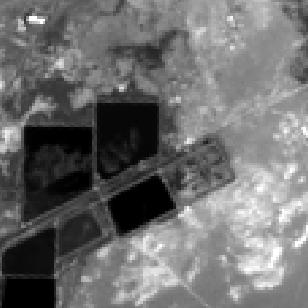}& &
\includegraphics[width=2cm]{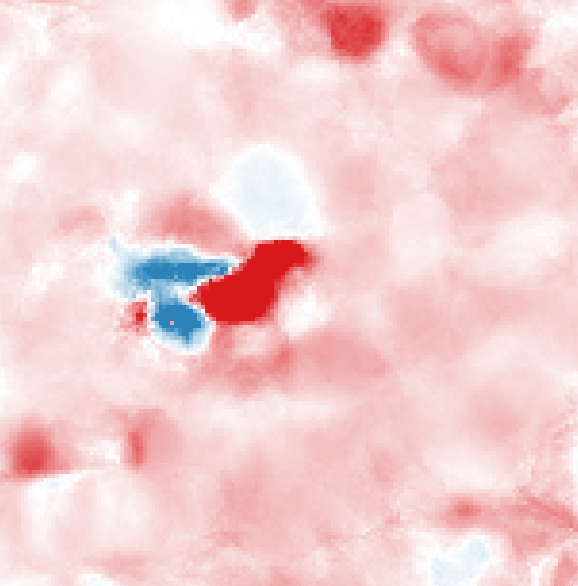}& 
\includegraphics[width=2cm]{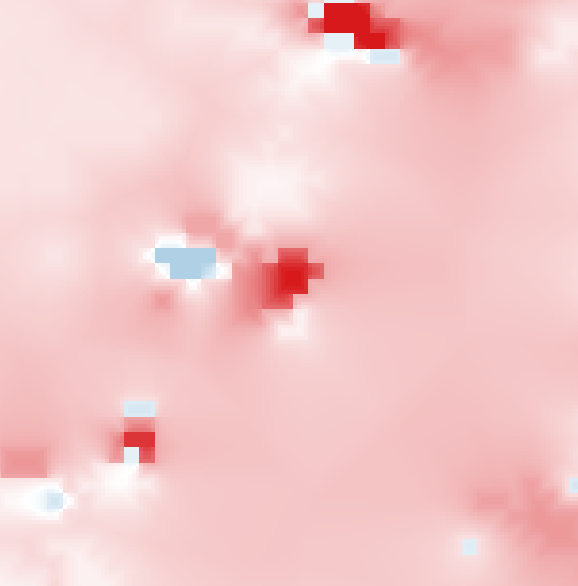}& &
\includegraphics[width=2cm]{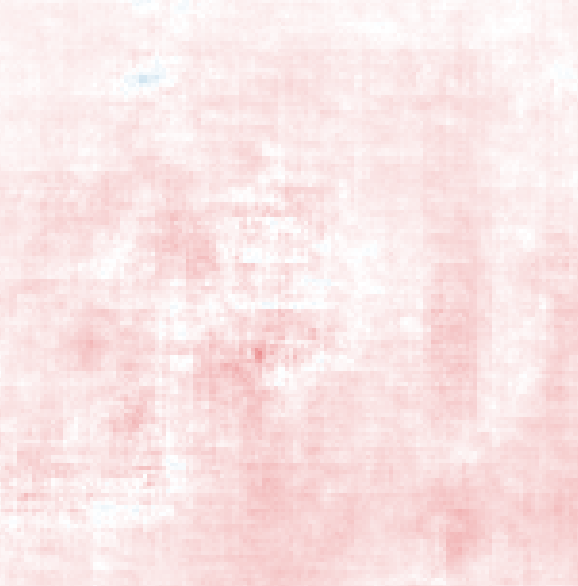}& 
\includegraphics[width=2cm]{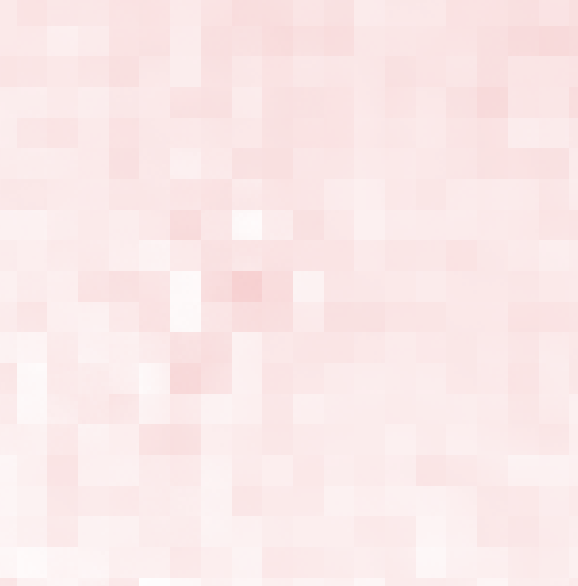}&
\includegraphics[width=2cm]{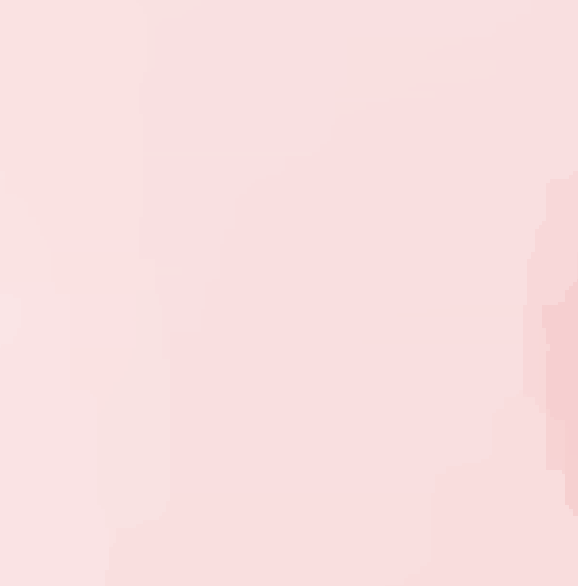} &
\includegraphics[height=1.8cm]{fig/00_scales/scale_landsat.png}
\\
\end{tblr}

\vspace{-5pt}
\caption{\textbf{Going from (a) high-resolution to (b)-(d) medium-resolution remote sensing optical images acquired before and after real seismic events} and the relative North-South displacement estimations. Classical models (1-2) become overly sensitive to temporal change while state-of-the-art data-driven models (3-4) fail to achieve subpixel precision. 
In the small to very small displacement regime, encoder-decoder models need iterative refinements and a non convex a-posteriori regularization to effectively estimate dense ground displacement fields (5). It boosts the localization and sharpness of seismic faults both for (a) high resolution and (b) medium resolution sensors, and improves the robustness to external perturbations of the signal such as (c) geological activities and (d) human activities, among others. 
\vspace{-3pt}
}
\label{fig:teaser}
\end{figure*}

\vspace*{-0.3cm}
\section {Introduction}
\label{sec:introduction}

Dense Earth surface displacement measurements are crucial for applications like monitoring co-seismic deformation \cite{Vanpuymbroeck00, LeprinceA07, Rosu15, Cheng21}, ice-flow dynamics \cite{Scambos92, Berthier05, Heid12}, and sand dune migrations \cite{hermas12}. In seismology, they help understand rupture fault geometry \cite{Michel02} and earthquake energy release \cite{Leprince07, Marchandon21}. While GNSS stations (e.g., GPS) provide precise measurements, their sparse distribution makes dense field measurements impractical. Optical geodesy offers a promising alternative, estimating dense displacement fields from pre- and post-seismic remote sensing images.

Dense displacement field estimation traditionally relies on computationally demanding \matching techniques \cite{Leprince07, Rosu15, montagnon2024sub}. Recently, deep learning has emerged as a promising alternative, with \gfn \cite{Montagnon24} being the first model to estimate dense ground displacement fields using an encoder-decoder framework and a semi-synthetic training dataset. 
While \gfn performs well near fault lines, it lags behind traditional methods farther from the fault.
Excelling in large video motion estimation tasks, \raft \cite{teed2020raft} is a natural candidate for ground deformation estimation. However, when trained using standard protocols~\cite{sun2022disentangling} with co-seismic data, the model struggles to capture small displacements and adapt to the unique challenges of ground deformation estimation (see Figure~\ref{fig:teaser}).

A key challenge is the lack of dense ground-truth data, requiring training on a semi-synthetic dataset generated using an Earth deformation simulator \cite{Montagnon24}. Seismic deformation typically causes broad, smooth changes with abrupt shifts near fault zones, demanding sub-pixel precision in displacement estimation, whether from high-resolution or medium-resolution sensors. Achieving model robustness for sub-pixel displacements is difficult, as it may lead to trivial zero-displacement solutions. Additionally, time gaps between images (typically weeks to months) introduce non-tectonic variations such as geological and anthropogenic change, complicating precise detection. 

In this work, we identify that explicit correlation layers (present in most recent deep learning optical flow architectures \cite{ilg2017flownet, sun2018pwc, teed2020raft, huang2022flowformer}) fail at estimating small to very small displacement and that iterative refinements with explicit warping layers and a correlation independent backbone achieve sub-pixel precision but produce noisy estimate because of temporal perturbations. To achieve
better generalization in real-world data, an appropriate regularization is thus needed. For that, we show that a non-convex variant of Total Variation is adapted to displacement fields with low-frequency patterns and sharp fault-line details.

Evaluating machine learning methods in scientific applications is challenging due to limited ground truth data and domain-specific expectations,  hard to capture with standard metrics. In this study, we evaluate our framework in two settings: (i) a quantitative comparison using semi-synthetic data from real satellite image pairs modified with simulated seismic impacts, and (ii) a qualitative evaluation on a real-world dataset from California’s Ridgecrest earthquake zone (2019), including high- and medium-resolution examples from three seismic shocks. Our framework demonstrates strong accuracy on the semi-synthetic benchmark and meets key geophysical expectations on real-world examples: precisely located and sharp fault discontinuities and smooth 
estimates in regions farther from the fault. 

\noindent In summary, our contributions are three-fold:
\vspace*{-0.1cm}
\begin{itemize}
\item We introduce an optical flow model specifically designed for ground deformation 
estimation, trained on semi-synthetic co-seismic data. Our model outperforms 
traditional geoscientific approaches, generalizes well to real-world scenarios
with small to very small displacements, preserves the photometric quality of
classical models, and is more robust to temporal perturbations.
\item Through an analysis of generic optical flow models, we identify that those 
relying on explicit correlation layers struggle with small displacements and
fail to generalize effectively to real-world data.
\item
We demonstrate the importance of two key components: (1) iterative refinements 
with explicit warping layers on a correlation-independent backbone, enabling
sub-pixel precision, and (2) a non-convex variant of Total Variation
regularization, which preserves fault-line sharpness while maintaining
smoothness elsewhere.
\end{itemize}

\vspace*{-0.2cm}
\section{Related Work}
\label{sec:related}

\myparagraph{Dense optical flow estimation}
Dense displacement field estimation for video motion is widely studied, spanning global variational methods \cite{horn1981determining, brox2004high}, local gradient-based techniques \cite{lucas1981iterative}, and \matching approaches \cite{weinzaepfel2013deepflow, revaud2015epicflow, revaud2016deepmatching, thewlis2016fully}. FlowNet \cite{dosovitskiy2015flownet} introduced end-to-end learning but falls short of traditional methods. FlowNet2 \cite{ilg2017flownet}, with stacked iterative refinements, surpasses them, validating the encoder-decoder paradigm. Since then, various improvements have tackled video motion estimation challenges.

A key challenge in video motion estimation is handling large displacement fields \cite{brox2009large}, addressed through explicit correlation layers and iterative refinements. FlowNet-C \cite{dosovitskiy2015flownet} introduced explicit correlation layers for feature comparison, later enhanced with pyramidal structures for multi-scale learning \cite{Ranjan_2017_CVPR, hui18liteflownet, sun2018pwc, sun2019models}. To improve efficiency, their search space is limited, restricting large displacement capture. Iterative refinements with warping layers \cite{ilg2017flownet, hur2019iterative} progressively refine flow estimates but can strain deep architectures \cite{dosovitskiy2015flownet, sun2018pwc, hur2019iterative}. RAFT \cite{teed2020raft} eliminates explicit warping layers with a GRU-based recurrent architecture, enabling over 10 iterations for better large-displacement estimation. Follow-up work improves correlation layers \cite{hofinger2020improving, Zhang_2021_ICCV, shi2023flowformer++} and adopts transformer-based backbones \cite{huang2022flowformer, shi2023flowformer++}. Beyond architecture, studies highlight the critical role of training datasets and data schedulers \cite{dosovitskiy2015flownet, ilg2017flownet, mayer2016large, sun2019models, sun2021autoflow, saxena2024surprising}.

Small displacements are underexplored in motion estimation due to their scarcity in video datasets. While FlowNet-SD \cite{ilg2017flownet} handles them with a non-iterative design, sub-pixel precision remains overlooked, offering little benefit for large displacements. Here, we re-introduce  iterative refinements with explicit warping layers for sub-pixel accuracy, removing reliance on the explicit correlation layer, which struggles with small to very small displacements.

\myparagraph{Dense optical flow estimation for optical geodesy}
In optical geodesy, dense displacement field estimation is typically done through \matching \cite{Leprince07, Rosu15, montagnon2024sub}, where each pixel in the second image is matched to a corresponding window in the first image. Unlike video motion estimation, which focuses on large displacements \cite{weinzaepfel2013deepflow, revaud2015epicflow, thewlis2016fully}, estimating dense sub-pixel displacements requires precise correlation within a sub-pixel search space, making it computationally demanding.
To address this, \gfn \cite{Montagnon24} leverages the FlowNet-SD architecture and introduces FaultDeform, a synthetic dataset for ground deformation. However, traditional video motion models are not optimized for sub-pixel accuracy in small displacements, limiting \gfnnospace’s effectiveness compared to \matching methods designed for this precise task.
\begin{table*}[t]
    \centering
    \resizebox{0.9\textwidth}{!}{%

\centering
\setlength{\tabcolsep}{4pt}
\begin{tabular}{cccccccc}
    \toprule
    \multirow{2}{*}{Method}  & Correlation  & Iterative & \multirow{2}{*}{Warping} & Separate & Intermediate &  \multirow{2}{*}{Regularization} \\
     & Independent  & Refinements & & Weights & Loss &  \\
     & \cl &   \ir & \wpg &  \ws& \il &  \ltvm \\
     \midrule
    FlowNet-C \cite{dosovitskiy2015flownet}& \ding{55} & \ding{55}& - & - & - & \ding{55}
    \\ 
    FlowNet-S \cite{dosovitskiy2015flownet} & \checkmark & \ding{55}& - & - & - & \ding{55}
    \\ 
    FlowNet-SD \cite{ilg2017flownet} / \gfn \cite{Montagnon24} &  \checkmark & \ding{55}& - & - & - & \ding{55}
    \\
    FlowNet2 \cite{ilg2017flownet} & \ding{55}& \checkmark& \checkmark & \checkmark & \ding{55} & \ding{55}
    \\
    
    IRR-FlowNet-S \cite{hur2019iterative} & \checkmark& \checkmark& \checkmark & \ding{55} & \ding{55} & \ding{55}
    \\
    \raft \cite{teed2020raft} & \ding{55}&   \checkmark & \ding{55} & - & \checkmark & \ding{55}
    
    \\
    \midrule
    Ours (joint weights)  & \checkmark& \checkmark & \checkmark & \ding{55} & \checkmark & \checkmark
    \\
    Ours (separate weights) & \checkmark& \checkmark & \checkmark & \checkmark & \checkmark & \checkmark
    \\

\end{tabular}

    }
    \vspace{-7pt} 
    \caption{\textbf{Essential components of our proposed method \microflow} and how they compare with standard optical flow models primarily designed for solving large video motions.}
    \label{tab:method_components}
    \vspace{-5pt} 
\end{table*}

\myparagraph{Regularizations for dense optical flows}
Variational approaches \cite{horn1981determining} frame optical flow estimation as a minimization problem, balancing data fidelity (brightness consistency) with gradient smoothness regularization. Early methods used a quadratic penalty \cite{horn1981determining}, while later ones employed an $L1$ penalty \cite{zach2007duality} to handle outliers and preserve flow discontinuities. Regularization is crucial as the data term alone is often under-constrained. In contrast, data-driven models incorporate regularization via tailored loss functions 
in recent video motion methods \cite{teed2020raft, sun2022disentangling, huang2022flowformer, shi2023videoflow, dong2024memflow}.
\section{Method}
\label{sec:method}

\begin{figure*}[ht]
\centering
\includegraphics[width=18cm]{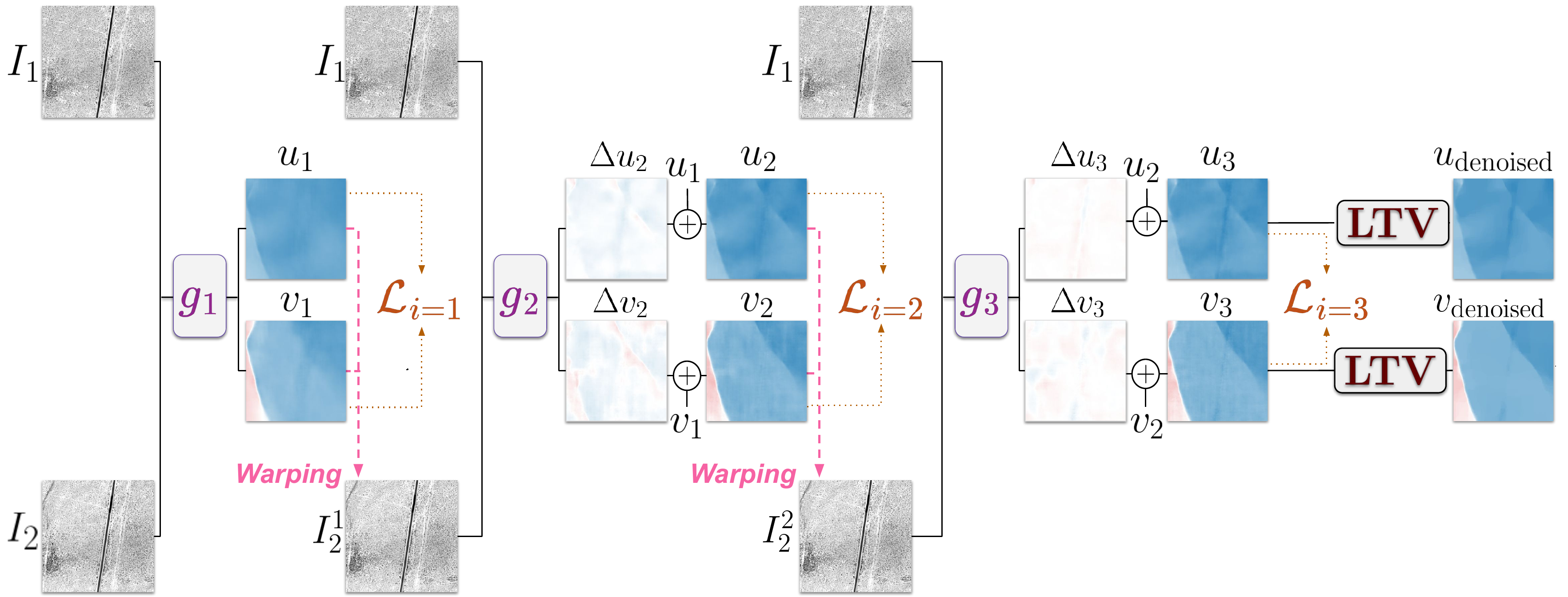}
\vspace{-8pt}
\caption{\textbf{Overview of \microflow for estimation dense ground displacement fields}. \microflow regresses dense displacement fields ($u_3$, $v_3$) by jointly encoding images $I_1$ and $I_2$ through a set of correlation independent encoder-decoder networks ($g_1$, $g_2$, $g_3$) with iterative refinements. It achieves sub-pixel precision and recovers missing faults (notably in the top right corner). An a-posteriori non-convex \ltv regularization reveals the original displacement field, maintaining the sharpness of the fault while preserving smoothness away from the fault to correct noisy estimates caused by temporal change.
}
\label{fig:method}
\vspace{-3pt}
\end{figure*}

\begin{figure}[ht]
\begin{subfigure}{\linewidth}
    \vspace{-10pt}
    \centering
    \includegraphics[width=8cm]{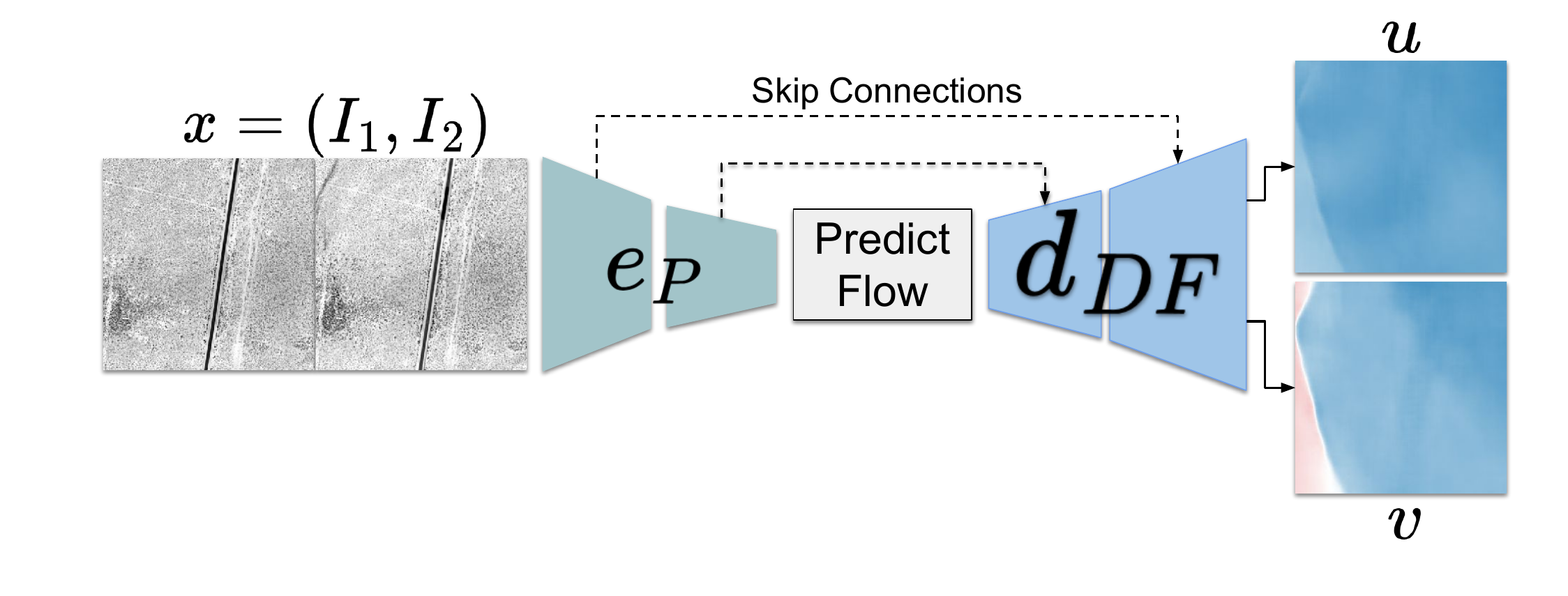}
     \vspace{-10pt}
    \caption{Correlation independent encoder-decoder networks.}
    \label{fig:corr_independent}
\end{subfigure}
\begin{subfigure}{\linewidth}
    \centering
    \includegraphics[width=8cm]{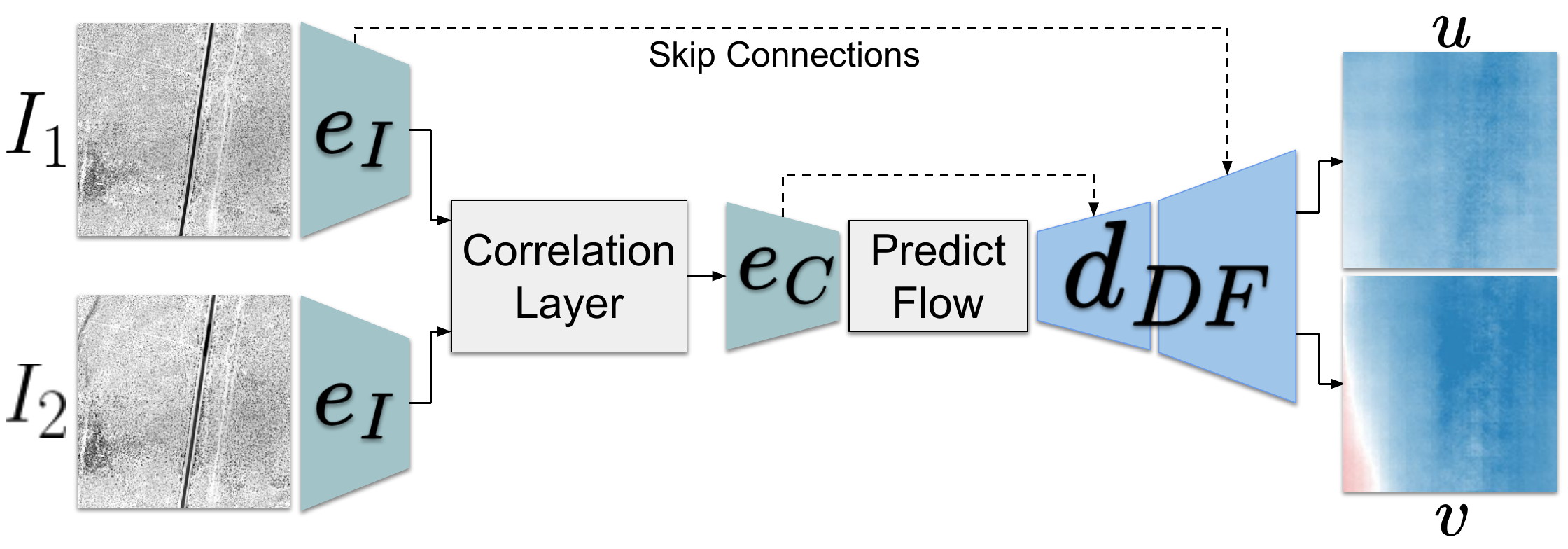}
     \vspace{-7pt}
    \caption{Correlation dependent encoder-decoder networks.}
    \label{fig:corr_dependent}
\end{subfigure}
\vspace{-8pt}
\caption{\textbf{Correlation dependent vs Correlation independent encoder-decoder networks}. Standard architectures rely on correlation dependent encoder-decoder networks, which are not robust to very small displacement fields. Instead, we use a correlation independent network as the backbone of our iterative procedure. 
}
\label{fig:correlation_method}
\vspace{-10pt}
\end{figure}

Given a pair of grayscale images $I_1$ and $I_2$, our goal is to estimate a dense displacement field $\vdf = (\vu, \vv)$ that maps each pixel ($x$, $y$) in $I_2$ to its corresponding position $(x', y') = (x+\vu(x,y), y + \vv(x,y))$ in $I_1$. 

In video motion estimation \cite{brox2009large, revaud2016deepmatching, dosovitskiy2015flownet}, image pairs are captured within fractions of a second. In optical geodesy, however, acquisition gaps span weeks to months, introducing perturbations that alter scene appearance and interact with ground displacement.
In this context, estimating the dense ground displacement field $\vdf_{\text{earth}} = (\vu_{\text{earth}}, \vv_{\text{earth}})$ requires inferring the mapping between points on the Earth’s surface, which doesn't translate exactly to a mapping between pixel intensities in the images.

Data-driven approaches use strong priors on the ground deformation field to directly estimate $\vdf_{\text{earth}}$. However, the lack of real-world annotations and the variability of temporal changes relative to displacement magnitude often lead to noisy estimates, $\vdf_{\text{noisy}}$. This poses an inverse problem: recovering the true displacement field $\vdf_{\text{earth}}$ from $\vdf_{\text{noisy}}$.

Guided by this observation, we propose a two-step approach to recover the true ground deformation field $\vdf_{\text{earth}}$. First, a correlation-independent encoder-decoder (Sec. \ref{sec:encoder_decoder}) with iterative refinements using explicit warping (Sec. \ref{sec:iterative}) estimates a noisy displacement field $\vdf_{\text{noisy}}$. Then, a regularization process (Sec. \ref{sec:regularization}) refines it to recover $\vdf_{\text{earth}}$.

\subsection{Encoder-decoder architecture}
\label{sec:encoder_decoder}

We estimate the displacement field $\vdf_{\text{noisy}}$, from the input images $I_1$ and $I_2 \in \real^{H \times W}$, using a modified U-Net architecture $g$ tailored for displacement field estimation. The encoder $e$ extracts features ($f_1$, $f_2$, ..., $f_r$) at each resolution ($2^1$, ... $2^r$) and at the coarsest level $2^r$, an initial displacement field $df_k=h(f_r)$ is computed. The decoder $d_{DF}$ progressively upsamples and refines $df_r$ through finer resolutions ($df_{r-1}$, ... $df_1$), using the corresponding features ($f_1$, $f_2$, ..., $f_r$), until the final displacement field $\vdf_{\text{noisy}} = df_{1}$ is obtained. We consider two types of encoders.

\myparagraph{Correlation independent encoder network (Figure \ref{fig:corr_independent})}
The input images are processed jointly. The stacked input $\vx = (I_1, I_2) \in \real^{H \times W \times 2}$ is processed by the encoder $e_{P}$, producing joint feature maps ($f_1$,  ..., $f_r$) at each resolution.

\myparagraph{Correlation dependent encoder network (Figure \ref{fig:corr_dependent})}
Each input image is processed separately by an image encoder network, $e_{\text{I}}$, generating two feature maps $f_{j,1}$ and $f_{j,2} \in \real^{h \times w \times c}$ at resolution $2^j$ with $j \in [1, r-1]$. Correspondences between the features maps $f_{j,1}$ and $f_{j,2}$ are extracted to form a 3D-correlation volume $C \in \real^{h \times w \times D}$, 
with $D=d^2$ defined as its number of planes, and $h=H/2^{j}$ and $w = W/2^{j}$ the spatial resolution of the feature maps, and $c$ the number of feature channels. This volume, computed at resolution $2^j$ or hierarchically, is then processed by a correlation encoder $e_{\text{C}}$ to produce feature maps at finer resolutions ($2^{j+1}$, ... $2^{r}$).
We use FlowNet-SD as our correlation-independent encoder-decoder, essential for sub-pixel precision in small displacements (Section \ref{sec:experiments}).

\subsection{Iterative refinements with explicit warping}
\label{sec:iterative}
Instead of estimating the displacement field in a single step, we iteratively refine it through a sequence $(\vdf_1, \vdf_2, ..., \vdf_n)$ \cite{ilg2017flownet, hur2019iterative}, where $n$ is the maximum number of iterations. In each iteration, we compute an update $\mathbf{\Delta df_i}$ as $\vdf_i = \vdf_{i-1} + \mathbf{\Delta df_i}$. The refined field $df_i$ is used to warp $I_2$ via a Spatial Transformer Network \cite{jaderberg2015spatial}, producing $I_2^i$, which is stacked with $I_1$ for the next iteration. The updated field is given by $\mathbf{\Delta df_i} = g_{i}(\vx_i)$. Networks $(g_1, ..., g_n)$ can either have separate weights ($g_1 \neq g_2 \neq g_n$) or share weights ($g_1 = g_2 = g_n = g$). In both cases, parameters are optimized end to end.

\subsection{Intermediate loss}
\label{sec:loss}

To train the model, we use an intermediate loss \cite{teed2020raft} instead of the standard $L_1$ loss on the final displacement field $ \vdf_n$. This loss progressively weights the differences between the intermediate predicted displacement field $\vdf_i$ and the ground truth $\vdf_{gt}$, defined as:
\begin{equation}
\mathcal{L} = \sum_{i=1}^{n} \mathcal{L}_{i} = \sum_{i=1}^{n} \gamma^{n-i} |\vdf_{gt} - \vdf_i| 
\end{equation}
where $\gamma$ is an attenuation factor, prioritizing later predictions. This loss is effective in our context as it enforces predicting a good estimate at each iteration instead of shifting at each iterations and compensating with the last iteration model.

\subsection{A posteriori regularization}
\label{sec:regularization}

To recover the original displacement field $\vdf_{\text{earth}}$ from a noisy estimate $\vdf_{\text{noisy}}$, we frame the problem as an optimization task aiming to denoise the field while preserving spatial smoothness:
\begin{equation}
 \vdf_{\text{denoised}} = \arg \min_{\vdf \in \real^{H \times W \times 2}} \left(\norm{\vdf - \vdf_{\text{noisy}}}_2^2 + \lambda \psi(\vdf) \right)
\end{equation}

The first term enforces data fidelity, penalizing deviations from the noisy data, while the second term, $\psi(\vdf)$, encourages smoothness, reducing high-frequency noise. The parameter $\lambda$ balances data fidelity and smoothness.

Instead of optimizing the displacement field $\vdf=(\vu, \vv)$ together, we optimize each direction, $\vu$ and $\vv$, independently. Focusing on $\vu$, for the remainder of this section, we now solve the following optimization problem:
\begin{equation} \label{eq:u_denoised}
\vu_{\text{denoised}} = \arg \min_{\vu} \left(\norm{\vu - \vu_{\text{noisy}}}_2^2 + \lambda \psi(\vu) \right)
\end{equation}

To gain insights, we first focus on 1D signals $\vu \in \real^{H}$ and $\vu_{\text{noisy}} \in \real^{H}$ and analyze different penalty functions $\psi$. This approach is extended to 2D images using alternating projection algorithms like Dykstra's, which solves the 1D optimization problem across rows and columns.

\myparagraph{L2 and L1 (Total Variation) penalty function}
The L2 gradient penalty, $\psi(\vu) = \norm{\nabla \vu}_2^2$, is convex and smooth. However, its quadratic nature tends to over-smooth solutions, especially near sharp variations. A common alternative is the total variation (TV) penalty, $\psi(\vu) = |\nabla \vu |$, which promotes piecewise constant solutions, preserving edges while smoothing within regions but potentially blurring discontinuities due to treating all variations equally.

\myparagraph{Log total variation (LTV) penalty function}
We propose a log total variation penalty $\psi(\vu) = \log(|\nabla \vu| + \epsilon)$. Due to the $log$ concavity, exact optimization is computationally expensive. However, the concavity enables an upper-bound approximation using the linearization of $\psi$ at a point $\beta$:
\begin{equation}
\phi_{\beta}(\vu) = \frac{|\nabla \vu|}{|\beta| + \epsilon}
\end{equation}

This allows iterative minimization via the reweighted-L1 technique~\cite{bach2012optimization}. Each iteration solves a weighted-L1 problem
\begin{equation}
 \vu_{\text{new}} \leftarrow \arg \min_{\vu \in \real^{H}} \left( \|\vu - \vu_{\text{noisy}}\|_2^2 + \lambda \sum_{i=1}^{H} \frac{|\nabla \vu[i]|}{|\nabla \vu^{\text{old}}[i]| + \epsilon} \right)
\end{equation}
where $\vu^{\text{old}}$ is the previous estimate.

\section{Experiments}
\label{sec:experiments}

\label{sec:details}

\myparagraph{Training details}
We train our models for $n=3$ iterations. Training details, optimizer, and data augmentations are in the supplementary material. Our \ltv regularization follows ProxTV\footnote{\url{https://github.com/albarji/proxTV}} with $\lambda=0.001$ and 3 iterations.

\myparagraph{Evaluation metrics}
We use the standard average point error metric (EPE) \cite{sun2014quantitative} to quantify photometric accuracy and introduce a smoothness metric ($\norm{\nabla \vf}_2^2$) to assess noise levels in near-fault and non-fault regions, ensuring adherence to physical constraints.

\myparagraph{Displacement ranges}
To assess model adaptability across displacement ranges, and thus, varying satellite image resolutions, we consider three ranges: very small (under a pixel), small (1–5 pixels), and medium (5–15 pixels).

\subsection{Datasets}

\myparagraph{Training on a semi-synthetic dataset}
Since ground truth data is unavailable, we train on the semi-synthetic FaultDeform dataset \cite{Montagnon24} \footnote{\url{https://doi.org/10.57745/AHEOVO}}, designed by geoscience experts. It includes realistic displacement fields from strike-slip fault discontinuities, generated using a 3D simulator based on historical earthquake data. The dataset features time intervals between pre- and post-seismic images ranging from weeks to months, capturing real-world variability.

\myparagraph{Real examples from various satellite imagery}
We evaluate on three datasets of pre- and post-seismic images from the Ridgecrest earthquake region, including Landsat-8 (15m resolution) and SPOT 6 (2m resolution) imagery covering a 45x45 km² area. It contains 9 pairs of Landsat images and 195 pairs of SPOT images, along with data from other satellites (Sentinel-2, Leica ADS80, Pleiades) with resolutions ranging from 15m to 50cm. This diverse dataset tests the model's robustness to varying spatial resolutions and temporal gaps between acquisitions, demonstrating its adaptability to different imaging conditions. Additional details are in the supplementary material.

\begin{table}[t]
    \resizebox{0.5\textwidth}{!}{%
    \begin{tblr}{
  colspec = {l | ccc |ccc},
  cell{1}{1} = {r=2}{c}, 
  cell{1}{2} = {c=3}{c}, 
  cell{1}{5} = {c=3}{c}, 
  cell{3}{1} = {c=7}{l}, 
  cell{7}{1} = {c=7}{l}, 
  cell{12}{1} = {c=7}{l}, 
  rowsep=1pt,  
  colsep=2pt
}
\toprule
Method &  \textbf{EPE $\downarrow$} & & & \textbf{Smoothness (non-fault) $\downarrow$ }\\
& very small & small & medium & very small & small & medium \\
\midrule
~~\textit{Baselines \matching} \\
\cosicorr~\cite{LeprinceA07} & 0.261 & 0.300 & 0.425 & 0.116 & 0.146 & 0.254 \\
\micmac~\cite{Rosu15} & 0.293 & 0.253 & - & 0.166 & 0.203 & - \\
\cnndis~\cite{montagnon2024sub} & 0.328 & 0.527 & - & - & - & -\\ 
\midrule
~~\textit{Baselines encoder-decoder} \vspace{2pt} \\
\gfn$^\dagger$ \cite{Montagnon24}&  0.136 & 0.239 & 0.293& 0.037&0.080&0.185\\
FlowNetS \cite{dosovitskiy2015flownet}& 0.150& 0.261& 0.352& 0.033&0.076&0.192 \\
FlowNetC \cite{dosovitskiy2015flownet} & 0.258&0.390&0.638&0.042&0.090&0.238 \\
RAFT \cite{teed2020raft} & 0.137&0.225&0.279&0.016&0.050&0.143 \\
\midrule
~~\textit{Our approach} \vspace{2pt} \\
\textbf{(Ours)}-shared weigths & \textbf{0.133}&0.221&0.250&\textbf{0.012}&\textbf{0.040}&\textbf{0.137}\\
\textbf{(Ours)}-sep. weigths & 0.136&\textbf{0.219}&\textbf{0.240}&0.022&0.051&0.151\\
\bottomrule
\end{tblr}

    }
    \caption{\textbf{Comparison with State-of-the-Art methods} for the dense ground displacement estimation task on the semi-synthetic FaultDeform dataset~\cite{Montagnon24}, using the EPE metric across various displacement ranges, from very small to large displacements. \gfn was originally trained using displacement range of [0.1, 50] and with the EPE loss. We report our results for displacement range of [0.1, 15]. $^\dagger$ denotes \gfn trained with the $\mathcal{L}_1$ loss.}
    \label{tab:sota}
    \vspace{-5pt}
\end{table}

\begin{figure}[ht]
\setlength\tabcolsep{2pt}
\begin{tblr}{
  colspec = {X[c,0.1]X[c,h]X[c,h]X[c,h]X[c,h]},
  stretch = 0,
  rowsep = 2pt,
  hlines = {red5, 0pt},
  vlines = {red5, 0pt},
}
&\textbf{IR } & \textbf{IR } &  \textbf{{ \raft }} & \textbf{{\raft }} \\
& \small\textbf{corr.} & \small \textbf{no corr.} & \small \textbf{{ corr. }} &\small \textbf{{no corr.}} \\
(a) & \includegraphics[width=1.9cm]{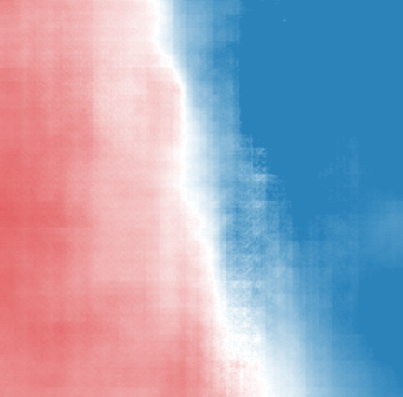} &  \includegraphics[width=1.9cm]{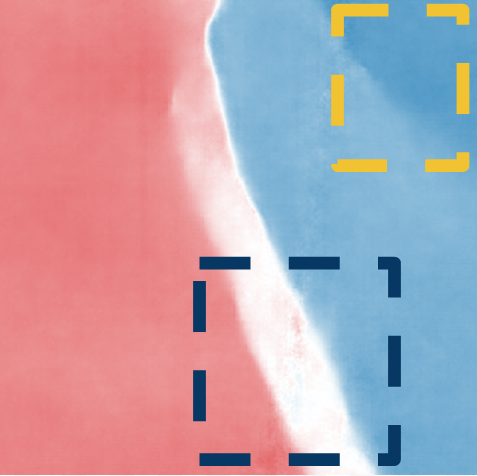}&
\includegraphics[width=1.9cm]{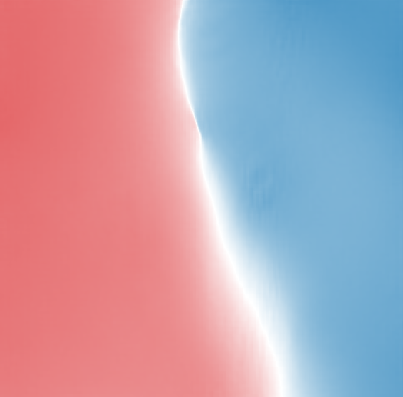}&
\includegraphics[width=1.9cm]{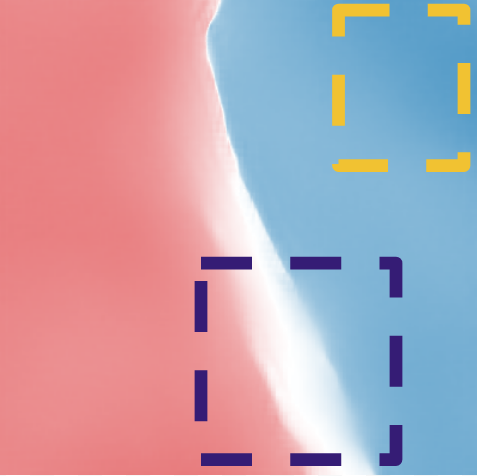} \\

(b) & \includegraphics[width=1.9cm]{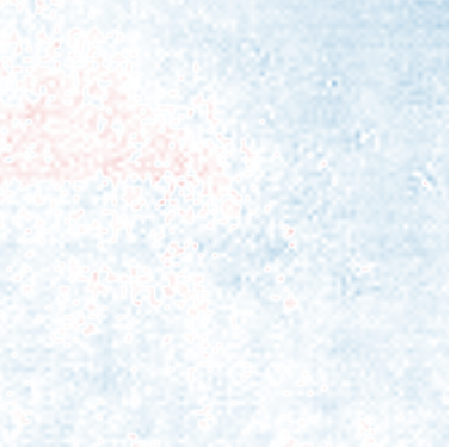} &  \includegraphics[width=1.9cm]{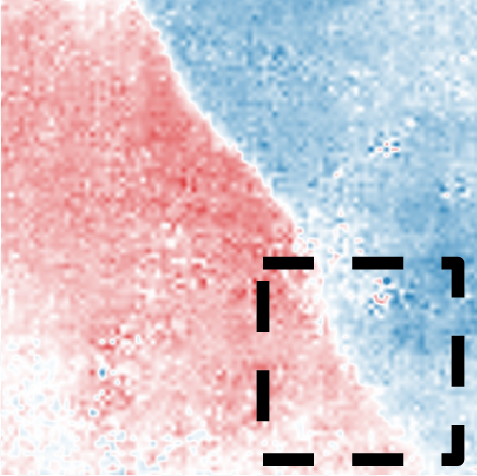}&
\includegraphics[width=1.9cm]{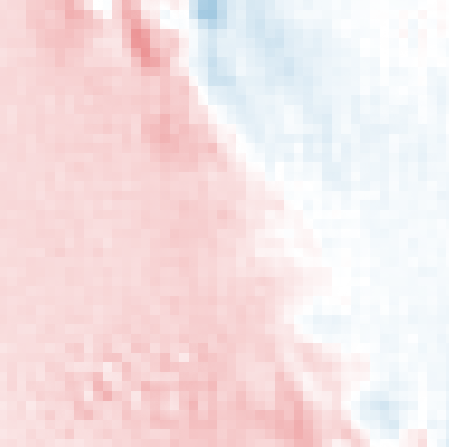}&
\includegraphics[width=1.9cm]{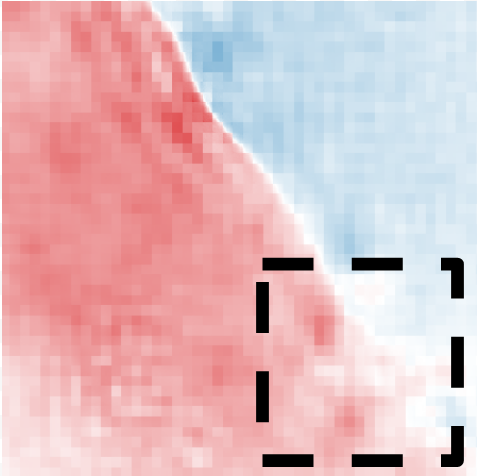} \\

\end{tblr}

\vspace{-10pt} 
\caption{\textbf{Correlation-dependent backbones lag behind in predicting small to very small displacement fields}, unlike their correlation-independent counterparts both for high resolution (a) and medium resolution (b) images. This is illustrated for two iterative algorithms, one with explicit warping (IR) and one with recurrent units (\raft).}
\label{fig:correlation}
\vspace{-5pt} 
\end{figure}

\begin{table}[t]
    \resizebox{0.5\textwidth}{!}{%
    \begin{tblr}{
  colspec = {ccccc|ccc|ccc},
  cell{1}{1} = {r=2}{c}, 
  cell{1}{2} = {r=2}{c}, 
  cell{1}{3} = {r=2}{c}, 
  cell{1}{4} = {r=2}{c}, 
  cell{1}{5} = {r=2}{c}, 
  cell{1}{6} = {c=3}{c}, 
  cell{1}{9} = {c=3}{c}, 
  cell{3}{1} = {c=6}{l}, 
  cell{6}{1} = {c=6}{l}, 
  cell{9}{1} = {c=6}{l}, 
  cell{12}{1} = {c=6}{l}, 
  rowsep=1pt,  
  colsep=2pt
}
\toprule
\cl & \ir & \ws & \il & \ltvm  & \textbf{EPE $\downarrow$} & & & \textbf{Smoothness (non-fault)$\downarrow$}  \\
& & & & & very small & small & medium & very small & small & medium\\
\midrule
~~\textit{Impact of \cl} \vspace{2pt} \\
\ding{55} & \ding{55} & \ding{55} & \ding{55} & \ding{55}  & 0.258&0.390&0.638&0.042&0.090&0.238\\
\checkmark & \ding{55} & \ding{55} & \ding{55} & \ding{55} &  0.136 & 0.239 & 0.293& 0.037&0.080&0.185\\
\midrule
~~\textit{Impact of \ir} \vspace{2pt} \\
\checkmark & \checkmark & \ding{55} & \ding{55} & \ding{55} & 0.135&0.241&0.285&0.087&0.108&0.180 \\
\checkmark & \checkmark & \checkmark & \ding{55} & \ding{55} & 0.138&0.230&0.257&0.064&0.095&0.178 \\
\midrule
~~\textit{Impact of \il} \vspace{2pt} \\
\checkmark & \checkmark & \ding{55} & \checkmark & \ding{55} & 0.134&0.222&0.252&0.073&0.095&0.182\\
\checkmark & \checkmark & \checkmark & \checkmark & \ding{55} & 0.137&0.221&0.242&0.063&0.089&0.175\\
\midrule
~~\textit{Impact of \ltvm} \vspace{2pt} \\
\checkmark& \checkmark & \ding{55} & \checkmark & \checkmark &  0.133&0.221&0.250&0.012&0.040&0.137\\
\checkmark &  \checkmark & \checkmark & \checkmark & \checkmark & 0.136&0.219&0.240&0.022&0.051&0.151\\
\bottomrule
\end{tblr}

    }
    \vspace{-5pt} 
    \caption{\textbf{Ablation for the different components of the models} described in Table \ref{tab:method_components}.}
    \label{tab:ablation_method}
    \vspace{-5pt} 
\end{table}

\begin{table}[ht]
    \centering
    \resizebox{0.5\textwidth}{!}{
    \begin{tblr}{
  colspec = {c | ccc |ccc},
  cell{1}{2} = {c=3}{c}, 
  cell{1}{5} = {c=3}{c}, 
  rowsep=1pt,  
  colsep=2pt
}
\toprule
Maximum number &  \textbf{EPE $\downarrow$} & & & \textbf{Smoothness (non-fault) $\downarrow$ }\\
of iterations & very small & small & medium & very small & small & medium \\
\midrule
$n=1$ & 0.136&0.239 & 0.293 & 0.037&0.080&0.185\\
$n=2$ & 0.136& 0.220&0.248 & 0.058&0.089&0.182\\
$n=3$ & 0.137&0.221 & 0.242 & 0.063&0.089&0.175 \\
$n=4$ & 0.132&0.216&0.239 &0.072&0.095&0.178\\
$n=5$ & 0.140 &0.235&0.262 &0.105&0.126&0.206 \\
\bottomrule
\end{tblr}
    }
    \vspace{-5pt} 
    \caption{\textbf{Impact of the maximum number of iterations $n$}} 
    \label{tab:ir_n}
\end{table}

\begin{table}[ht]
\vspace{-5pt} 
    \resizebox{0.5\textwidth}{!}{%
    \begin{tblr}{
  colspec = {l | ccc |ccc},
  cell{1}{2} = {c=6}{c}, 
  cell{2}{2} = {c=3}{c}, 
  cell{2}{5} = {c=3}{c}, 
  rowsep=1pt,  
}
\toprule
 & \textbf{Smoothness} \\
Param & near-fault $\uparrow$ & & & non-fault $\downarrow$ & & \\
& very small & small & medium & very small & small & medium \\
\midrule
Reference&0.580&3.427&11.489&0.005&0.029&0.095\\
No reg & 0.321&2.577&7.826&0.063&0.089&0.175\\
LTV reg & 0.289&2.601&7.947&0.022&0.051&0.151\\
\bottomrule
\end{tblr}
    }
    \vspace{-5pt}
    \caption{\textbf{Impact of the regularization both in the near-fault and non-fault areas}.}
    \label{tab:ablation_ltv_summary}
    \vspace{-12pt} 
\end{table}

\begin{figure*}[h]
    \centering
    \resizebox{0.95\textwidth}{!}{
    \includegraphics{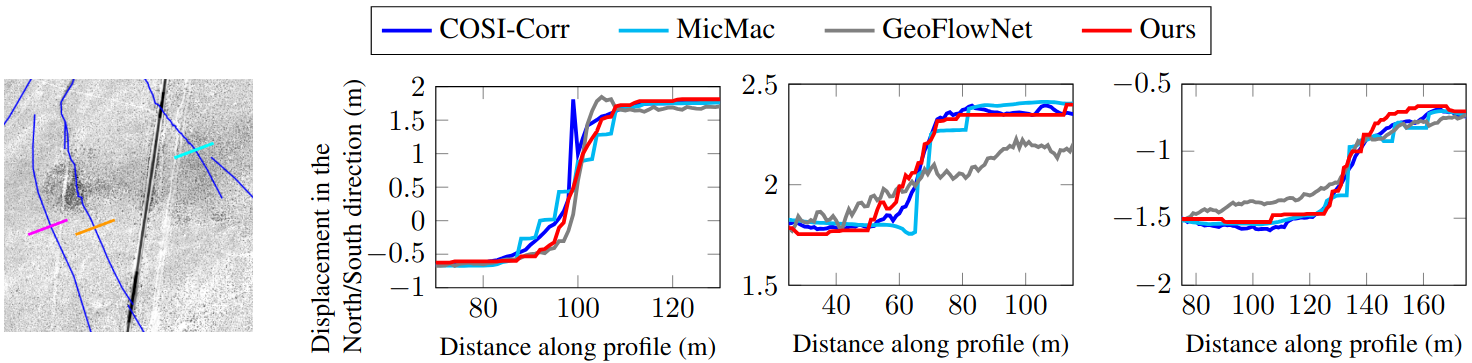}
    }
    \vspace{-10pt} 
    \caption{\textbf{Displacement profiles} comparing estimates from \cosicorr, \micmac, \gfn, and our proposed method, across the three faults depicted in orange, magenta and cyan on the left sub-figure.} 
    \label{fig:high_fault_analysis}
    \vspace{-10pt} 
\end{figure*}

\begin{figure}[t]
\setlength\tabcolsep{2pt}
\begin{tblr}{
  colspec = {X[c,0.1]X[c,h]X[c,h]X[c,h]X[c,h]},
  stretch = 0,
  rowsep = 2pt,
  hlines = {red5, 0pt},
  vlines = {red5, 0pt},
}
& \textbf{$I_1$}& \small\textbf{\shortstack{1. \raft}} & \small \textbf{\shortstack{2. \gfn}} &\small \textbf{\shortstack{3. Ours}} \\
(a) & \includegraphics[width=1.9cm]{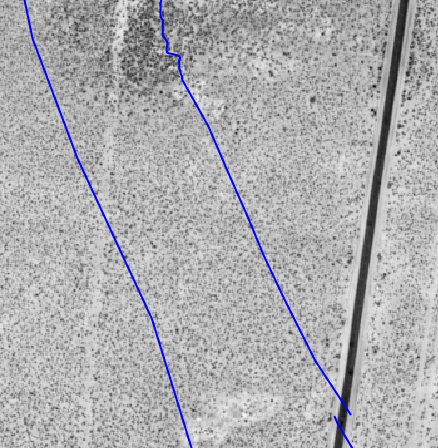} &  \includegraphics[width=1.9cm]{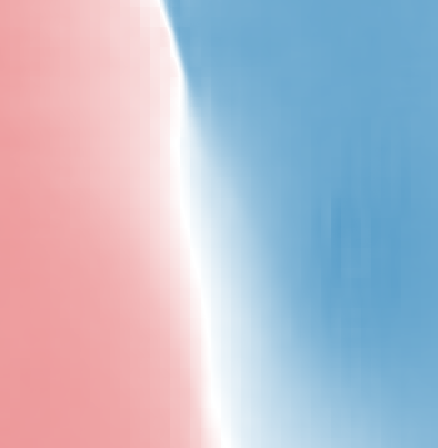}&
\includegraphics[width=1.9cm]{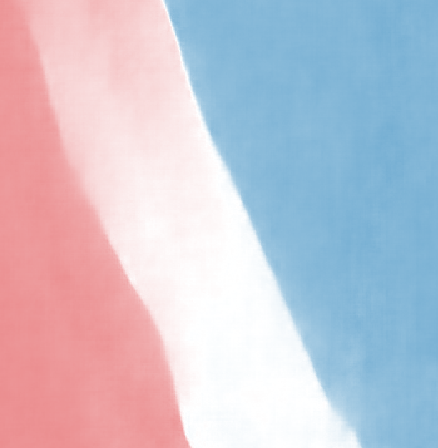}&
\includegraphics[width=1.9cm]{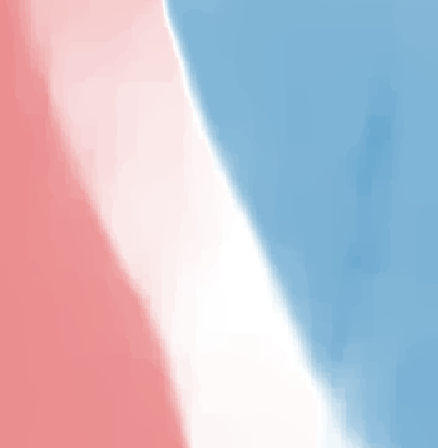} \\

(b) & \includegraphics[width=1.9cm]{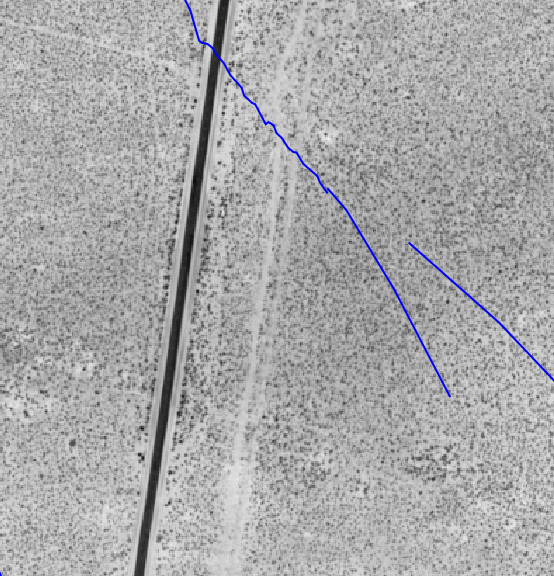} &  \includegraphics[width=1.9cm]{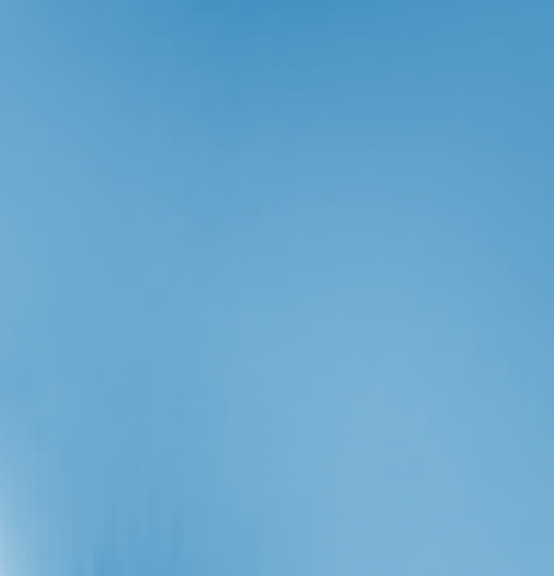}&
\includegraphics[width=1.9cm]{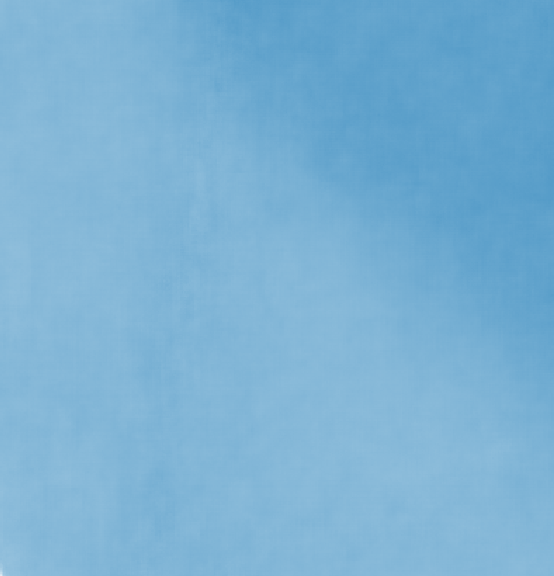}&
\includegraphics[width=1.9cm]{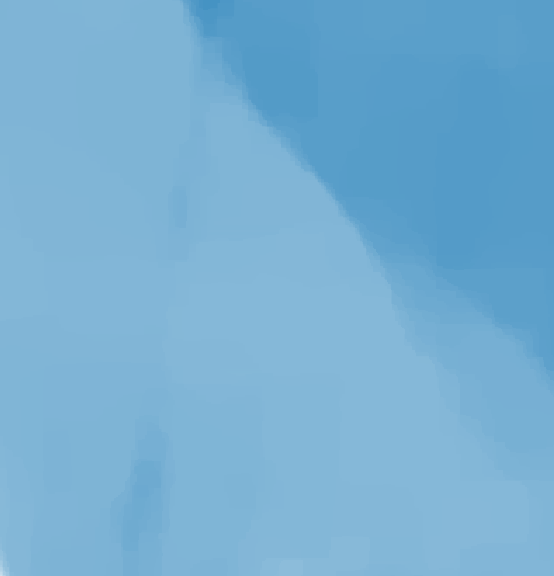} \\

(c) & \includegraphics[width=1.9cm]{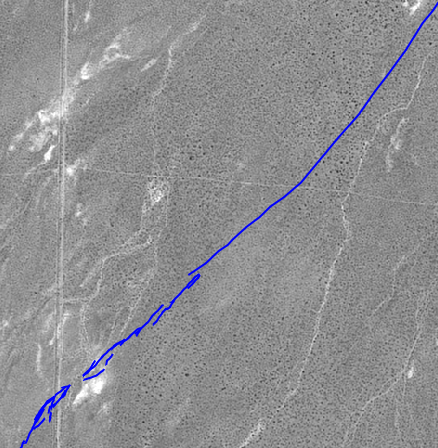} &  \includegraphics[width=1.9cm]{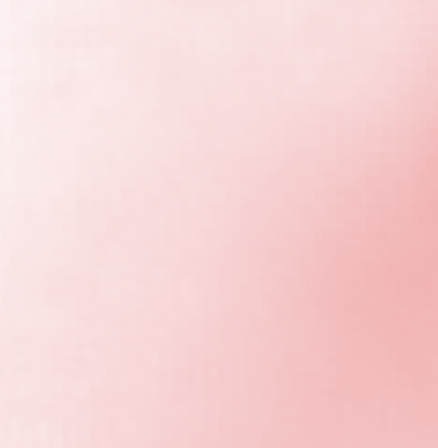}&
\includegraphics[width=1.9cm]{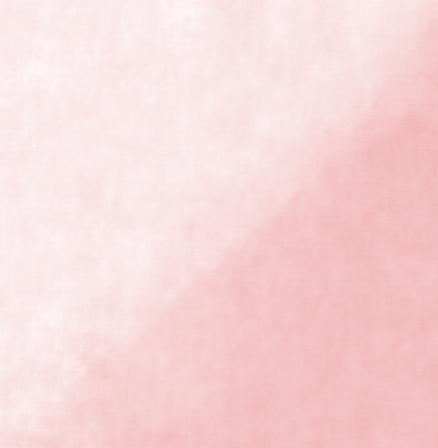}&
\includegraphics[width=1.9cm]{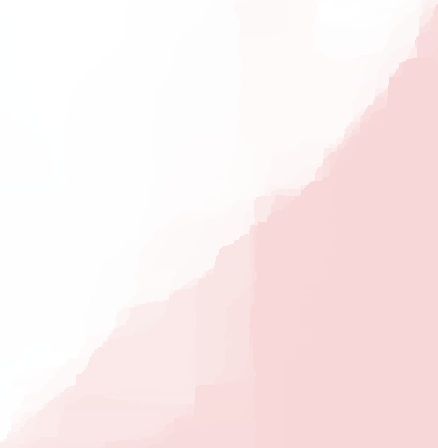} \\

\end{tblr}

\vspace{-10pt} 
\caption{\textbf{The need for sub-pixel precision:} achieved with \microflow (3), but unmet by correlation-dependent backbones (1) and non-iterative models (2).}
\label{fig:subpixel}
\vspace{-15pt} 
\end{figure}

\begin{figure}[ht]
    \centering
    \setlength\tabcolsep{2pt}
\begin{tblr}{
  colspec = {X[c,h]X[c,h]},
  stretch = 0,
  rowsep = 2pt,
  hlines = {red5, 0pt},
  vlines = {red5, 0pt},
  cell{3}{1}={c=2}{c}, 
  cell{5}{1}={c=2}{c}, 
  cell{7}{1}={c=2}{c}, 
  cell{9}{1}={c=2}{c}, 
}
\textbf{Without LTV} & \textbf{With LTV} \\

\begin{overpic}[width=4cm]{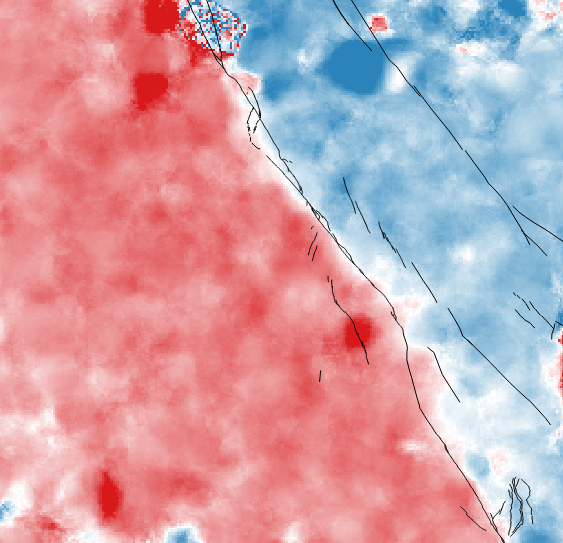} 
\put(5,5){\includegraphics[width=1cm]{fig/00_scales/scale_landsat.png}} 
\end{overpic}
& \includegraphics[width=4cm]{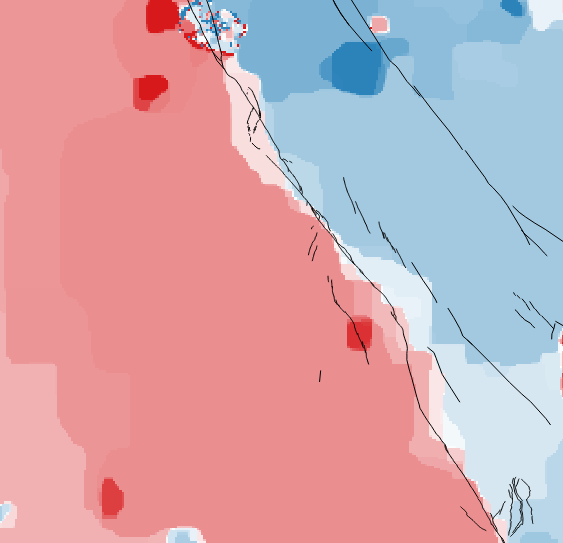}  \\
COSI-Corr \\
\includegraphics[width=4cm]{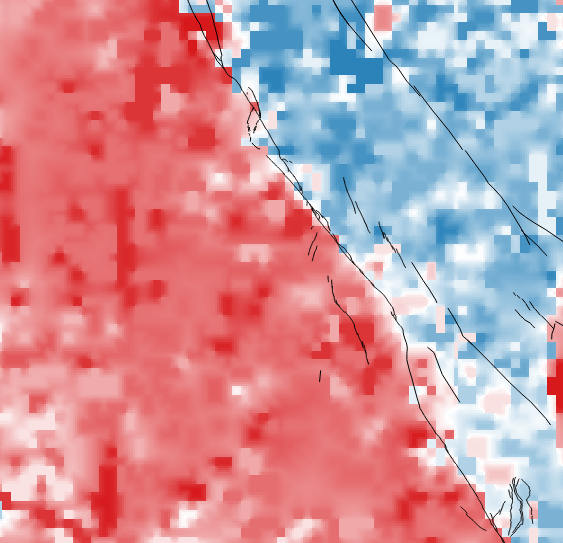}
&\includegraphics[width=4cm]{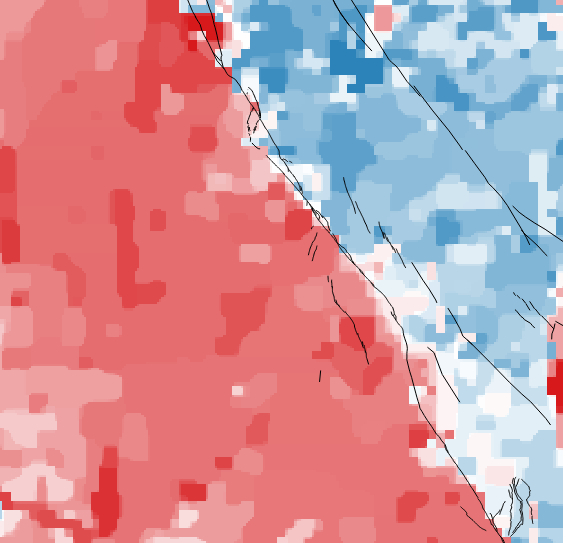}\\
\micmac \\
\includegraphics[width=4cm]{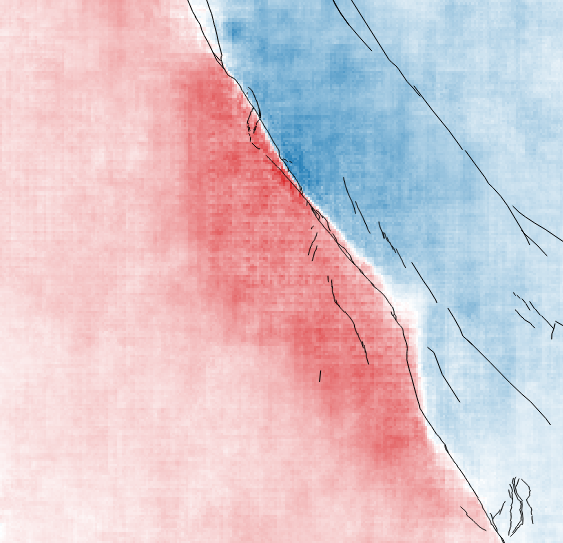}
&\includegraphics[width=4cm]{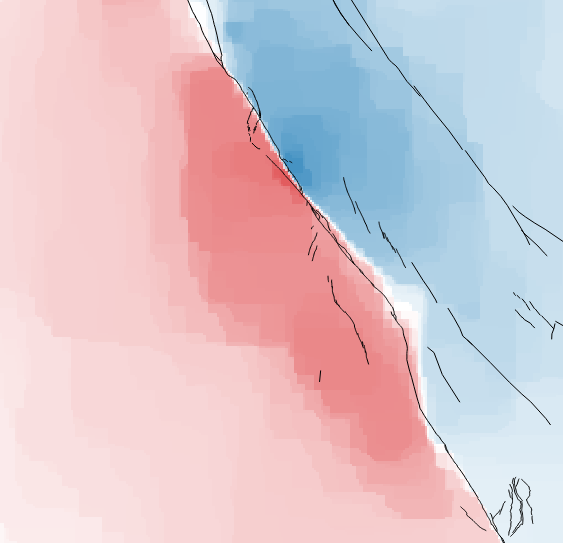}\\
\gfn \\
\includegraphics[width=4cm]{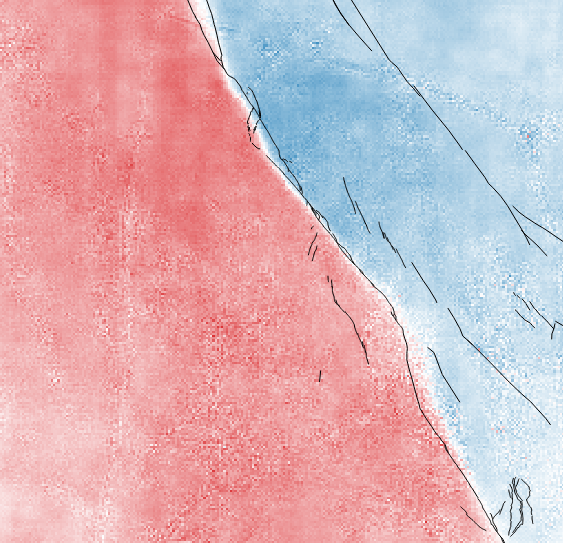}
&\includegraphics[width=4cm]{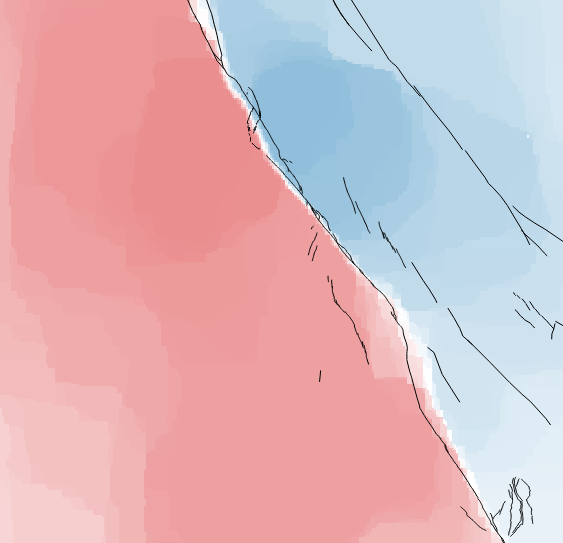}\\
Ours \\ 
\end{tblr}

    \caption{\textbf{Impact of LTV a-posteriori regularization on four models, \cosicorr, \micmac, \gfn and \microflow} for a 2820x2820m area acquired with \landsat. Expert annotations are in black, highlighting the precision of the fault location for the two data-driven approaches over traditional \matching ones.}
    \label{fig:zoom}
\end{figure}

\subsection{Comparison with state-of-the-art models}

With limited annotated data, quantitative evaluation is challenging, so best practices often rely on semi-synthetic benchmarks as proxies for real-world performance. Although imperfect, these benchmarks guide model design.

Table~\ref{tab:sota} presents a comparison on the test set of the semi-synthetic FaultDeform dataset. Our models were trained with different training parameters as for the original \gfn. For a fair comparison, we retrained \gfn with our setup, resulting in improved performance over their reported results. We compare the models using the standard EPE metric for the entire image and our new smoothness metric for regions away from the fault, which assesses the estimate's consistency with expected earth motion constrained by physics. 

Overall, deep learning models substantially outperform \matching methods on all metrics across all displacement ranges, except for FlowNet-C, which we discuss in section \ref{sec:ablation}. The combination of a correlation independent encoder-decoder network with iterative refinements, and our non-convex regularization greatly enhances accuracy for both small and large displacements, while maintaining stability for very small displacements.

\subsection{Ablation study}
\label{sec:ablation}

Table \ref{tab:ablation_method} summarizes the impact of each model's component.  

\myparagraph{Correlation dependent vs correlation independent networks (\cl)}
\label{sec:correlation}
The correlation layer is a key component of most recent video motion estimation models where medium to large displacements prevail. However, in small to very small displacement regimes, correlation dependent networks struggle to estimate accurate displacement fields when their correlation independent counterpart achieve better results as illustrated in Figure \ref{fig:correlation} and Table \ref{tab:ablation_method}. 

\myparagraph{Explicit warping vs recurrent units (\wpg)}
\label{sec:warping}
We compare the estimates with and without explicit warping during the iterative process in Figure \ref{fig:correlation} (columns 2 and 4). If iterative process with recurrent unit provides a decent estimate in terms of photometric accuracy, iterative process needs warping to recover sharp faults with precise location (black rectangle) and achieve sub-pixel precision (dark blue and yellow rectangles).

\myparagraph{Iterative refinements (\ir)}
Iterative refinements trained with the intermediate loss significantly enhance the photometric performance for small to medium displacements achieving a relative boost of 7$\%$ for small displacements and up to 17$\%$ for medium displacements. The model with separate weights outperforms the shared-weight model for medium displacements, while the shared-weight model is better for very small displacements. Both models perform equally in the small displacement range.

\myparagraph{Impact of the maximum number of iterations $n$}
Table~\ref{tab:ir_n} shows that increasing iterations improves the model until a limit where performance declines. While $n=4$ yields higher accuracy at the cost of smoothness, we choose $n=3$ as a trade-off between EPE and smoothness.

\myparagraph{Impact of the intermediate loss (\il)}
The intermediate loss aligns each iteration's estimate with ground deformation, enhancing the accuracy of the final estimate. It significantly improves the average point error with a slight gain in smoothness.

\myparagraph{Impact of the a-posteriori regularization (\ltvm)}
The non-convex penalty provides modest photometric gains but preserves data fidelity and reduces noise, as shown by the smoothness metric. Table \ref{tab:ablation_ltv_summary} highlights its impact in near-fault and non-fault regions, maintaining sharpness near faults while smoothing elsewhere. Ground truth smoothness serves as a reference.
Ablations over $\lambda$ and $k$ are detailed in the supplementary (Sec. A).

\subsection{Qualitative results on real examples}
\label{sec:qualitative}

\myparagraph{Achieving sub-pixel precision for small to large displacement fields}
While classical models like \cosicorr and \micmac are designed specifically for the task to achieve sub-pixel precision, previous data-driven models trained solely on the semi-synthetic dataset fall short, as shown in Figure \ref{fig:subpixel}. 
The lack of sub-pixel precision leads to errors such as merging two faults into one and mislocalizing them (\raft, example a) or failing to accurately detect the main fault and secondary fault locations (\raft and \gfn, example b). In contrast, our proposed model successfully achieves sub-pixel precision, providing a tool to detect not only the main fault but also secondary ones.

\myparagraph{In the near-fault regions, the sharpness of the estimated displacement fields is preserved}
To assess how well our model preserves sharpness in estimated displacement fields, we compare displacement profiles for three faults (Figure \ref{fig:high_fault_analysis}). The geophysics baselines, \cosicorr and \micmac, align in displacement levels and fault boundaries, but \cosicorr exhibits aliasing, and \micmac over-smooths. \gfn captures the primary fault’s sharpness but misestimates the other two, with one fault missing entirely. Our model overcomes these issues, achieving accurate displacement levels and preserving sharp profiles for all three faults.

\myparagraph{In the non fault regions, the noise generated by the temporal change is largely reduced}
We analyze regions farther from faults using medium-resolution (\landsat) data to assess estimates amid temporal changes. Ground deformation should be smooth in these areas due to physical constraints.
We compare our model's estimates with those from \cosicorr, \micmac, \raft, and \gfn in Figure \ref{fig:teaser} (c) and (d) and the Figure C in the appendix. Classical models generate high correlation noise, even with regularization (\micmac), while \raft produces pixelated estimates due to its correlation-dependent backbone. In contrast, data-driven methods with correlation-independent backbones yield smoother estimates, effectively enhanced by regularization. Using the smoothness metric, we quantitatively validate these qualitative observations across 10 images, obtaining smoothness values of 0.125, 0.082 and 0.036, for \cosicorr, \micmac and \microflow respectively.

\myparagraph{The combination of data-driven models and non convex regularization works effectively and lead to improved fault location}
We compare displacement estimates from our proposed method with those from \cosicorr, \micmac, and \gfn in a region near the fault, as shown in the left column of Figure~\ref{fig:zoom}. The traditional methods, \cosicorr and \micmac, exhibit significantly higher noise compared to the deep learning models, supporting the idea that incorporating data priors helps mitigate the impact of strong temporal perturbations. When applying our non-convex regularization to all four estimates (Figure~\ref{fig:zoom}, right column), noise levels decrease across all models. However, the two data-driven approaches, which integrate prior knowledge of seismic fault behavior, achieve the most precise and physically consistent fault localization. Moreover, we can see that GeoFlowNet is fading far from the fault, which is not coherent with the non-data-driven methods. On the other hand, our \microflow keeps a non-zero deformation even far from the fault.

\section{Concluding Remarks}
\label{sec:conclusion}

In this work, we introduce \microflow, a novel optical flow model addressing the unique challenges posed by ground deformation estimation, a critical task for the geoscience community. Unlike standard video motion optical flow task with medium to large displacements, this task involves small to very small displacements, requiring sub-pixel precision for accurate estimates and robustness to high temporal variations from the image acquisition process. 

\microflownospace's correlation-independent backbone, enhanced with iterative refinements and explicit warping, achieves subpixel precision while preserving fault-line sharpness through non-convex regularization.

With limited manually annotated data, quantitatively assessing model performance on real data remains challenging. We first rely on semi-synthetic benchmarks, though imperfect, but offering valuable insights and allowing quantitative comparisons. Our proposed method \microflow shows significant improvement over two widely used methods in the geoscience community, \cosicorr and \micmac. 

We also perform a comprehensive qualitative evaluation using real-world pre- and post-seismic images from the Ridgecrest earthquake region, captured by medium- and high-resolution sensors. Our results show that \microflow achieves photometric accuracy across small and very small displacement ranges—a first for a deep learning-based optical flow model.

\section{Acknowledgement}
The authors would like to sincerely thank Tristan Montagnon for his helpful discussions.
This project was supported by ANR 3IA MIAI@Grenoble Alpes (ANR-19-P3IA-0003) and by ERC grant number 101087696 (APHELEIA project).
This work was granted access to the HPC resources of IDRIS under the allocation [AD011015113, AD010115035] made by GENCI.
{
    \small
    \bibliographystyle{ieeenat_fullname}
    \bibliography{main}
}


\renewcommand{\thetable}{\Alph{table}}
\renewcommand{\thefigure}{\Alph{figure}}
\clearpage
\setcounter{page}{1}
\appendix
\maketitlesupplementary

\section{Additional model analysis and ablations}
\label{sec:sup_ablations}

\paragraph{What does each iteration bring in the (\ir) process?}
Table \ref{tab:ir_step} illustrates the gains from each iteration for a fixed value of $n=3$. The second iteration notably improves accuracy across all displacement scales, while the third iteration provides marginal benefits. For very small displacements, however, iterative refinements introduce slight degradation, adding noise to the estimates—a common effect seen in iterative methods. In such cases, the best model is the one after the first iteration, which will be taken for the low-resolution satellites.

\begin{table}[h]
    \centering
    \resizebox{0.3\textwidth}{!}{
    \begin{tblr}{
  colspec = {c |c |ccc},
  rowsep = 1pt,
  row{1} = {font=\bfseries},
  cell{1}{3} = {c=3}{c}, 
  cell{1}{1} = {r=2}{c}, 
  cell{1}{2} = {r=2}{c}, 
}
\toprule
$i$ & \ws & \textbf{EPE $\downarrow$} \\
    &     & very small & small & medium \\
\midrule
$i=1$ & \ding{55} & 0.134 & 0.239 & 0.319 \\
$i=2$ & \ding{55} & 0.142 & 0.226 & 0.260 \\
$i=3$ & \ding{55} & 0.150 & 0.222 & 0.252 \\
\midrule
$i=1$ & \checkmark & 0.137 & 0.237 & 0.298 \\
$i=2$ & \checkmark & 0.138 & 0.222 & 0.246 \\
$i=3$ & \checkmark & 0.138 & 0.221 & 0.242 \\
\bottomrule
\end{tblr}
    }
    \vspace{-5pt} 
    \caption{\textbf{Impact of the iteration parameter $i$} for a fixed value of $n=3$.}
    \label{tab:ir_step}
\end{table}

\begin{figure}[h]
  \centering 
  \vspace{-10pt} 
  \setlength\tabcolsep{2pt}
\begin{tblr}{
  colspec = {X[c,h]X[c,h]},
  stretch = 0,
  rowsep = 2pt,
  hlines = {red5, 0pt},
  vlines = {red5, 0pt},
}
\includegraphics[width=3.5cm]{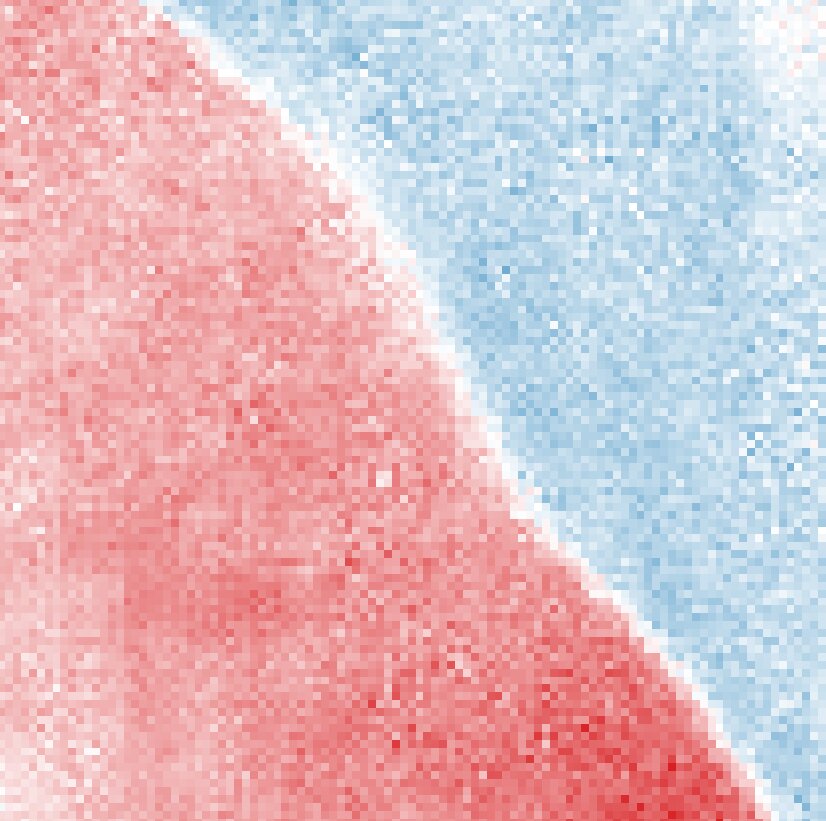} & 
\includegraphics[width=3.5cm]{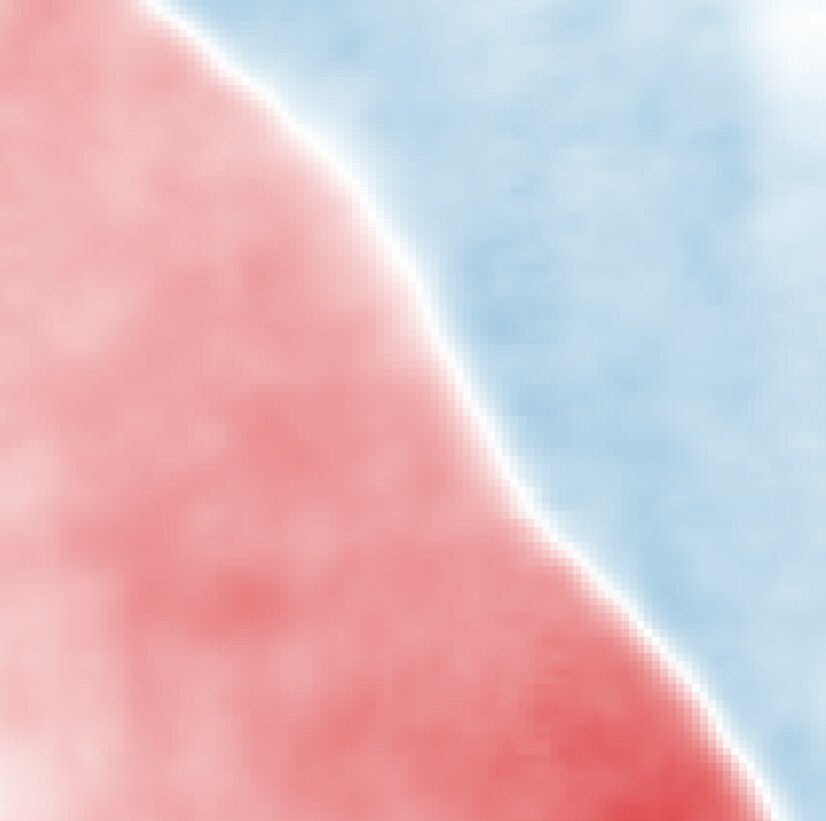} \\
(a) No Regularization & (b) L2 gradients \\
\includegraphics[width=3.5cm]{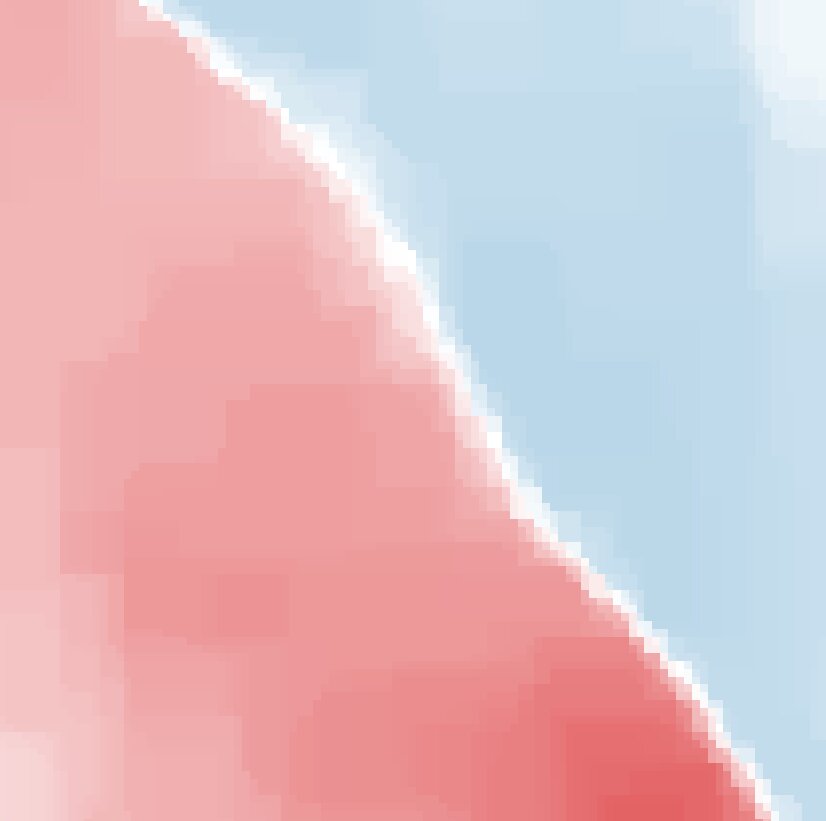} & 
\includegraphics[width=3.5cm]{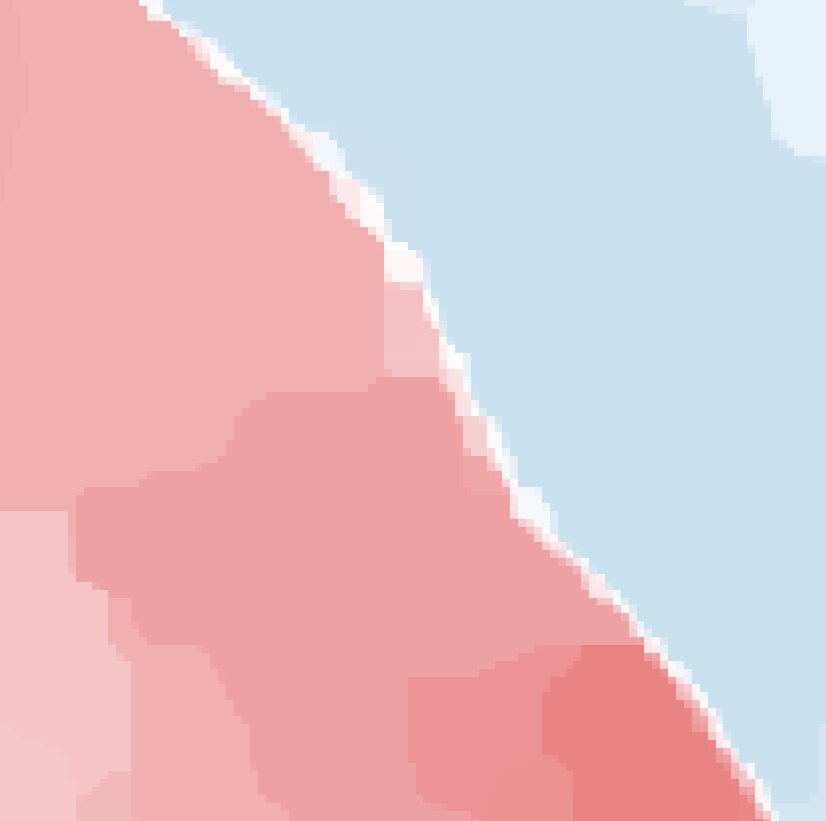} \\
(c) TV  & (d) LTV \\
\end{tblr}

  \vspace{-10pt} 
  \caption{\textbf{Comparison of the denoised image produced with different penalty functions during the a-posteriori regularization}, with $k=3$, and $\lambda=0.001$ for LTV, $\lambda=0.01$ for TV, $\lambda=0.1$ for L2 gradients.}
  \label{fig:penalty_functions}
\end{figure}

\begin{table}[h]
\vspace{-5pt} 
    \resizebox{0.5\textwidth}{!}{%
    \begin{tblr}{
  colspec = {l | ccc |ccc},
  cell{1}{2} = {c=6}{c}, 
  cell{2}{2} = {c=3}{c}, 
  cell{2}{5} = {c=3}{c}, 
  cell{4}{1} = {c=7}{l}, 
  cell{6}{1} = {c=7}{l}, 
  cell{9}{1} = {c=7}{l}, 
  cell{13}{1} = {c=7}{l}, 
  rowsep=1pt,  
}
\toprule
Param & \textbf{Smoothness} \\
& near-fault $\uparrow$ & & & non-fault $\downarrow$ & & \\
& very small & small & medium & very small & small & medium \\
\midrule
~~\textit{Groundtruth smoothness} \\
Reference & 0.580 & 3.427 & 11.489 & 0.005 & 0.029 & 0.095 \\
\midrule
~~\textit{\ltvm: Impact of the regularization} \\
No reg & 0.321 & 2.577 & 7.826 & 0.063 & 0.089 & 0.175 \\
LTV reg & 0.289 & 2.601 & 7.947 & 0.022 & 0.051 & 0.151 \\
\midrule
~~\textit{\ltvm: Number of iterations $k$} \\
$k=1$ & 0.297 & 2.585 & 7.920 & 0.027 & 0.057 & 0.156 \\
$k=2$ & 0.292 & 2.597 & 7.953 & 0.024 & 0.053 & 0.152 \\
$k=3$ & 0.289 & 2.601 & 7.947 & 0.022 & 0.051 & 0.151 \\
\midrule
~~\textit{\ltvm: $\lambda$} \\
$\lambda=0.0001$ & 0.305 & 2.573 & 7.861 & 0.033 & 0.062 & 0.156 \\
$\lambda=0.001$ & 0.289 & 2.601 & 7.947 & 0.022 & 0.051 & 0.151 \\
$\lambda=0.01$ & 0.270 & 2.471 & 7.476 & 0.012 & 0.037 & 0.151 \\
\bottomrule
\end{tblr}

    }
    \vspace{-10pt} 
    \caption{Quantitative ablations on the $k$ and $\lambda$ parameter for the \ltvm.}
    \label{tab:ablation_ltv_smoothness}
    \vspace{-12pt} 
\end{table}

\begin{figure*}[h]
  \centering 
  \vspace{-10pt} 
  \resizebox{0.9\textwidth}{!}{%

\setlength\tabcolsep{2pt}
\begin{tblr}{
  colspec = {X[c,h]X[c,h]X[0.2,h]X[c,h]X[c,h]X[c,h]},
  stretch = 0,
  rowsep = 2pt,
  hlines = {red5, 0pt},
  vlines = {red5, 0pt},
  cell{1}{1} = {c=2}{c}, 
}
\textbf{Image pair} && &\textbf{\smash{MicMac}}&\textbf{\smash{ \cosicorr}}  & \textbf{ Ours}\\
\includegraphics[width=2.6cm]{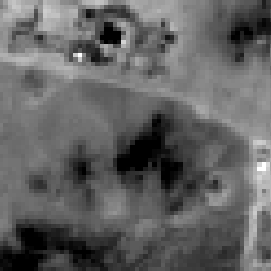} &  \includegraphics[width=2.6cm]{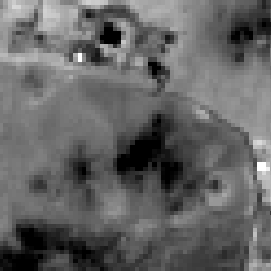}& &
\includegraphics[width=2.6cm]{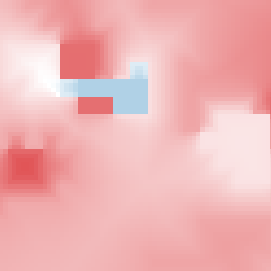}& 
\includegraphics[width=2.6cm]{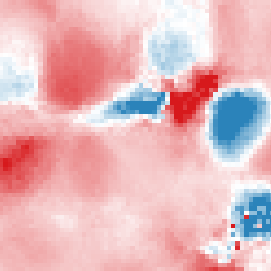}&
\includegraphics[width=2.6cm]{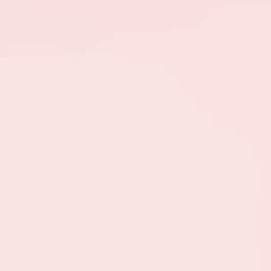} 
\\
\includegraphics[width=2.6cm]{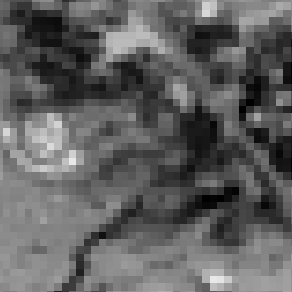} &  \includegraphics[width=2.6cm]{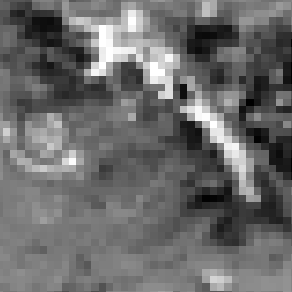}& &
\includegraphics[width=2.6cm]{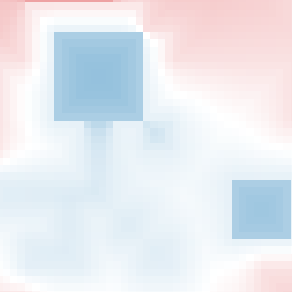}& 
\includegraphics[width=2.6cm]{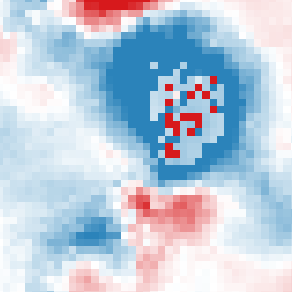}&
\includegraphics[width=2.6cm]{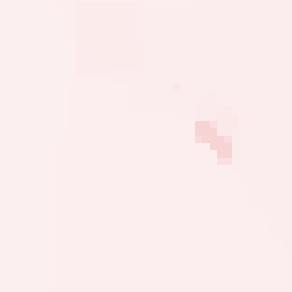} 
\\
 \includegraphics[width=2.6cm]{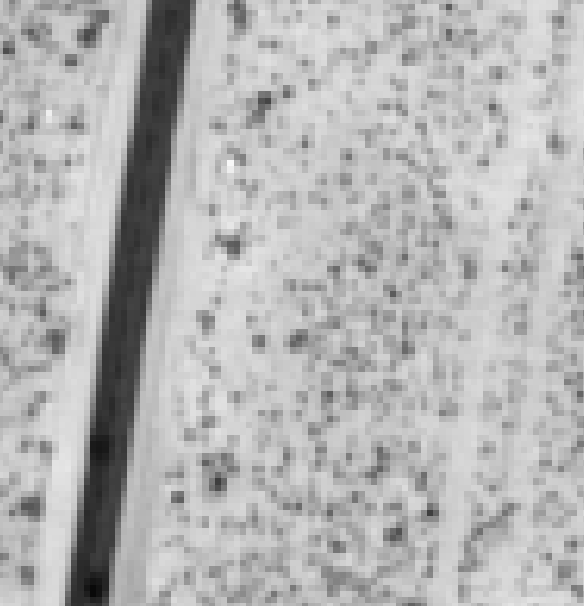} &  \includegraphics[width=2.6cm]{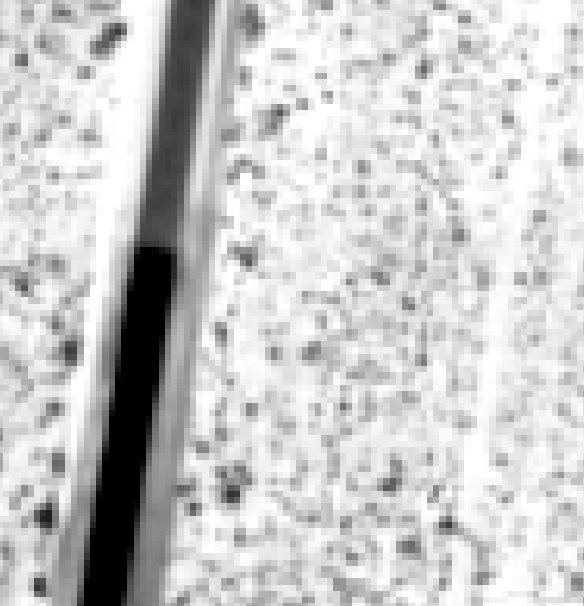}& &
\includegraphics[width=2.6cm]{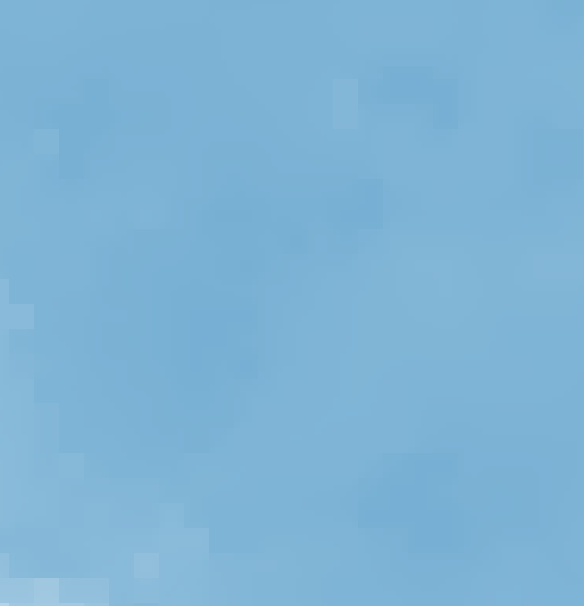}& 
\includegraphics[width=2.6cm]{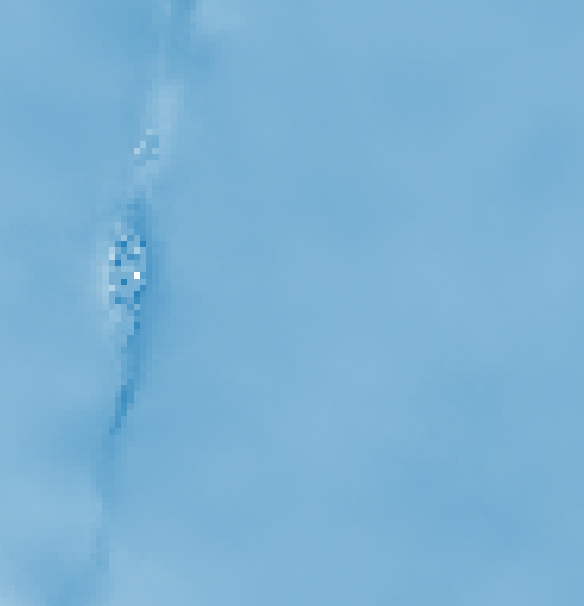}&
\includegraphics[width=2.6cm]{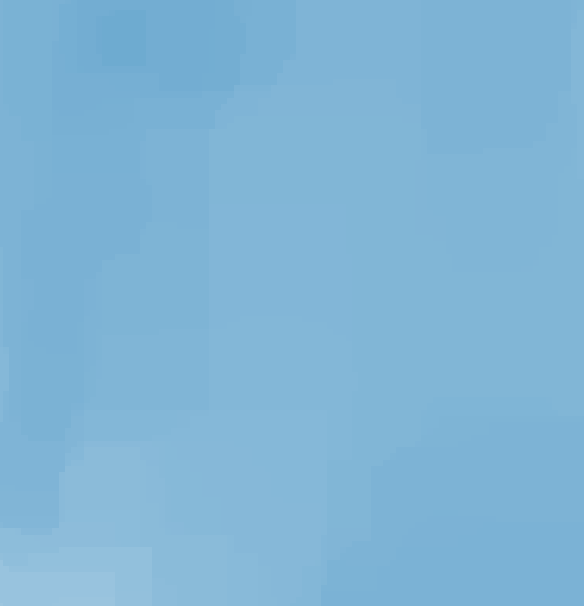} \\

\includegraphics[width=2.6cm]{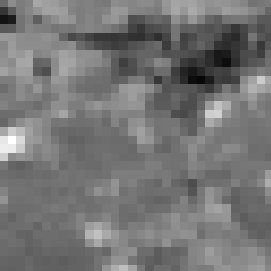} &  \includegraphics[width=2.6cm]{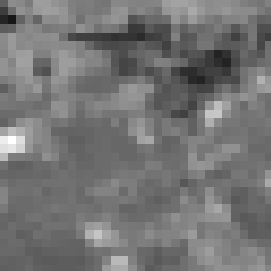}& &
\includegraphics[width=2.6cm]{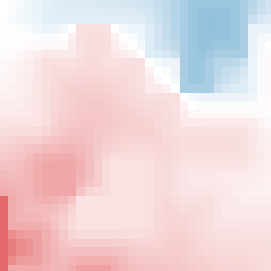}& 
\includegraphics[width=2.6cm]{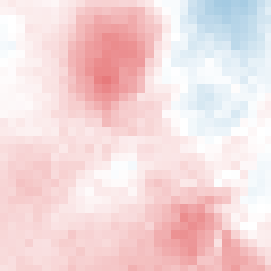}&
\includegraphics[width=2.6cm]{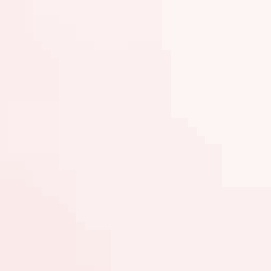} 
\\
 \includegraphics[width=2.6cm]{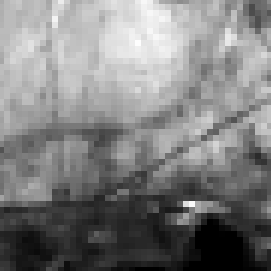} &  \includegraphics[width=2.6cm]{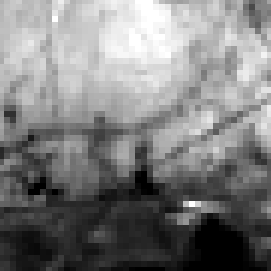}& &
\includegraphics[width=2.6cm]{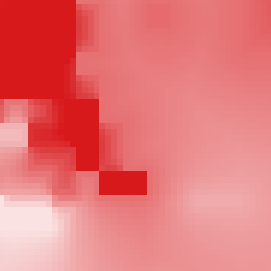}& 
\includegraphics[width=2.6cm]{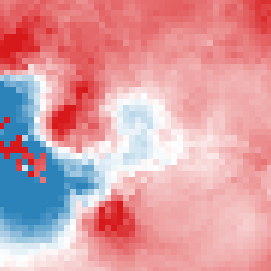}&
\includegraphics[width=2.6cm]{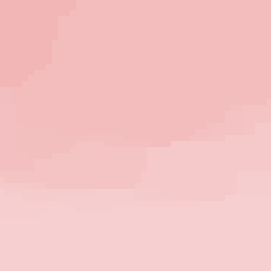} 
\\
\end{tblr}

  }
  \vspace{-10pt} 
  \caption{\textbf{Additional examples of vegetation and anthropogenic change}, with the corresponding \micmac, \cosicorr and \microflow estimates.}
  \label{fig:new_examples}
\end{figure*}

\begin{table*}[h]
\vspace{-5pt} 
    \centering
    \resizebox{0.7\textwidth}{!}{%
    \begin{tblr}{
  colspec = {ccccc|ccc|ccc},
  cell{1}{1} = {r=2}{c}, 
  cell{1}{2} = {r=2}{c}, 
  cell{1}{3} = {r=2}{c}, 
  cell{1}{4} = {r=2}{c}, 
  cell{1}{5} = {r=2}{c}, 
  cell{1}{6} = {c=3}{c}, 
  cell{1}{9} = {c=3}{c}, 
  cell{3}{1} = {c=6}{l}, 
  cell{7}{1} = {c=6}{l}, 
  cell{10}{1} = {c=6}{l}, 
  rowsep=1pt,  
  colsep=4pt
}
\toprule
\cl & \ir & \ws & \il & \ltvm  & \textbf{EPE $\downarrow$} & & & \textbf{Smoothness (non-fault)$\downarrow$}  \\
& & & & & very small & small & medium & very small & small & medium\\
\midrule
~~\textit{Impact of \ir} \vspace{2pt} \\
\checkmark & \ding{55}& \ding{55} & \ding{55} & \ding{55} & 0.138\stddev{0.002}&0.236\stddev{0.001}&0.305\stddev{0.010}&0.039\stddev{0.002}&0.081\stddev{0.001}&0.188\stddev{0.006} \\
\checkmark & \checkmark & \ding{55} & \ding{55} & \ding{55} & 0.138\stddev{0.002}&0.237\stddev{0.003}&0.279\stddev{0.012}&0.082\stddev{0.004}&0.102\stddev{0.004}&0.174\stddev{0.006} \\
\checkmark & \checkmark & \checkmark & \ding{55} & \ding{55} & 0.189\stddev{0.006}&0.224\stddev{0.009}&0.249\stddev{0.012}&0.065\stddev{0.006}&0.094\stddev{0.007}&0.177\stddev{0.007} \\
\midrule
~~\textit{Impact of \il} \vspace{2pt} \\
\checkmark & \checkmark & \ding{55} & \checkmark & \ding{55} & 0.137\stddev{0.001}&0.227\stddev{0.006}&0.256\stddev{0.004}&0.077\stddev{0.004}&0.098\stddev{0.004}&0.183\stddev{0.003}\\
\checkmark & \checkmark & \checkmark & \checkmark & \ding{55} & 0.141\stddev{0.003}&0.216\stddev{0.007}&0.239\stddev{0.005}&0.067\stddev{0.004}&0.090\stddev{0.004}&0.176\stddev{0.005}\\
\midrule
~~\textit{Impact of \ltvm} \vspace{2pt} \\
\checkmark& \checkmark & \ding{55} & \checkmark & \checkmark &  0.137\stddev{0.001}&0.225\stddev{0.005}&0.254\stddev{0.004}&0.022\stddev{0.002}&0.045\stddev{0.002}&0.133\stddev{0.002}\\
\checkmark &  \checkmark & \checkmark & \checkmark & \checkmark & 0.140\stddev{0.003}&0.215\stddev{0.007}&0.238\stddev{0.005}&0.023\stddev{0.002}&0.045\stddev{0.001}&0.133\stddev{0.002}\\
\bottomrule
\end{tblr}

    }
    \vspace{-10pt} 
    \caption{\textbf{Multiple runs of the model} and its ablations for each component for seed 1, 5, 10.}
    \label{tab:multiseed}
    \vspace{-5pt} 
\end{table*}


\paragraph{Depth of the encoder-decoder network.}
Using a deeper architecture like FlowNet-S significantly degrades performance, as shown in Table 2. We hypothesize that this is due to the nature of our input data, which consists of satellite imagery containing predominantly small to very small displacements, making deeper networks less effective in this context. Consequently, unlike video motion estimation tasks that heavily rely on FlowNet-S or FlowNet-C, we base our model on the FlowNet-SD architecture.

\begin{figure*}[h]
    \centering
    \resizebox{\textwidth}{!}{
    \setlength\tabcolsep{1pt}
\begin{tblr}{
  colspec = {X[c,0.5]X[c,h]X[c,h]X[c,h]X[c,h]X[c,h]X[c,h]},
  stretch = 0,
  rowsep = 2pt,
  hlines = {red5, 0pt},
  vlines = {red5, 0pt},
}
& \textbf{$\lambda = 0$} &\textbf{$\lambda = 0.0001$} & \textbf{$\lambda = 0.001$} & \textbf{$\lambda = 0.01$} & \textbf{$\lambda = 0.1$} & \textbf{$\lambda = 1$}\\
(a) & \includegraphics[width=2.5cm]{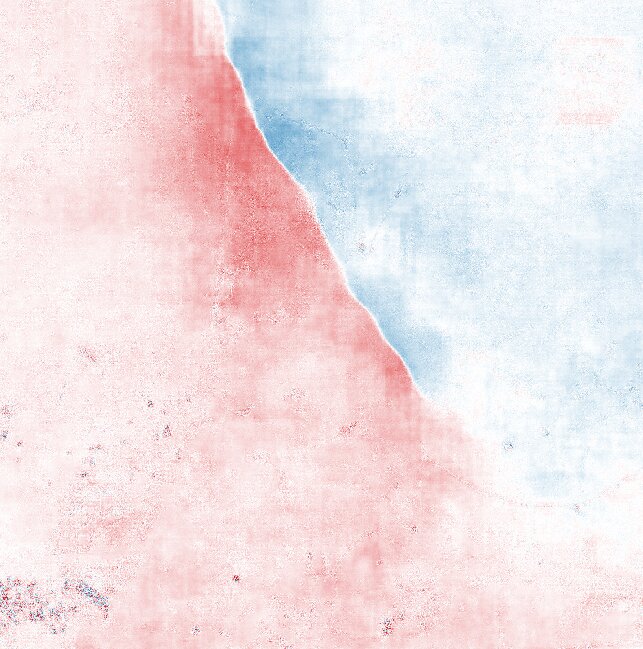} &  \includegraphics[width=2.5cm]{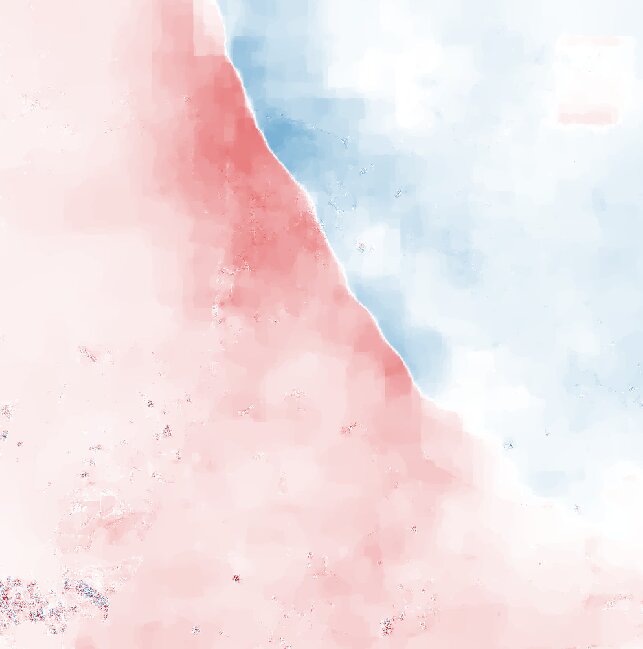}& 
\includegraphics[width=2.5cm]{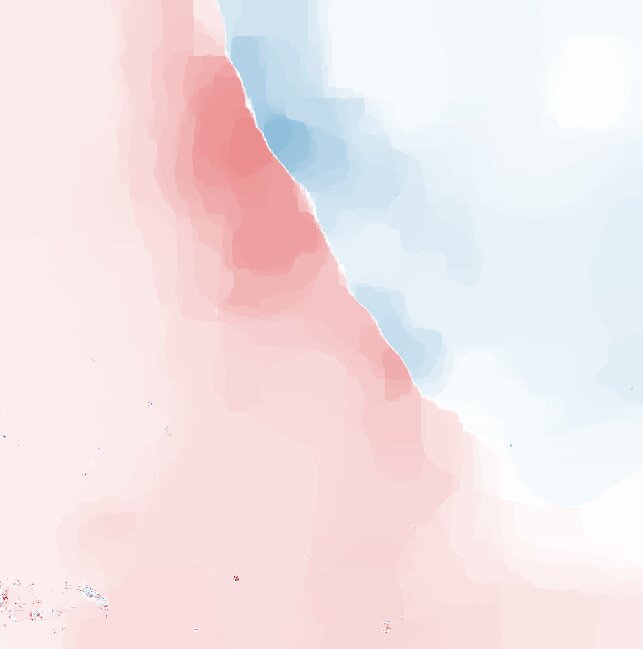}&
\includegraphics[width=2.5cm]{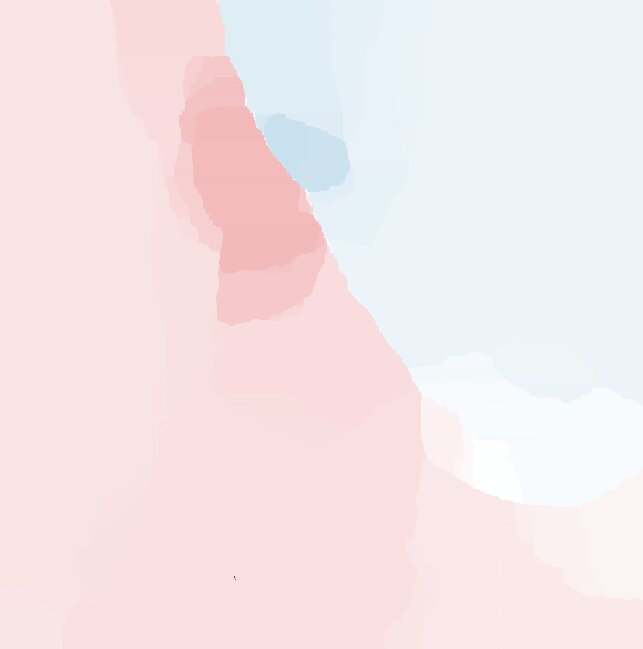}& 
\includegraphics[width=2.5cm]{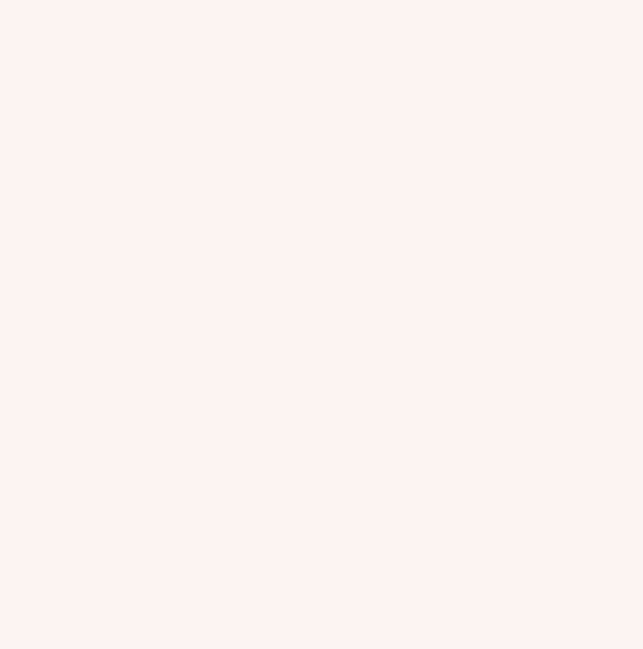}&
\includegraphics[width=2.5cm]{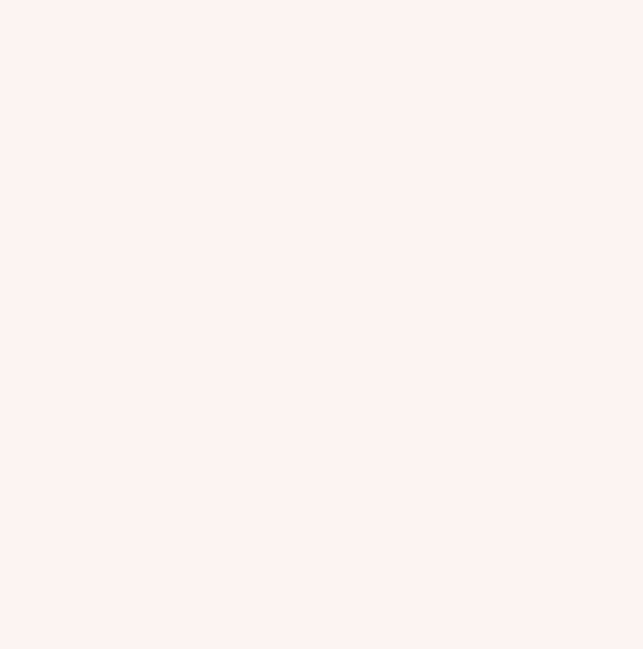} \\

(b) & \includegraphics[width=2.5cm]{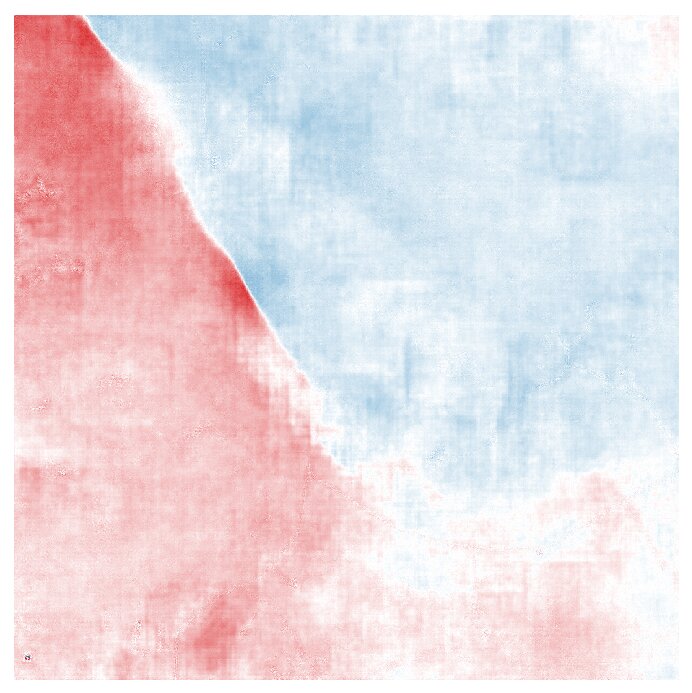} &  \includegraphics[width=2.5cm]{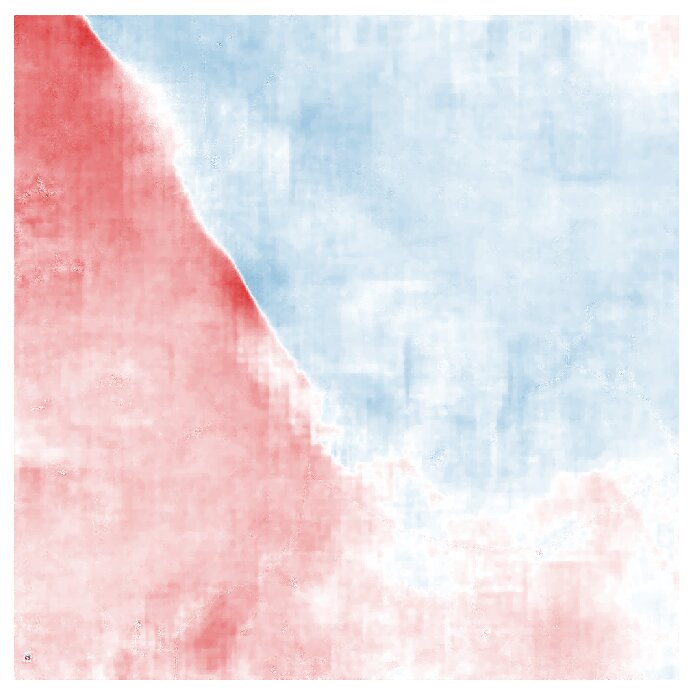}& 
\includegraphics[width=2.5cm]{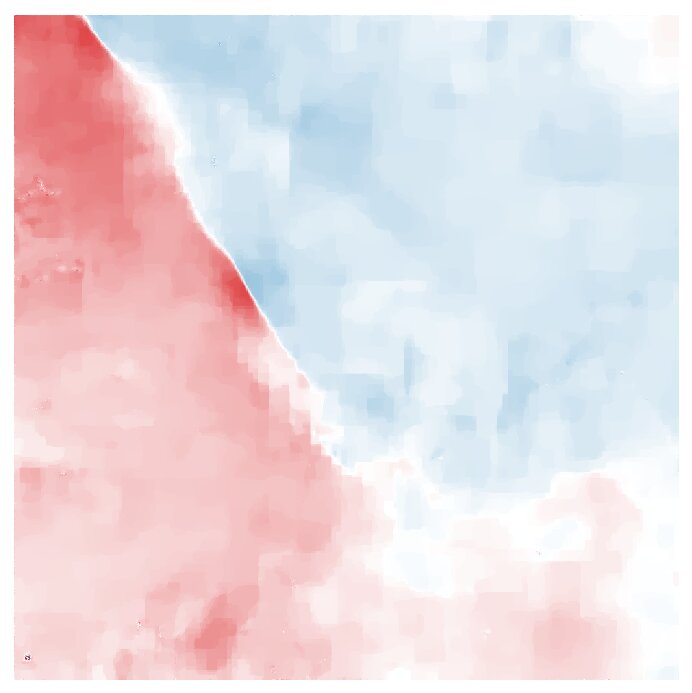}&
\includegraphics[width=2.5cm]{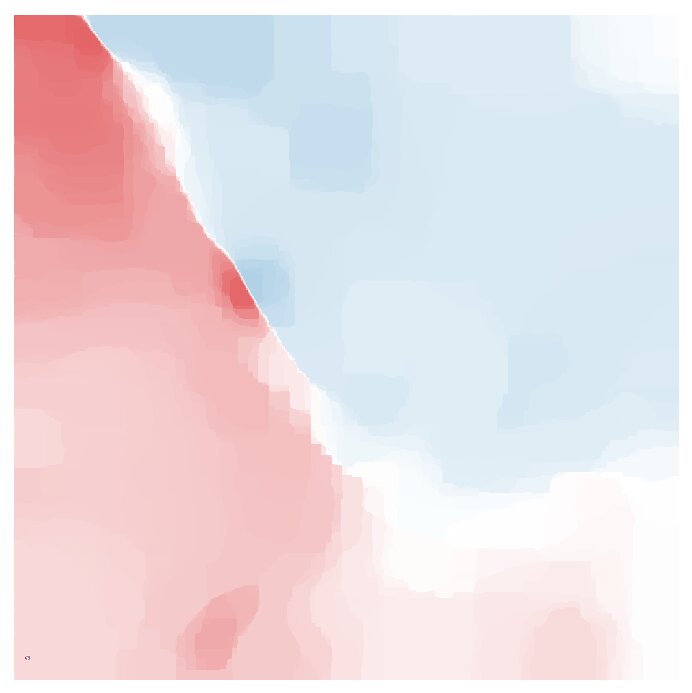}& 
\includegraphics[width=2.5cm]{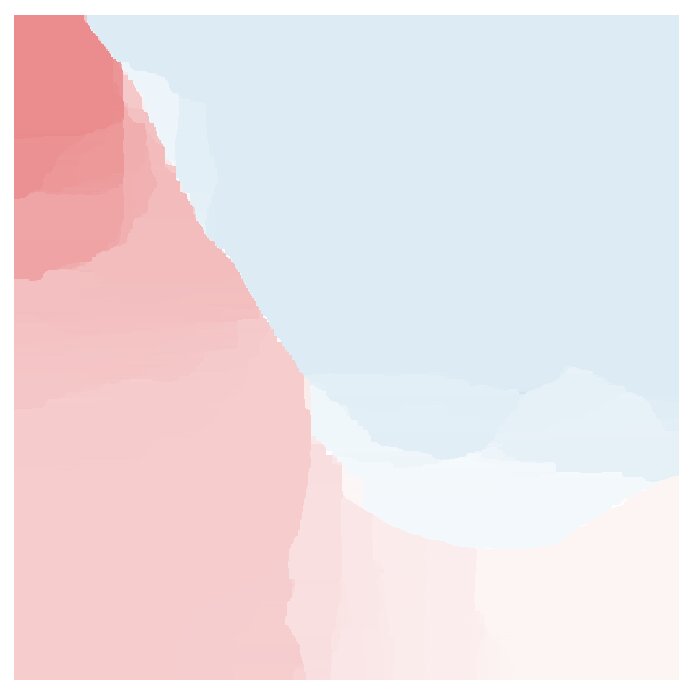}&
\includegraphics[width=2.5cm]{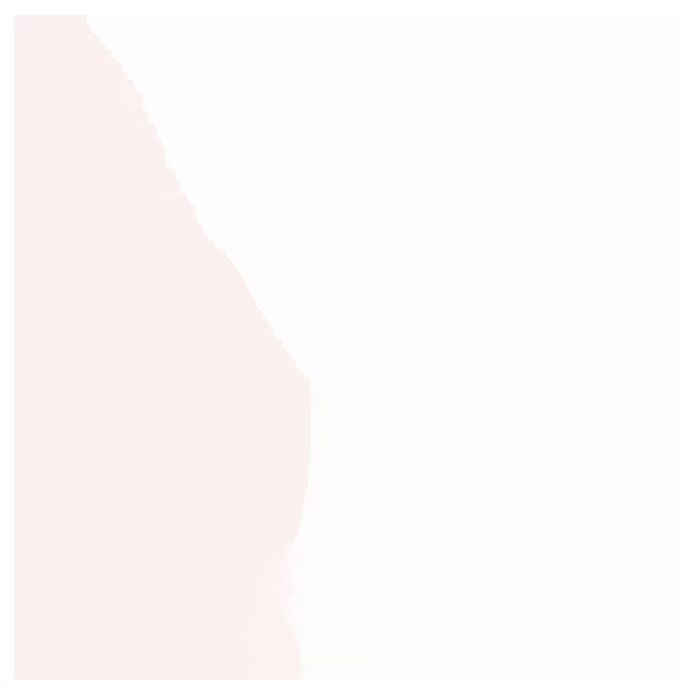} 
\\
(c) & \includegraphics[width=2.5cm]{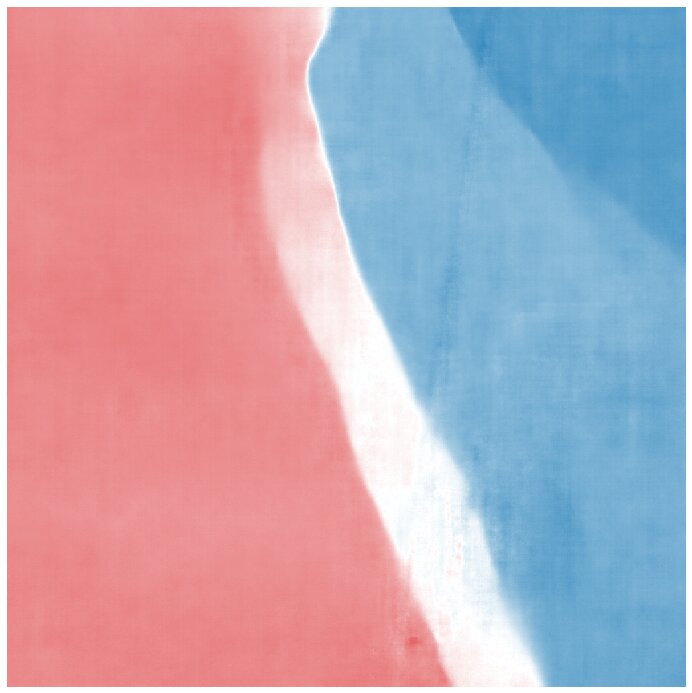} &  \includegraphics[width=2.5cm]{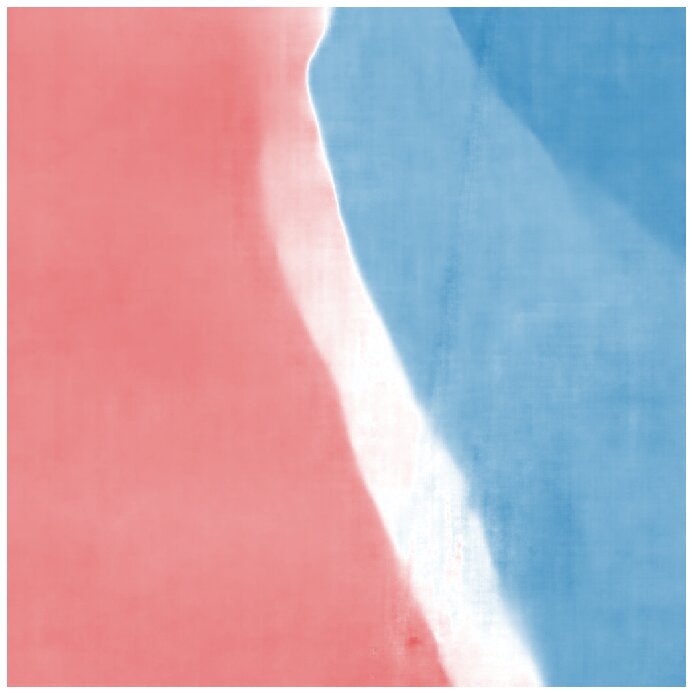}& 
\includegraphics[width=2.5cm]{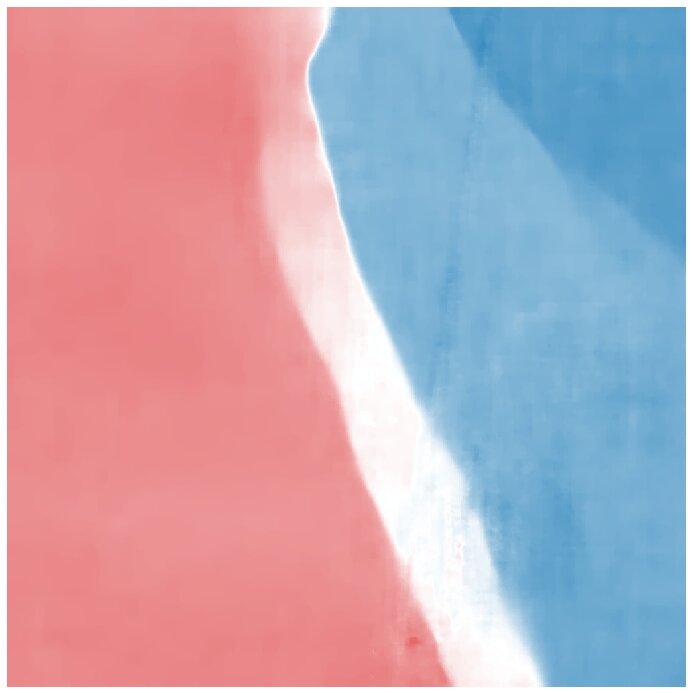}&
\includegraphics[width=2.5cm]{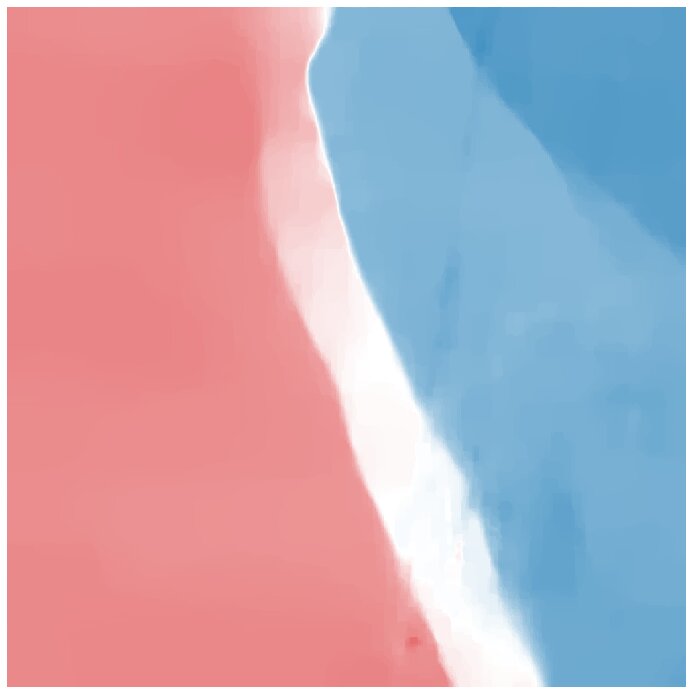}& 
\includegraphics[width=2.5cm]{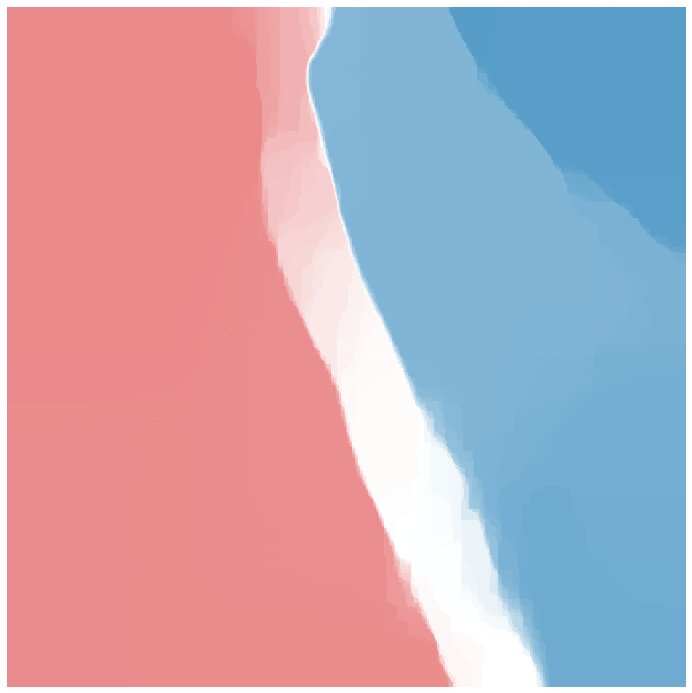}&
\includegraphics[width=2.5cm]{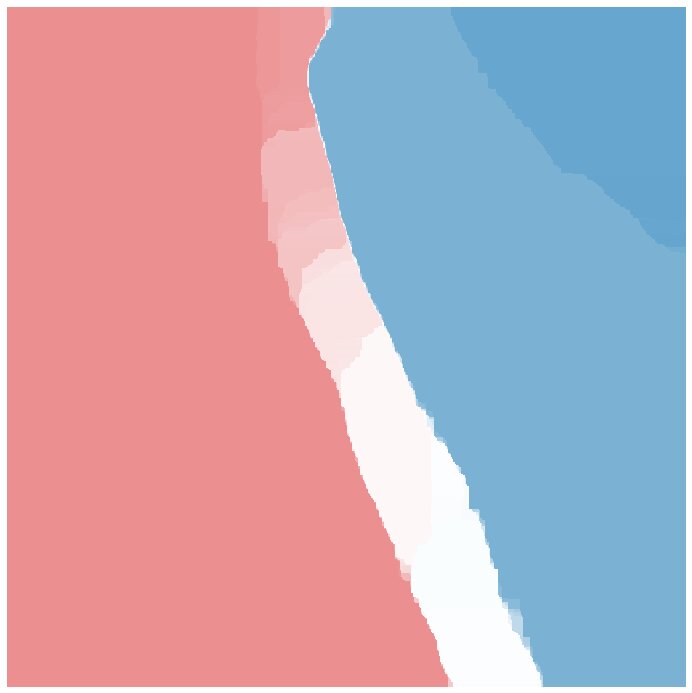} 
\\
(d) & \includegraphics[width=2.5cm]{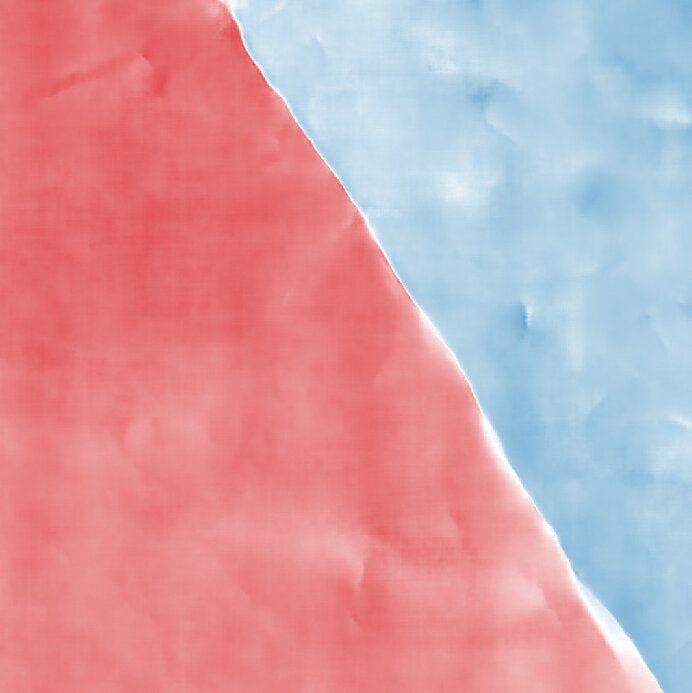} &  \includegraphics[width=2.5cm]{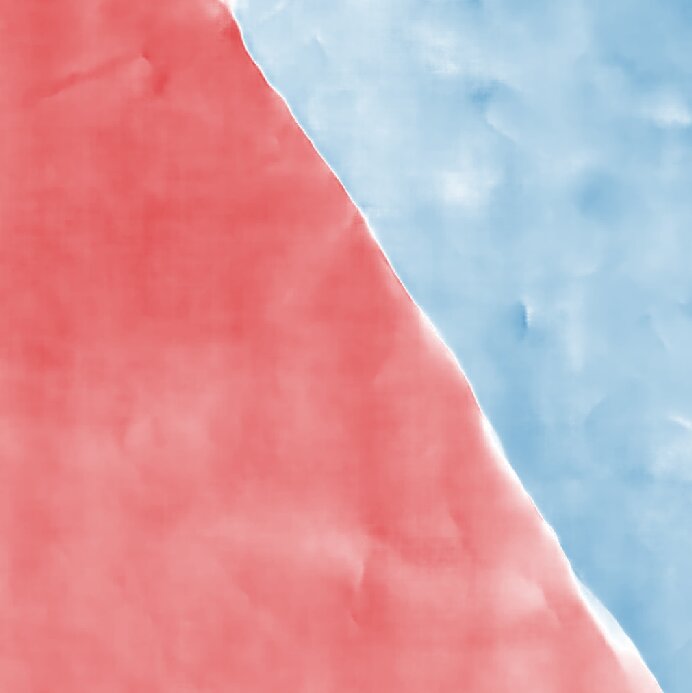}& 
\includegraphics[width=2.5cm]{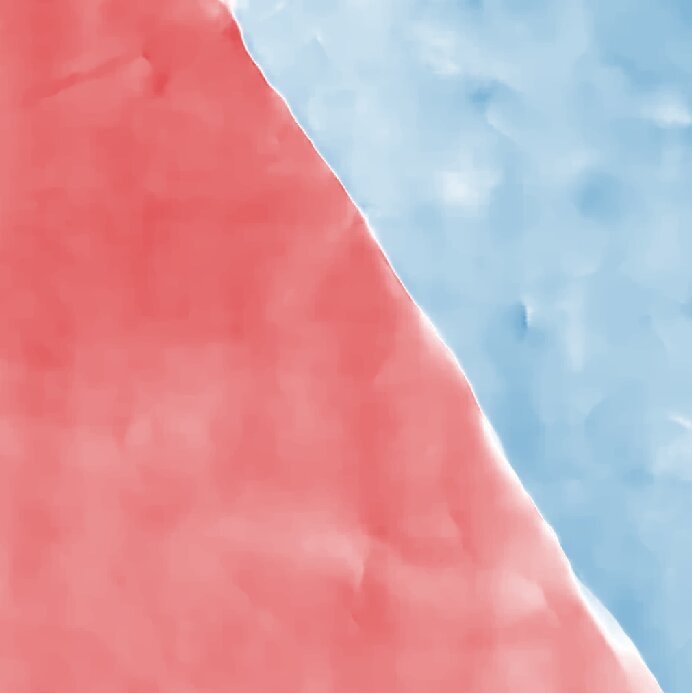}&
\includegraphics[width=2.5cm]{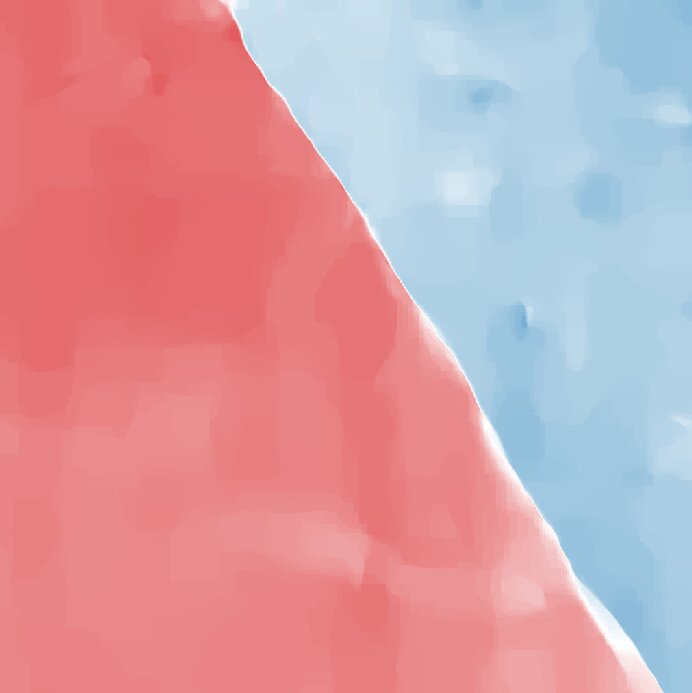}& 
\includegraphics[width=2.5cm]{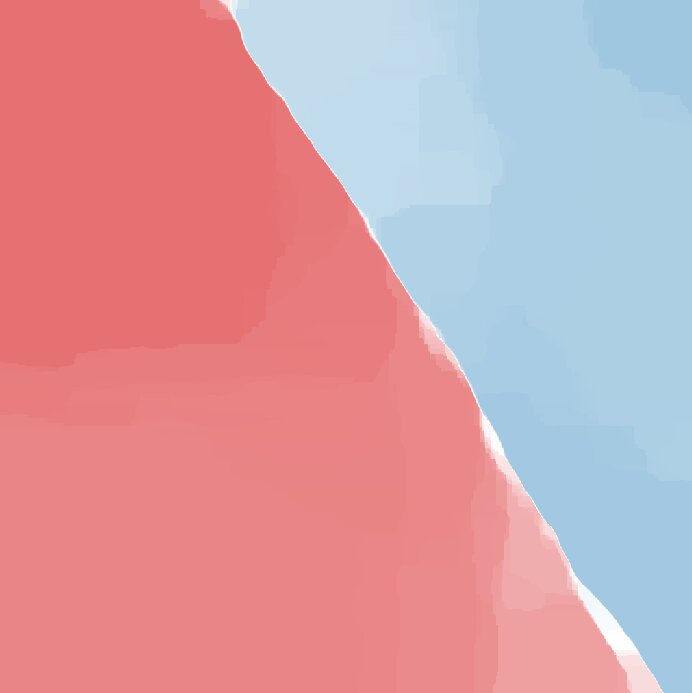}&
\includegraphics[width=2.5cm]{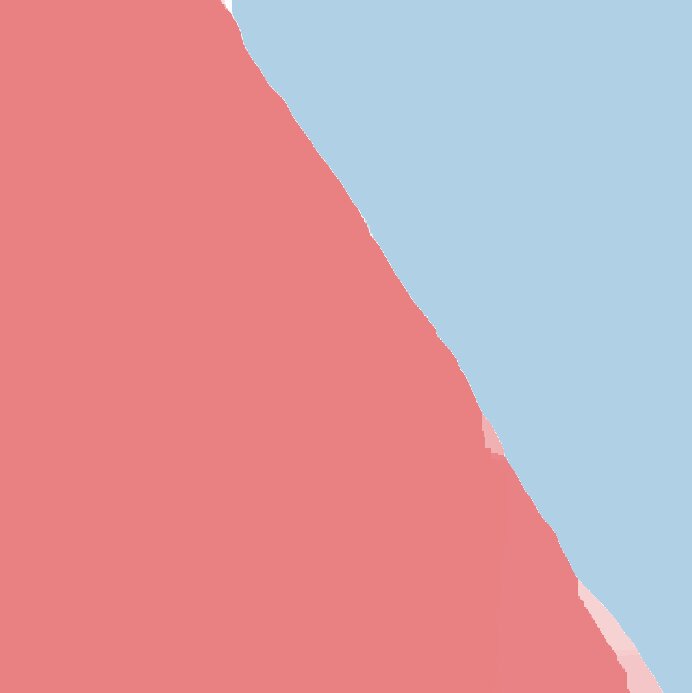} 
\\
\end{tblr}

    }
    \vspace{-10pt}
    \caption{\textbf{Effect of the $\lambda$ parameter in the \ltv regularization}, evaluated for the number of iteration $k=3$ and using images captured with different Earth observation sensors: a) \landsat (15m/pixel) b) \sentinel (10m/pixel) c) \ads (1m/pixel) d) \pleiades (50cm/pixel).}
    \label{fig:impact_lambda}
\end{figure*}

\begin{figure*}[h]
    \centering
    \resizebox{\textwidth}{!}{
    \setlength\tabcolsep{2pt}
\begin{tblr}{
  colspec = {X[c,h]X[c,h]X[c,h]X[c,h]X[c,h]X[c,h]},
  stretch = 0,
  rowsep = 2pt,
  hlines = {red5, 0pt},
  vlines = {red5, 0pt},
}
\textbf{$k = 0$} &\textbf{$k = 1$} & \textbf{$k = 2$} & \textbf{$k = 3$} & \textbf{$k = 4$} & \textbf{$k = 5$} \\
\includegraphics[width=2.7cm]{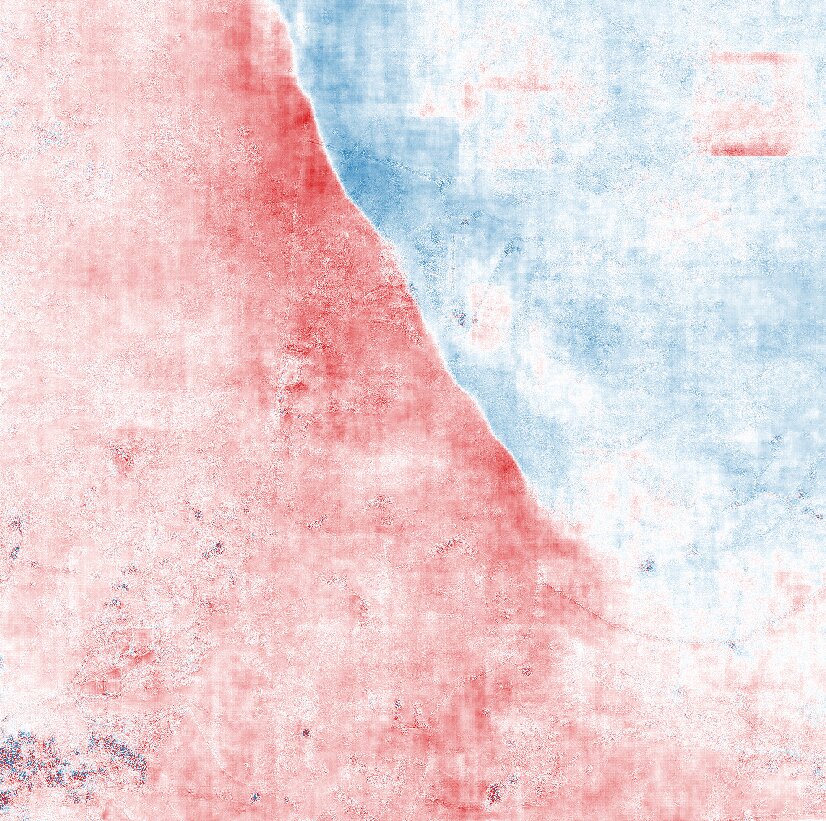} &  
\includegraphics[width=2.7cm]{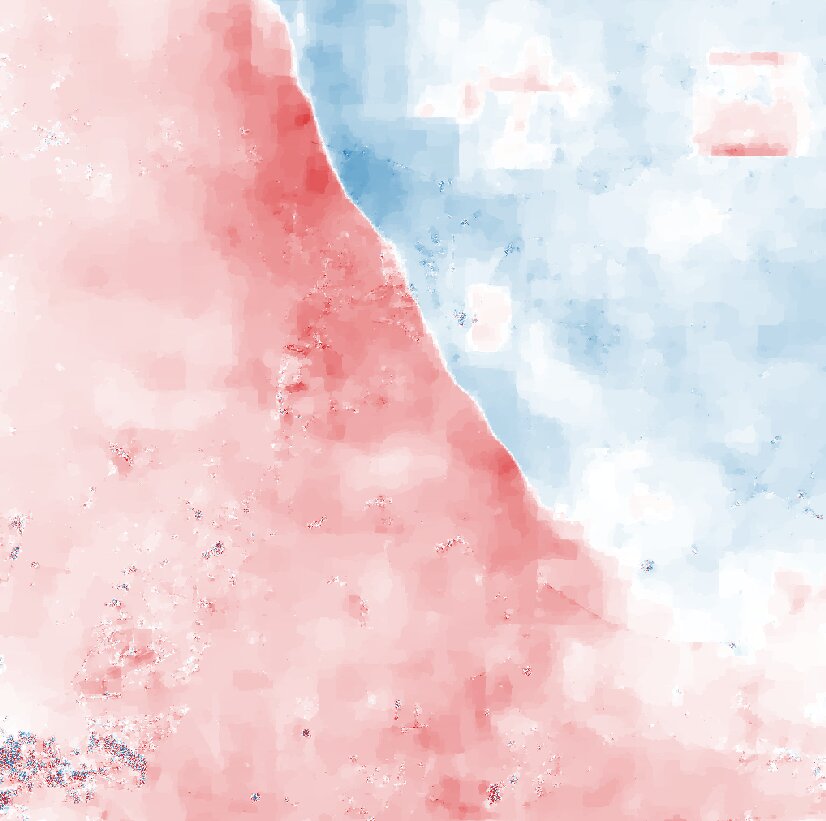}& 
\includegraphics[width=2.7cm]{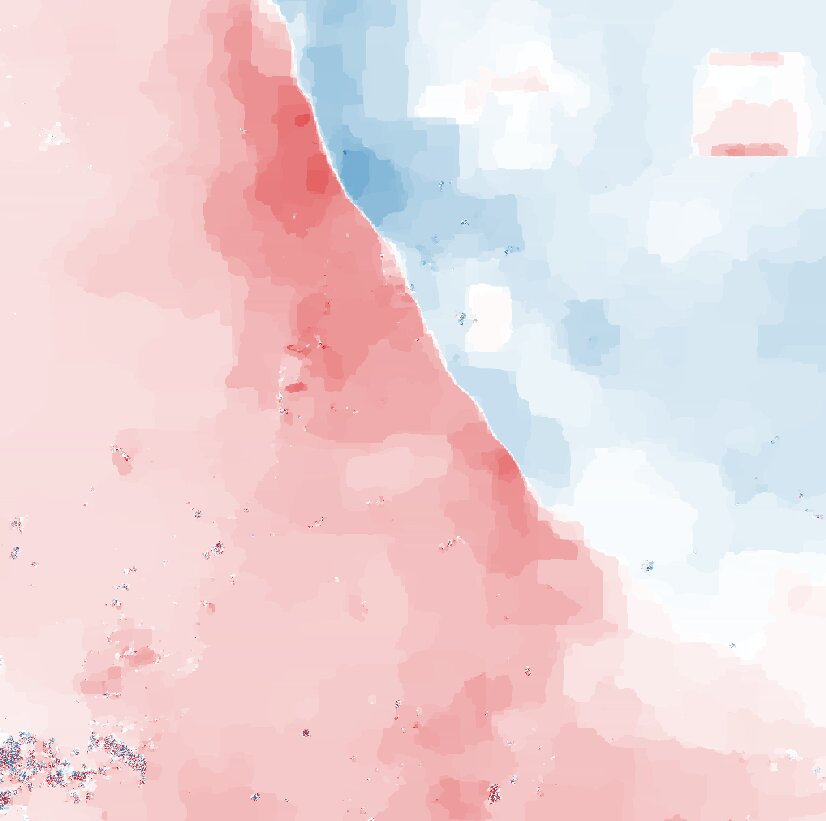}& 
\includegraphics[width=2.7cm]{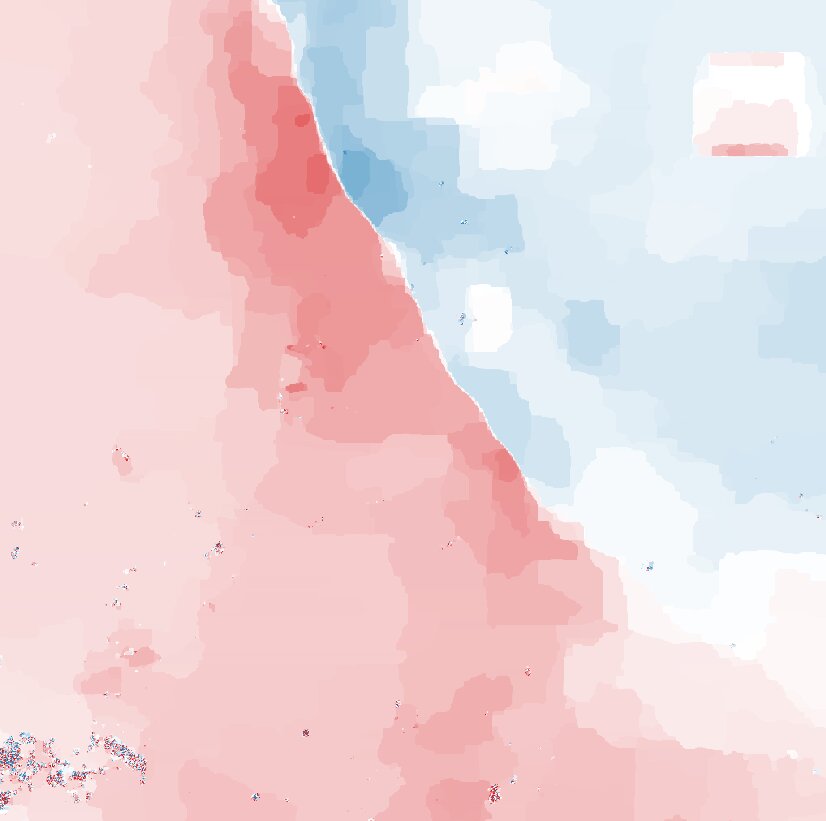}& 
\includegraphics[width=2.7cm]{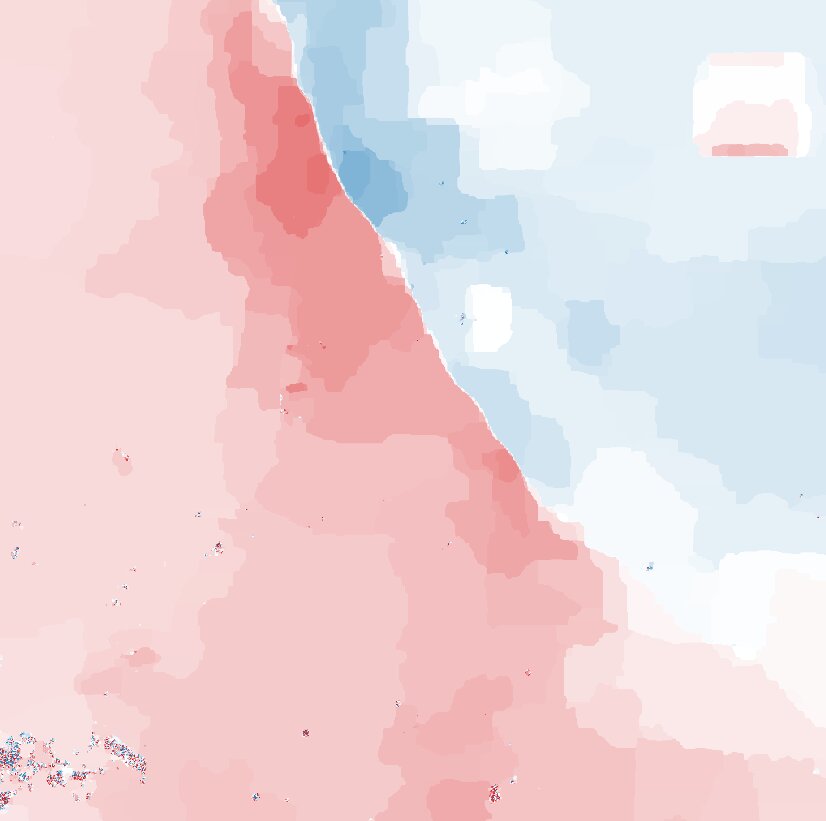}& 
\includegraphics[width=2.7cm]{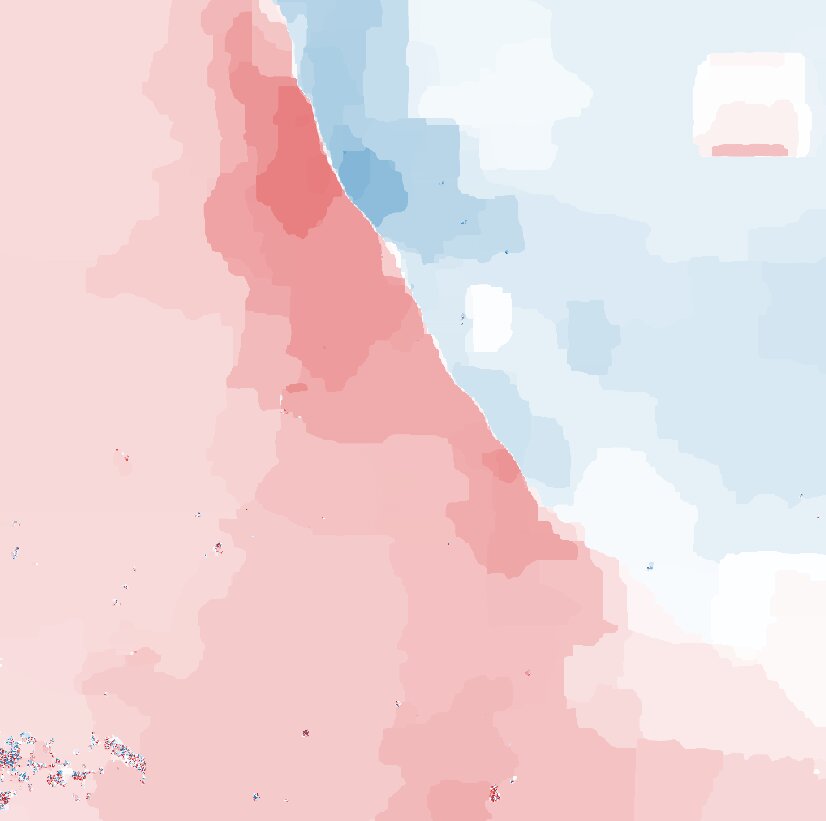}
\end{tblr}

    }
    \vspace{-10pt}
    \caption{\textbf{
    Effect of the number of iterations $k$ for approximating the \ltv penalty}, evaluated for $\lambda=0.001$ and using images captured with the \landsat satellites (15m/pixel). }
    \label{fig:impact_k}
\end{figure*}

\paragraph{Impact of the penalty functions.}

Figure \ref{fig:penalty_functions} shows a zoomed-in version of the denoised images in the $v$ direction produced for the three penalty functions for k varying between 1 and 5, and a fixed $\lambda$ value for each penalty function (specifically $\lambda=0.001$ for LTV, $\lambda=0.01$ for TV, $\lambda=0.1$ for L2 gradients). Our convex penalty allows to keep the fault line sharp while smoothing the noise away from the sharp while other penalty functions overly smooth the fault line.

\paragraph{Hyperparameters for \ltvm regularization.}

Table \ref{tab:ablation_ltv_smoothness} presents a quantitative ablation of $\lambda$ and $k$, with qualitative results in Figures \ref{fig:impact_lambda} and \ref{fig:impact_k}.
Excessively large $\lambda$ leads to over-smoothed solutions, while very small $\lambda$ keeps the signal noisy. We choose $\lambda=0.001$ as a balance. Though iterations improve results, they increase computational cost, so we set $k=3$ to balance smoothness and efficiency.

\begin{table*}[h]
    \centering
    \resizebox{0.7\textwidth}{!}{%
    \begin{tblr}{
  colspec = {ccccc|ccc|ccc},
  cell{1}{1} = {r=2}{c}, 
  cell{1}{2} = {r=2}{c}, 
  cell{1}{3} = {r=2}{c}, 
  cell{1}{4} = {r=2}{c}, 
  cell{1}{5} = {r=2}{c}, 
  cell{1}{6} = {c=3}{c}, 
  cell{1}{9} = {c=3}{c}, 
  cell{3}{1} = {c=6}{l}, 
  cell{7}{1} = {c=6}{l}, 
  cell{10}{1} = {c=6}{l}, 
  rowsep=1pt,  
  colsep=4pt
}
\toprule
\cl & \ir & \ws & \il & \ltvm  & \textbf{EPE $\downarrow$} & & & \textbf{Smoothness (non-fault)$\downarrow$}  \\
& & & & & very small & small & medium & very small & small & medium\\
\midrule
~~\textit{Impact of \ir} \vspace{2pt} \\
\checkmark & \ding{55}& \ding{55} & \ding{55} & \ding{55} & 0.135\stddev{0.002}&0.762\stddev{0.007}&3.743\stddev{0.002}&0.039\stddev{0.002}&0.081\stddev{0.006}&0.071\stddev{0.009} \\
\checkmark & \checkmark & \ding{55} & \ding{55} & \ding{55} & 0.134\stddev{0.002}&0.782\stddev{0.018}&4.164\stddev{0.015}&0.080\stddev{0.016}&0.130\stddev{0.013}&0.146\stddev{0.022} \\
\checkmark & \checkmark & \checkmark & \ding{55} & \ding{55} & 0.134\stddev{0.003}&0.950\stddev{0.012}&4.261\stddev{0.006}&0.061\stddev{0.006}&0.097\stddev{0.005}&0.100\stddev{0.004} \\
\midrule
~~\textit{Impact of \il} \vspace{2pt} \\
\checkmark & \checkmark & \ding{55} & \checkmark & \ding{55} & 0.133\stddev{0.002}&0.599\stddev{0.021}&3.632\stddev{0.017}&0.074\stddev{0.003}&0.132\stddev{0.002}&0.148\stddev{0.004}\\
\checkmark & \checkmark & \checkmark & \checkmark & \ding{55} & 0.133\stddev{0.003}&0.804\stddev{0.007}&3.760\stddev{0.006}&0.053\stddev{0.001}&0.090\stddev{0.001}&0.088\stddev{0.002}\\
\midrule
~~\textit{Impact of \ltvm} \vspace{2pt} \\
\checkmark& \checkmark & \ding{55} & \checkmark & \checkmark &  0.132\stddev{0.002}&0.597\stddev{0.021}&3.632\stddev{0.017}&0.021\stddev{0.003}&0.059\stddev{0.002}&0.065\stddev{0.004}\\
\checkmark &  \checkmark & \checkmark & \checkmark & \checkmark & 0.132\stddev{0.002}&0.803\stddev{0.007}&3.760\stddev{0.007}&0.013\stddev{0.001}&0.034\stddev{0.001}&0.029\stddev{0.002}\\
\bottomrule
\end{tblr}

    }
    \vspace{-10pt} 
    \caption{\textbf{Quantitative analysis when training a specialized model on very small displacements only} for multiple runs (seed 1, 5, 10).}
    \label{tab:multiseed_very_small}
    \vspace{-5pt} 
\end{table*}


\paragraph{Impact of multiple runs of the model.}
We report mean and standard deviation after running our models and its component ablation with 3 different seeds (1, 5, 10) in Table \ref{tab:multiseed}, while Table 2 and 3 (main paper) report the results for a seed of 1.

\paragraph{Training a specialized model for very small displacements only.}
To improve the estimation of very small displacement fields, we train a model specifically tailored to this range. The mean and standard deviation of its performance are reported in Table \ref{tab:multiseed_very_small}. As anticipated, the model does not generalize well to small or medium displacements. However, it delivers improved performance on very small displacements, which is consistent with observations from real-world examples. Following those observations, we report the estimate for this model in Table 2 and 3 of the main paper.

\paragraph{Additional examples of temporal changes.}
The time gaps (weeks to months) between acquisition introduce temporal changes, which affects the estimates of the different model. Figure \ref{fig:new_examples} provides additional examples of those perturbations. While the classical baselines present high noise, our proposed method, \microflow, largely reduces the noise.

\begin{figure*}[h]
    \begin{subfigure}{\linewidth}
        \centering
        \includegraphics[width=0.55\linewidth]{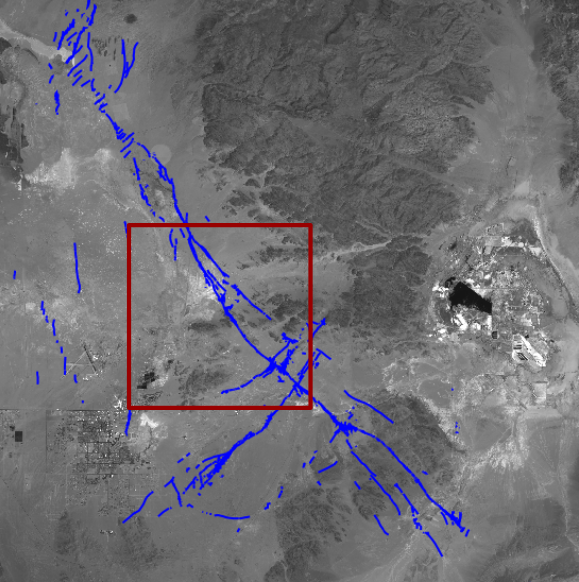}
    \end{subfigure}
    \vspace{-10pt}
    \caption{\textbf{The joint post-seismic image of the Ridgecrest area acquired with \landsat sensor (15m/pixel)}. The full image is 2991x2980 pixels², and is composed of 9 pairs of 1024x1024 pixels² for our dataset. Expert annotations are overlaid in blue. The dark red highlighted rectangle corresponds to the zoomed-in region depicted in Figure \ref{fig:ridgecrest_dataset}.}
    \label{fig:ridgecrest_landsat}
\end{figure*}

\section{Additional details}

\begin{figure*}[h]
    \begin{subfigure}{\linewidth}
        \centering
        \includegraphics[width=0.55\linewidth]{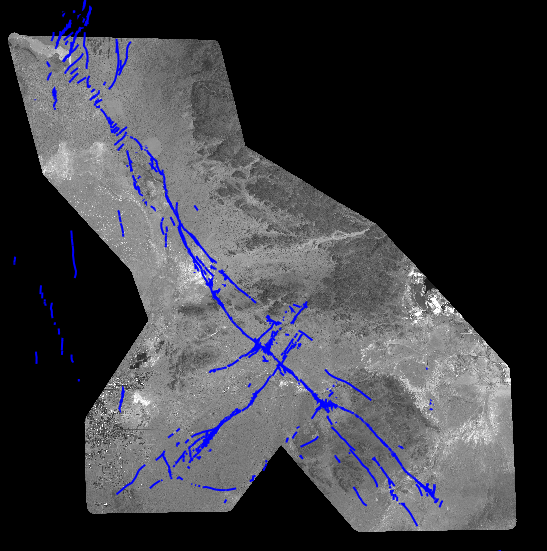}
    \end{subfigure}
    \vspace{-10pt}
    \caption{\textbf{The joint post-seismic image of the Ridgecrest area acquired with SPOT-6 sensor (2m/pixel)}. The full image is 20000 pixels², and is composed of 195 pairs of 1024x1024 pixels² for our dataset. Expert annotations are overlaid in blue.}
    \label{fig:ridgecrest_spot}
\end{figure*}

\begin{figure*}[h]
    
    \begin{subfigure}{\linewidth}
        \centering
        \includegraphics[width=0.7\linewidth]{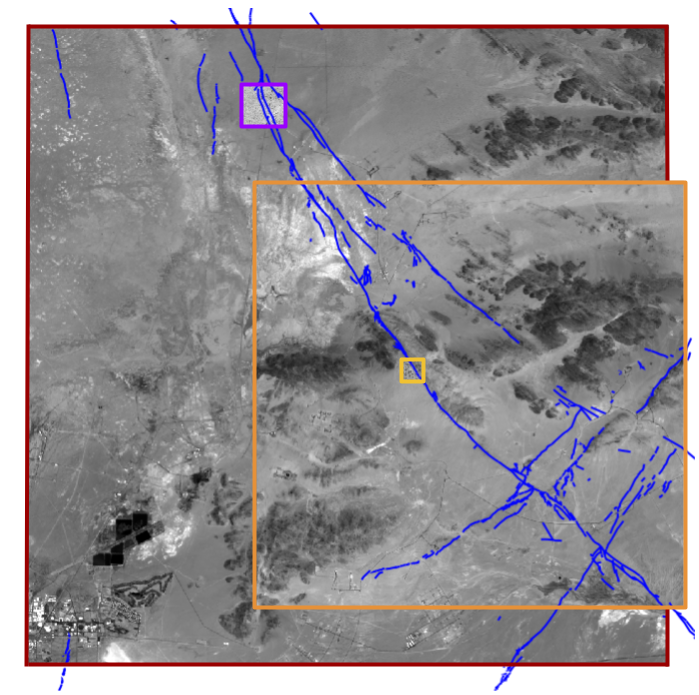}
    \end{subfigure}
    \vspace{1pt}
    \begin{subfigure}{\linewidth}
        \centering
        \includegraphics[width=0.9\linewidth]{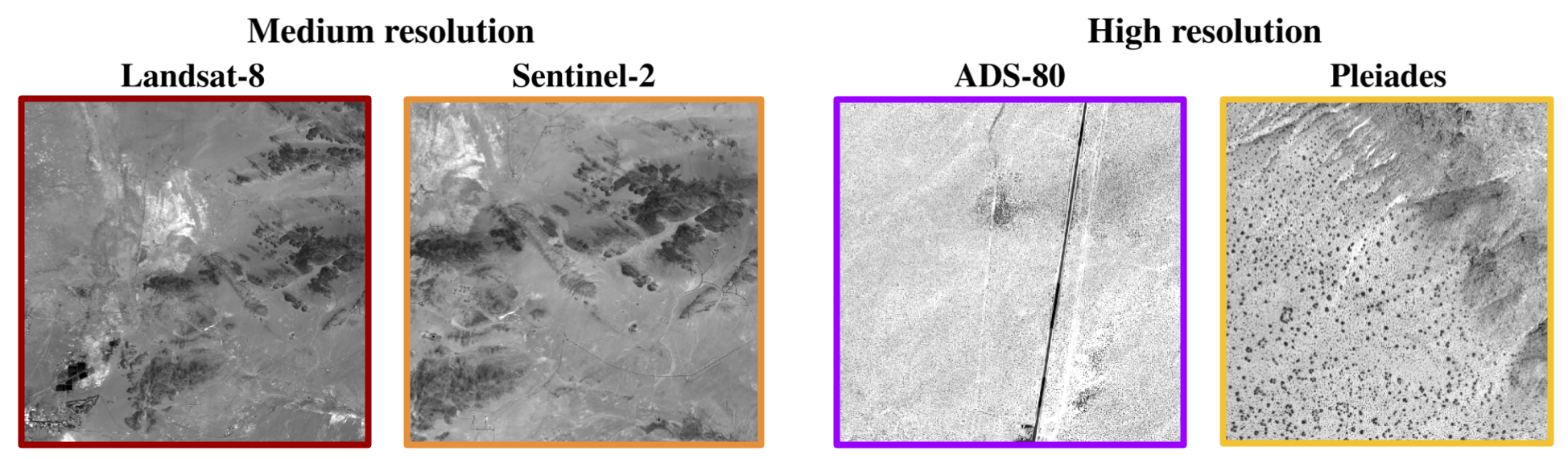}
    \end{subfigure}
    \vspace{-10pt}
    \caption{\textbf{Four post-seismic images of the RidgeCrest area}, captured using two medium-resolution satellites: \landsat (15m/pixel, dark red) and \sentinel (10m/pixel, orange), and two high-resolution sensors: \ads (1m/pixel, purple) and \pleiades (50cm/pixel, yellow). Expert annotations are overlaid in blue.}
    \label{fig:ridgecrest_dataset}

\end{figure*}

\subsection{Optical geodesy sensors.}
Early satellite sensors (Landsat-4) provided medium spatial resolutions ranging from 30 to 80 meters in multi-spectral bands due to technical limitations in camera optics, sensor design, data storage, and processing power. This medium resolution allowed sensors to cover large ground areas without overwhelming the systems. As technology advanced, satellite capabilities improved significantly (SPOT-6, Sentinel2, Landsat8) enabling resolutions as fine as 30-40 cm in the panchromatic band (WorldView4, GeoEye1).
However, high-resolution imagery requires lower-altitude flights or orbits, capturing finer details per pixel but limiting the field of view. Covering larger areas with high-resolution images involves multiple passes, increasing costs, processing time, and complexity while also raising the risk of interference, such as cloud cover, which complicates data acquisition.
The choice between medium and high spatial resolution imagery depends on the specific applications that rely on ground deformation estimates as inputs and their required level of precision, whether they need accuracy at the centimeter scale or coverage at the kilometer scale. A model capable of effectively handling both large and small displacements offers greater flexibility for various scientific applications.

\begin{figure*}[h]
    
    \setlength\tabcolsep{2pt}
\begin{tblr}{
  colspec = {X[c,h]X[c,h]X[c,0.2]},
  stretch = 0,
  rowsep = 2pt,
  hlines = {red5, 0pt},
  vlines = {red5, 0pt},
}
\textbf{\cosicorr} & \textbf{\raft}\\
\includegraphics[width=6cm]{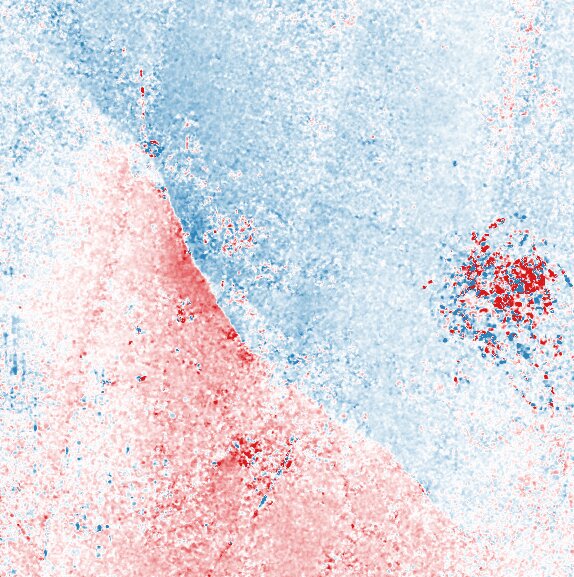} & 
\includegraphics[width=6cm]{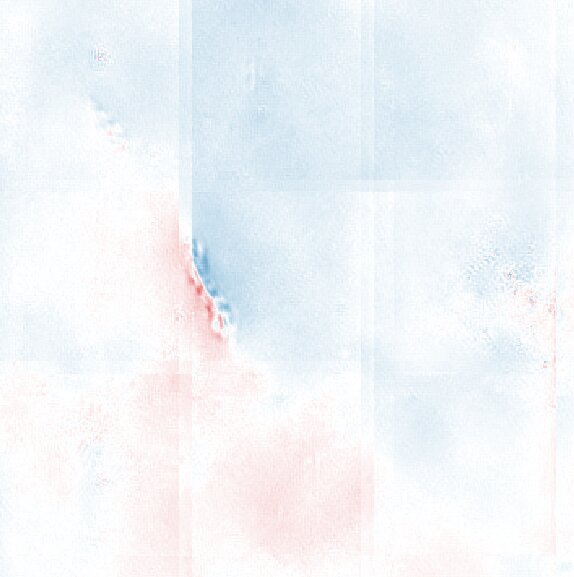} & \includegraphics[height=1.8cm]{fig/00_scales/scale_landsat.png}
\\
\\
\textbf{\ir without \ltvm} & \textbf{\gfn} \\
\includegraphics[width=6cm]{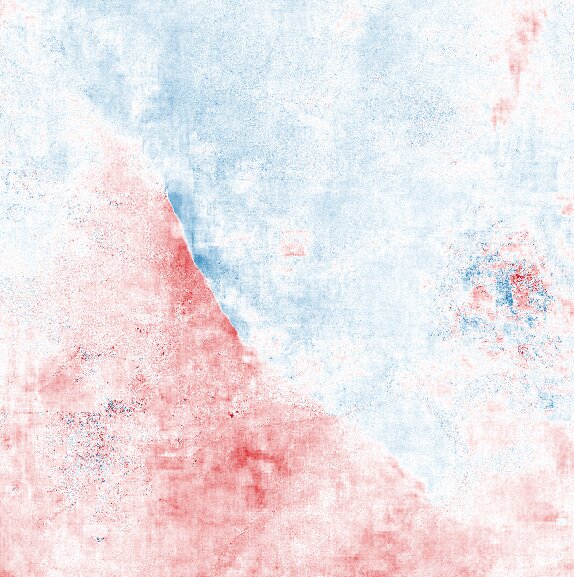}&
\includegraphics[width=6cm]{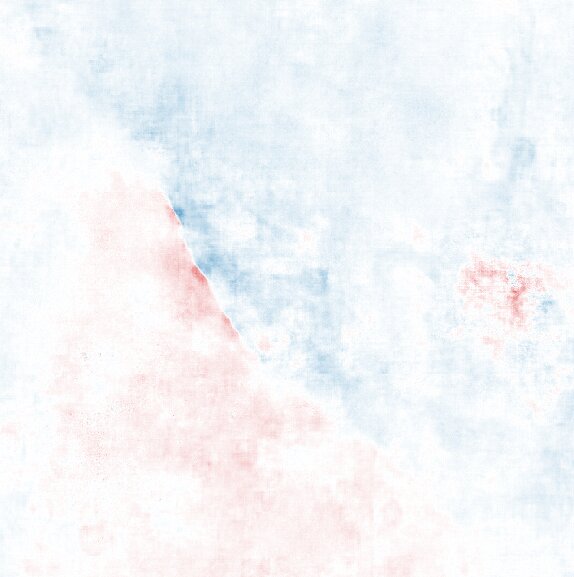} &
\\ 
\\
\textbf{MicroFlow ($\lambda=0.001$)} & \textbf{MicroFlow ($\lambda=0.01$)}\\
\includegraphics[width=6cm]{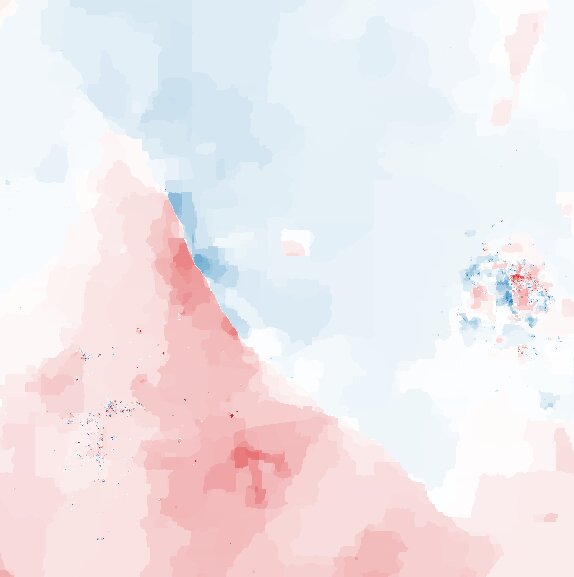} &
\includegraphics[width=6cm]{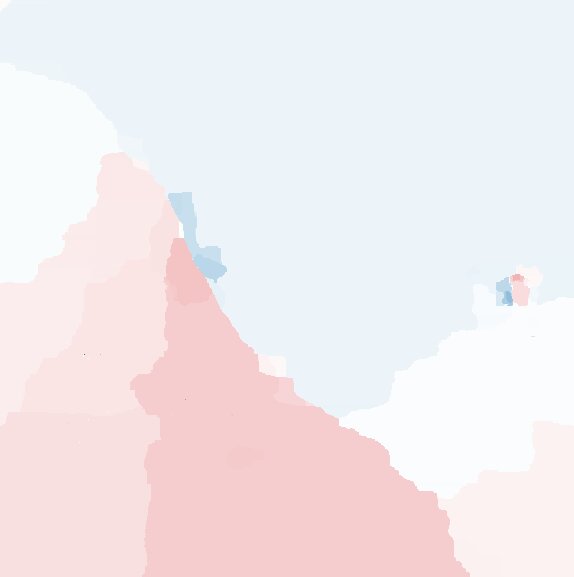} &

\end{tblr}

    \vspace{-10pt}
    \caption{North-south ($v$ direction) ground deformation estimates across the entire Ridgecrest area acquired from \landsat sensors.}
    \label{fig:result_ridgecrest_landsat_ns}

\end{figure*}

\begin{figure*}[h]
    
    \setlength\tabcolsep{2pt}
\begin{tblr}{
  colspec = {X[c,h]X[c,h]X[c,0.2]},
  stretch = 0,
  rowsep = 2pt,
  hlines = {red5, 0pt},
  vlines = {red5, 0pt},
}
\textbf{\cosicorr} & \textbf{\raft}\\
\includegraphics[width=6cm]{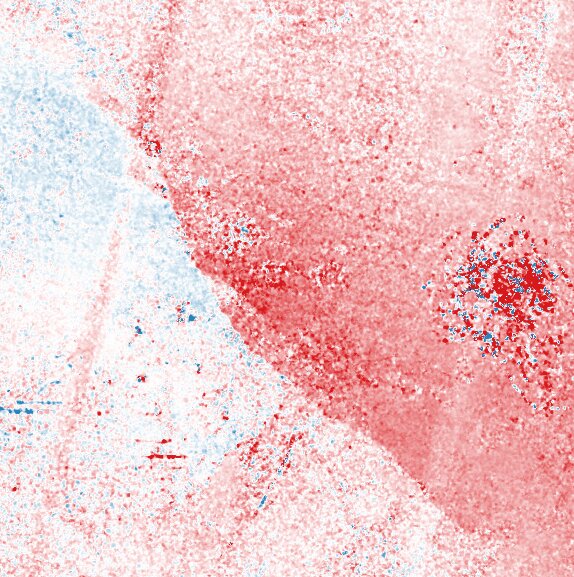} & 
\includegraphics[width=6cm]{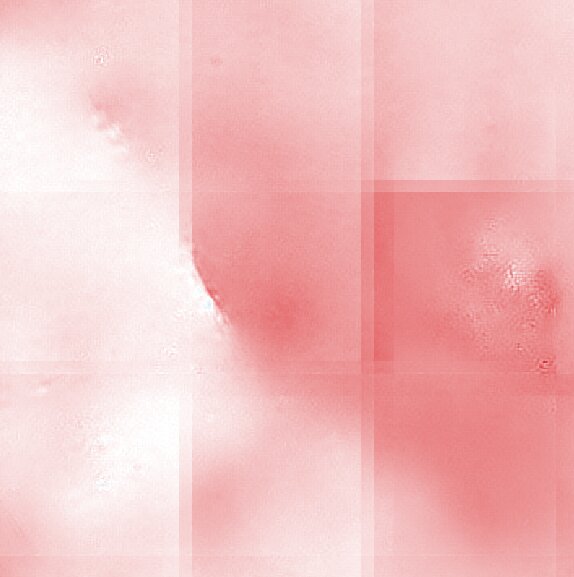} & \includegraphics[height=1.8cm]{fig/00_scales/scale_landsat.png}\\
\\
\textbf{\ir without \ltvm} & \textbf{\gfn} \\
\includegraphics[width=6cm]{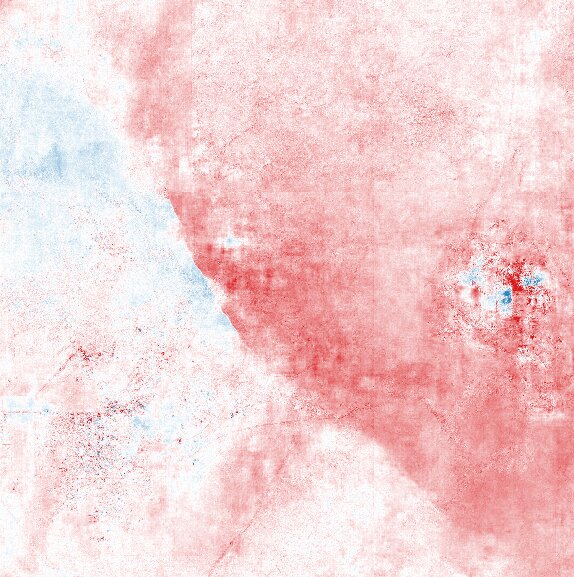}&
\includegraphics[width=6cm]{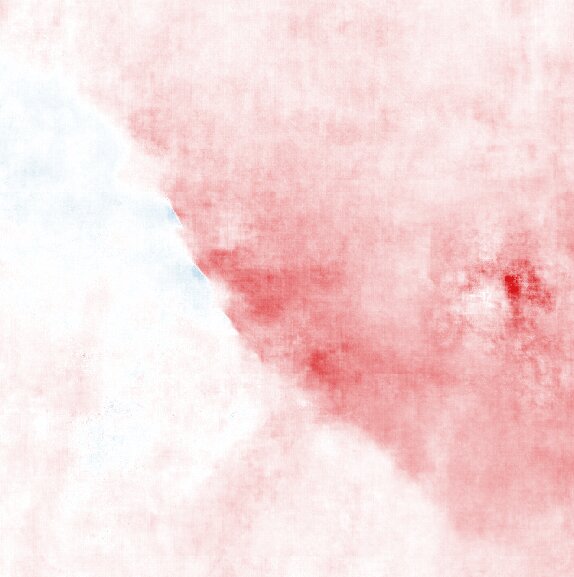} &\\ 
\\
\textbf{MicroFlow ($\lambda=0.001$)} & \textbf{MicroFlow ($\lambda=0.01$)}\\
\includegraphics[width=6cm]{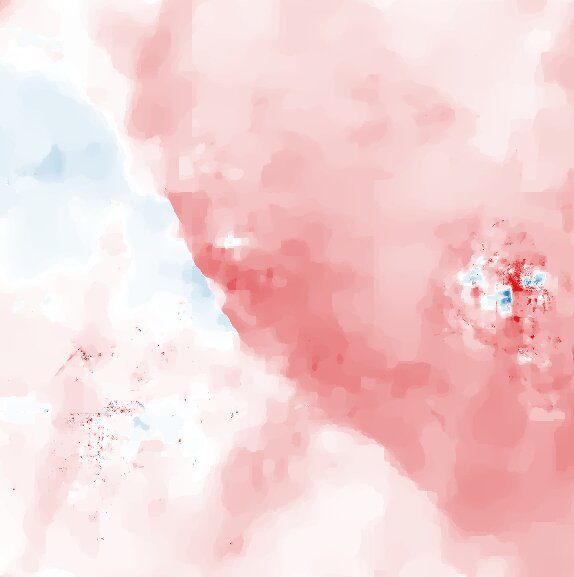} &
\includegraphics[width=6cm]{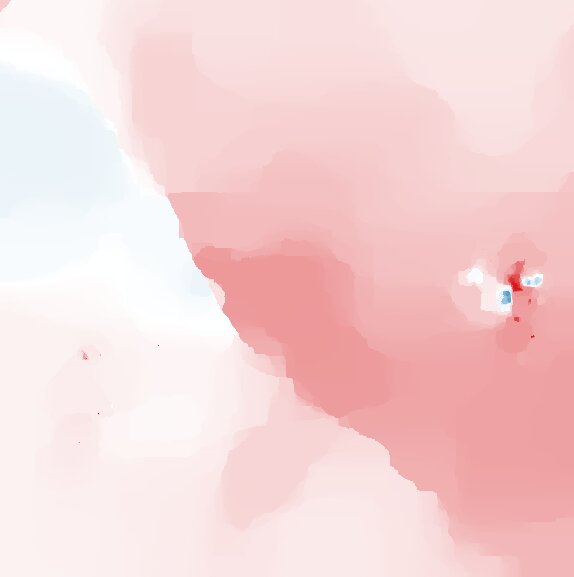} &

\end{tblr}

    \vspace{-10pt}
    \caption{East-west ($u$ direction) ground deformation estimates across the entire Ridgecrest area acquired from \landsat sensors.}
    \label{fig:result_ridgecrest_landsat_ew}

\end{figure*}

\myparagraph{High resolution sensors}
The Airborne Digital Sensor, \ads, operated by Leica Geosystems, has been in use since 2008. It employs a pushbroom scanning technique, capturing high-resolution imagery line-by-line as an aircraft flies over the area of interest. This method results in continuous, linear strips of data with a spatial resolution of 1 meter, ensuring consistent image quality throughout the surveyed region.

The Pleiades Satellite Constellation, \pleiades, operated by the French space agency CNES since 2011, consists of two agile Earth observation satellites. These satellites also use a pushbroom scanning technique to capture imagery line-by-line as they orbit the Earth. Unlike the \ads, the \pleiades satellites collect targeted scenes or swaths rather than continuous strips. They achieve a very high spatial resolution of up to 50 cm per pixel, enabling detailed observations of the Earth's surface.

SPOT-6 is a commercial Earth-imaging satellite owned and operated by Airbus Defence and Spaceoptical launched in 2012.  The SPOT images have a 60-km footprint and resolution of 1.5 m.

\myparagraph{Medium resolution sensors}
The \landsat satellite, operated jointly by the United States Geological Survey (USGS) and NASA, collects medium-resolution imagery in both multi-spectral and panchromatic bands through a push-broom scanning technique. It provides a spatial resolution of 15 meters for the panchromatic band, facilitating extensive land coverage. 

The \sentinel satellites, operated by the European Space Agency (ESA), collect high-resolution multi-spectral imagery using a push-broom scanning technique. The mission consists of two identical satellites, Sentinel-2A and Sentinel-2B, which provide a spatial resolution ranging from 10 to 60 meters across 13 spectral bands.

\subsection{RidgeCrest-2019 evaluation, details about image acquisition.}
Our evaluation focuses on the Ridgecrest earthquake zone in California, an area of 45x45 km², where three significant shocks in 2019 caused substantial earth deformations. Studying these deformations offers insights into broader geological processes and their potential influence on future earthquakes that could impact millions of residents. The images used in this study were acquired on September 15, 2018 and July 24, 2019. They capture surface motion of both the foreshock and mainshock Ridgecrest events. We group them into three sub-datasets of pre- and post-seismic images of this area.  

The first dataset (Figure \ref{fig:ridgecrest_landsat} contains 9 medium resolution image pairs acquired with the \landsat sensor over the entire region. 
The second dataset (Figure \ref{fig:ridgecrest_spot} contains 195 high resolution image pairs acquired with the SPOT-6 sensor over the entire region \footnote{\url{https://journals.sagepub.com/doi/pdf/10.1177/87552930241280255}}.
The third dataset contains 2 high-resolution images acquired with the \ads and \pleiades sensors and 2 medium-resolution images acquired with the \landsat and \sentinel sensors.

\subsection{Training details}
During training of the iterative encoder-decoder network, we use the Adam optimizer and clip gradients to the range $[-1, 1]$. The batch-size is fixed to 32. We use random cropping, random horizontal flipping and random vertical flipping as data augmentations.
For the \gfn model, we adopt the learning rate schedule proposed by \cite{Montagnon24}, starting with an initial learning rate of 0.0001. The learning rate is halved every 20 epochs over the course of the training on a dataset of 20,000 examples. For the iterative models, we begin with a learning rate of 0.0001, and apply the same schedule.
We stop training every model after 40 epochs. 

\section{Additional qualitative analysis}
Figures \ref{fig:result_ridgecrest_landsat_ns} and \ref{fig:result_ridgecrest_landsat_ew} compare the classical \cosicorr method, known for photometric accuracy, with three deep learning models—\raft, \gfn, and \microflownospace—trained only on the FaultDeform dataset without finetuning or test-time training.

As discussed earlier, \raft struggles with accuracy and fails to precisely locate the fault. \gfn quickly fades away from the fault, deviating from \cosicorr. In contrast, \microflow maintains non-zero deformation far from the fault, achieves high photometric quality, and significantly reduces noise, illustrating its potential for large-scale analysis.

\end{document}